\documentclass[10pt,twocolumn,letterpaper]{article}

\usepackage[pagenumbers]{cvpr}   %

\usepackage{booktabs}
\usepackage{multirow}
\usepackage{amsmath,amssymb}
\usepackage{xspace}
\usepackage{bm}
\usepackage{dblfloatfix}   %
\usepackage{algorithm}
\usepackage{algpseudocode}
\usepackage{enumitem}
\usepackage{float}         %
\makeatletter
\newcommand\FloatBarrier{\par\begingroup\let\@elt\relax\edef\@tempa{\@botlist\@deferlist\@dbldeferlist}\def\@tempb{}\ifx\@tempa\@tempb\else\clearpage\fi\endgroup}
\makeatother

\newcommand{\ours}{CLOSE\xspace}
\newcommand{\M}{\mathcal{M}}
\newcommand{\pyes}{P(\text{yes})}

\newcommand{\sig}{$^{*}$}
\DeclareMathOperator{\LR}{LR}

\usepackage{multicol}

\definecolor{cvprblue}{rgb}{0.21,0.49,0.74}
\usepackage[breaklinks,colorlinks,allcolors=cvprblue]{hyperref}
\newcommand{\bref}[1]{}

\title{CLOSE: A Calibrated Label-free Online Self-Evaluating Verifier\\for Object Hallucination in Vision-Language Models}

\author{Zhaoxi Wang \qquad Jiahao Lai\\
Monash University\\
{\tt\small \{zwan0594, jlai0072\}@student.monash.edu}}

\begin{document}
\maketitle
\begin{abstract}
Vision-language models frequently describe objects absent from the image because of language priors.
Yet the model itself is already a strong judge of its own output: asked separately whether an object is present, it rejects 44\%--62\% of its hallucinations, and its yes probability reaches an AUROC of 0.90--0.95 without training.
This judgment, however, is not directly usable: (1) its probabilities lack a common scale across models; (2) for the remaining 38\%--56\%, the model still answers yes with probabilities that overlap those of real objects, so no threshold separates them.
We therefore propose \ours, a Calibrated Label-free Online Self-Evaluating verifier that enforces this judgment during generation.
Its threshold is a quantile of a memory rather than a probability and thus depends only on ranks; the memory is cold-started from unlabeled images and updated online.
A scorer self-trained on pseudo-labels from four hidden-state signals is fused with self-asking by rank to separate what self-asking leaves mixed.
Refused objects are rolled back and banned through a laddered back-off, and a recognition table intercepts their reappearance under other wordings.
On COCO, refusing 20\% and 40\% of object mentions reduces the CHAIRs of LLaVA-1.5-7B by 42\% and 87\% (F1 +1.9 and $-$6.1 points); refusing 20\% and 35\% reduces the reachable hallucinations of Qwen3-VL-8B (official counts minus CHAIR-script miscounts) by 79\% and 96\% (F1 +0.5 and $-$5.3).
An external vocabulary further extends the check beyond the official vocabulary.
\end{abstract}

\section{Introduction}
\label{sec:intro}

\begin{figure}[t]
  \centering
  \includegraphics[width=\columnwidth]{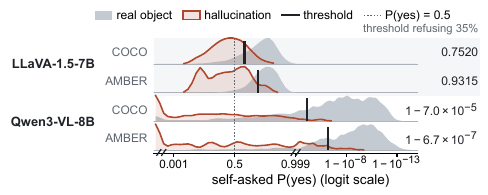}
  \caption{\textbf{Self-asking has no common scale across models} (development set, original captions, official annotations). $P(\text{yes})$ of real objects (grey) and hallucinations (red); black: threshold refusing 35\% of mentions (value at right); dotted: $P(\text{yes}){=}0.5$, right of which hallucinations are still affirmed; the axis is compressed 2$\times$ beyond 0.999 and values below $10^{-4}$ are drawn at the left edge (breaks).}
  \label{fig:scale}
\end{figure}

\begin{figure*}[t]
  \centering
  \includegraphics[width=0.9\textwidth]{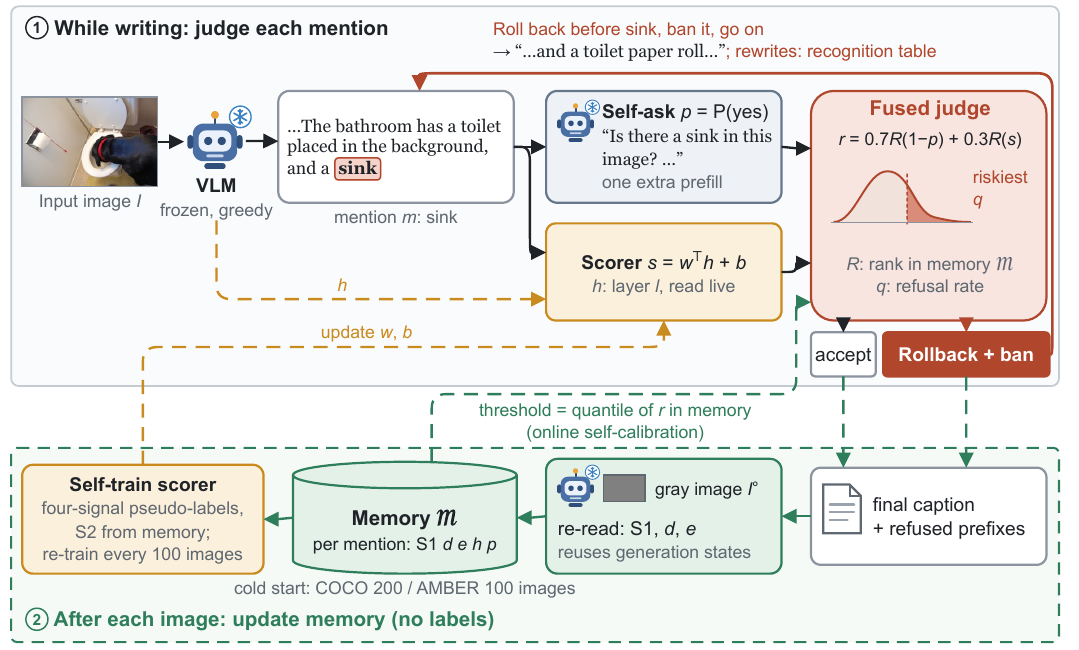}
  \caption{\textbf{Pipeline} (LLaVA-1.5-7B, $q{=}20\%$). Top: per-mention decision, rollback, ban and recognition table. Bottom (dashed): after each image, all mentions enter the memory. Alignment with existing verifiers: Fig.~\ref{fig:F1}.}
  \label{fig:method}
\end{figure*}

Object hallucination remains a major obstacle to deploying vision-language models (VLMs) in trustworthy applications~\cite{rohrbach2018object,li2023evaluating}.
Mainstream training-free methods gate and intervene in the latent space or on attention~\cite{liu2024paying,li2025the,liu2025reducing,wu2026rethinking,zou2026adaptive,zou2025look}\bref{liu2025reducing}; on long LLaVA-1.5-7B captions, they reduce CHAIRs by 13\%--67\% in the latent space~\cite{wu2026rethinking,zou2026adaptive} and by 8\%--83\% on attention~\cite{liu2024paying}, yet some hallucinations remain, and strong settings cause repetition and hurt fluency~\cite{zhong2026adaiat,zohrabi2026haloprobe}.

Methods that train a probe to judge whether an object is hallucinated~\cite{yang2026hallucinations,zohrabi2026haloprobe,park2025halloc,kogilathota2026halp} usually require labels, from a few thousand annotated mentions~\cite{yang2026hallucinations,zohrabi2026haloprobe} to over a hundred thousand samples~\cite{park2025halloc}, whereas real applications mostly involve unlabeled data, new datasets, and vocabularies other than the official CHAIR one.

Latent-space gates and probes alike make one judgment: is this object in the image?
Is there a more effective and purer way to make this judgment?
We find that the most direct way is to let the model ask itself this question.

\paragraph{Finding 1: the model itself is a strong judge.}
For each object in a caption we show the model only the image and ask ``Is there a \textit{X} in this image?''.
The model denies 44\%--62\% of its own hallucinations, and its yes probability separates hallucinations from real objects with an AUROC of 0.90--0.95 on LLaVA-1.5-7B and Qwen3-VL-8B, on COCO and AMBER (Sec.~\ref{sec:D.4}). This agrees with the observation that models discriminate better than they generate, which prior work uses only for evaluation or training~\cite{chen2026online,kadavath2022language,li2023evaluating}; we exploit it directly during generation.

\paragraph{Finding 2: the judgment is not directly usable.}
(1)~It has no common scale across models: to refuse the same 35\% of mentions, the probability threshold is 0.75 on LLaVA-1.5-7B but 0.9999 on Qwen3-VL-8B (\cref{fig:scale}), so no unified threshold can be set.
(2)~For the remaining 38\%--56\% of its hallucinations, the model still answers yes when asked separately; their probabilities overlap with those of real objects, so no threshold separates them.

We therefore propose \ours, a calibrated, label-free, online self-evaluating verifier that executes the self-asked judgment during generation (\cref{fig:method}).
On the same LLaVA-1.5-7B long captions on COCO, CHAIRs drops by 42\%--93\% as the refused share rises from 20\% to 50\%; for Qwen3-VL-8B, reachable hallucinations (official counts minus CHAIR-script miscounts) fall by 79\% when refusing 20\% and from 313 to 14 when refusing 35\%.

\paragraph{Contributions.}
(1) \textbf{An online self-calibrating fused judge.} Self-asking and a linear probe on the hidden states, the \emph{scorer}, are fused by rank; the scorer raises the risk rank of hallucinations the model itself affirms, correcting the blind spot of self-asking. The threshold is a quantile updated online with the test stream, so the probability scale of self-asking no longer affects the decision. To this end, we maintain a test-stream memory that stores the self-asked probability, hidden state and signals of every mention and supplies the ranks, the threshold and the scorer's training data.
(2)~\textbf{A label-free, small-data scorer.} The scorer is self-trained only from four hidden-state signals; its cold start needs only 100--200 unlabeled images. Because its training needs no annotation, the check is not limited to the official vocabulary: substituting an external vocabulary suffices to catch hallucinated objects outside the official vocabulary, and the method applies directly to new, unannotated data.
(3)~\textbf{Executing the self-asked decision during generation.} A refused object is rolled back under a laddered ban, and the recognition table catches it when the model writes it back in other words. Without any training, the quantile threshold, the rollback with laddered ban, and the recognition table already enforce plain self-asking.

\section{Related work}
\label{sec:related}

\begin{figure*}[t]
  \centering
  \includegraphics[width=\textwidth]{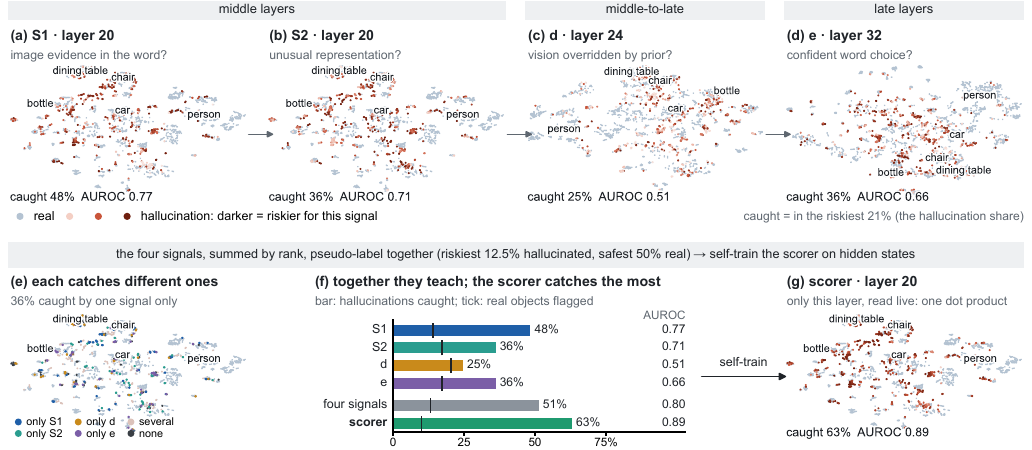}
  \caption{\textbf{Four signals, each covering one stage, complementarily teach the scorer} (Qwen3-VL-8B, COCO development set). Points: mentions on a t-SNE of each component's layer; darker hallucinations are riskier; ``caught'': among the riskiest 21\% (the hallucination share).}
  \label{fig:signal}
\end{figure*}

\paragraph{Training-free mitigation.}
Contrastive decoding (VCD \cite{leng2024mitigating}, M3ID \cite{favero2024multimodal}) compares outputs under original and perturbed inputs, and latent-space and attention interventions~\cite{liu2024paying,li2025the,liu2025reducing,wu2026rethinking,zou2026adaptive,zou2025look}\bref{liu2025reducing} such as VISTA~\cite{li2025the} and PAI~\cite{liu2024paying} restore visual information at inference.
These methods modulate decoding as a whole, and gated ones use internal signals only to decide when and how strongly to modulate; none decides to keep or remove individual objects, whereas we decide once at each object mention.

\paragraph{Origins of the four signals.}
Each of the four signals derives from one existing label-free cue (\cref{fig:signal}).
S1 follows image-replacement contrasts: the output change under a replaced image shows how much a word relies on the image~\cite{leng2024mitigating,favero2024multimodal,sun2025exploring}.
S2 follows representation-based anomaly detection: nearest-neighbor distance to past samples exposes unusual representations~\cite{sun2022outofdistribution}.
$d$ follows the finding that visual information fades over generation~\cite{favero2024multimodal,li2025the,sun2025exploring} and is written into text positions mainly in the middle layers~\cite{wang2025mllm,zhang2025crossmodal,jiang2025devils}\bref{zhang2025crossmodal}; VISTA~\cite{li2025the} steers hidden states on this basis.
$e$ follows the logit lens, whose intermediate-layer word distributions reflect confidence in the word choice~\cite{belrose2023eliciting,chuang2024dola}.
These signals do not intervene in decoding; they only provide pseudo-labels for the scorer.
Training-free internal-signal detectors (VisER~\cite{hasanebrahimi2026viser} and those evaluated in~\cite{chen2026robustness}) also need no labels, but report only AUROC and are not applied during generation.

\paragraph{Self-asking and self-verification.}
Asking separately dates back to P(True)~\cite{kadavath2022language} and POPE-style probing~\cite{li2023evaluating}.
IntroConformal~\cite{atabuzzaman2026introconformal} also reads yes/no logits and filters captions post hoc sentence by sentence, with a threshold set on a labeled calibration set; OSCAR~\cite{chen2026online} shows that models discriminate better than they generate, but uses self-asking only as a training reward; CoVe~\cite{dhuliawala2024chainofverification}, Self-Val~\cite{liu2026countering}, R-CoV~\cite{xie2026rcov} and LogicCheckGPT~\cite{wu2024logical} verify after generation and then rewrite or remove objects, without checking that the rewrite follows the verification.

\paragraph{Probe- and training-based verifiers.}
Most hallucination detectors are hidden-state probes trained with labels~\cite{samuel2026uniprobe,zohrabi2026haloprobe,park2025halloc,kogilathota2026halp,wu2026truthlens}.
VGD~\cite{yang2026hallucinations} shares our check-and-rollback skeleton, but trains its verifier on more than 4{,}000 labeled mentions and calibrates its threshold to a labeled recall target, and is not updated after deployment; REVERSE~\cite{wu2025generate} instead fine-tunes the model itself and backtracks and resamples when it detects a hallucination.
Most threshold-setting methods also rely on labels: conformal methods~\cite{atabuzzaman2026introconformal,angelopoulos2021a,angelopoulos2024conformal,gibbs2021adaptive} need labeled calibration data or label feedback, BICR~\cite{khanmohammadi2026grounded} needs correctness labels and does not intervene in generation, and CORAL~\cite{liu2026mitigating} sets a per-image threshold without accumulating across images.
HaloScope~\cite{du2024haloscope} and TSV~\cite{park2025steer} find truthfulness directions without labels and with a few labeled exemplars, respectively, but both are trained offline on text-only models.
After a rollback, banning words forces the model to paraphrase~\cite{candussio2026dont}, producing hallucinations that fixed-vocabulary metrics miss~\cite{al2024aloha}, which motivates our recognition table.

\section{Analysis}
\label{sec:analysis}

\begin{figure}[t]
  \centering
  \includegraphics[width=\columnwidth]{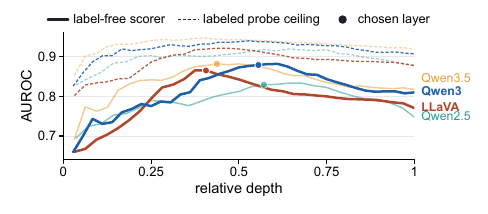}
  \caption{\textbf{The scorer reads the middle layers} (development set, COCO and AMBER averaged; Qwen3, Qwen3.5, Qwen2.5: Qwen3-VL-8B, Qwen3.5-9B, Qwen2.5-VL-7B). Solid: label-free scorer; dashed: labeled linear-probe ceiling; dots: chosen layers.}
  \label{fig:layer}
\end{figure}

\paragraph{Analysis 1: hallucinations are readable from hidden states, best in the middle layers.}
Across layers, the scorer (linear in the hidden state at a mention's first token) and a labeled probe ceiling both peak at a relative depth of about 0.4--0.6 (\cref{fig:layer}), an inverted U on every backbone, consistent with reports that the middle layers carry object-level visual information~\cite{wang2025mllm,zhang2025crossmodal,jiang2025devils}\bref{zhang2025crossmodal}.
The layers are chosen once on the labeled development set (1{,}000 COCO and 160 AMBER images, official annotations).
Trained only on pseudo-labels from four signals, the scorer attains 90\%--94\% of the labeled probe ceiling at the chosen layers.

\paragraph{Analysis 2: four signals, each covering one stage, are complementary teachers.}
Along the path from image to word (\cref{fig:signal}), S1 checks whether image evidence is written into the word, S2 whether the representation is unusual, $d$ whether the language prior overrides vision, and $e$ the confidence of the word choice. They catch different hallucinations: on Qwen3-VL-8B, 36\% are caught by only one signal (\cref{fig:signal}e).
No signal is strong on all backbones: on the COCO development set, $d$ is strong and S2 weak on LLaVA-1.5-7B (AUROC 0.72 and 0.53), whereas the reverse holds on Qwen3-VL-8B (0.51 and 0.71).

\begin{figure}[t]
  \centering
  \includegraphics[width=\columnwidth]{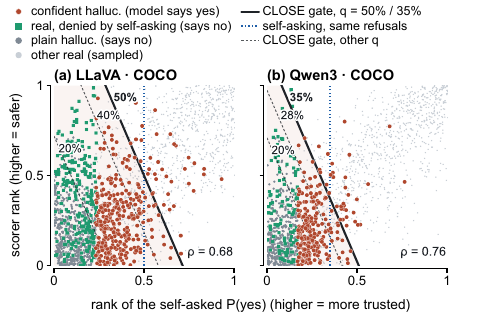}
  \caption{\textbf{Self-asking and the scorer make different errors} (COCO development set). Red: confident hallucinations, passed by self-asking and mostly ranked high-risk (low rank) by the scorer. Lines: \ours gate (bold: $q{=}50\%$ / $35\%$, lower left refused; dashed: other $q$) and plain self-asking (dotted).}
  \label{fig:ask}
\end{figure}

\begin{figure}[t]
  \centering
  \includegraphics[width=\columnwidth]{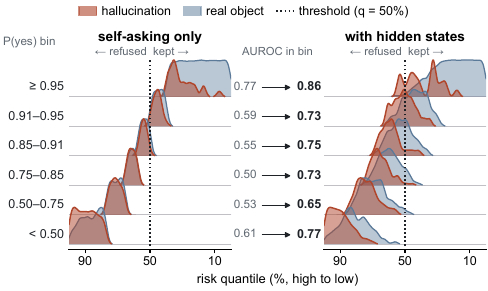}
  \caption{\textbf{Within each self-asked $\pyes$ bin (row), hidden states separate hallucinations from real objects} (LLaVA-1.5-7B, COCO development set, official annotations).}
  \label{fig:stage}
\end{figure}

\paragraph{Analysis 3: hidden states correct self-asking.}
On the original captions, plain self-asking reaches an AUROC of 0.90--0.95, and fused with the scorer 0.92--0.96 (Tab.~\ref{tab:D1}).
The two make different errors: across the eight development groups, the scorer ranks 68\%--82\% of the confident hallucinations among the riskiest 35\%, but only 17\%--29\% of the real objects the model also affirms, and its rank correlation with self-asking is only 0.46--0.76 (\cref{fig:ask}; Sec.~\ref{sec:D.1}). VGD~\cite{yang2026hallucinations} similarly finds, with labeled attention features, that confident hallucinations remain identifiable from internal signals.
\Cref{fig:stage} bins mentions by the self-asked $\pyes$: within a bin, self-asking barely separates hallucinations from real objects, with a within-bin AUROC of only 0.50--0.77, whereas adding hidden states yields 0.65--0.86.
The correction is targeted: the changed decisions are only 4.5\%--10.6\% of the mentions and lie right at the self-asking threshold (Fig.~\ref{fig:D3}); at the same refusal budget, the four backbones together admit a net 22 fewer hallucinations on COCO (95\% CI [6, 39]) and 8 fewer on AMBER (not significant, Fig.~\ref{fig:D4}(b)).


\section{Method}
\label{sec:method}

\ours consists of a gate and an intervention (\cref{fig:method}).
The \textbf{memory} $\M$ stores the signals, hidden states and self-asked probabilities of processed mentions, including refused ones, and is initialized from unlabeled images at cold start. $R(\cdot)$ denotes the rank within the memory divided by its size.

\subsection{Scorer: trained post hoc, read while writing}
\label{sec:head}

\paragraph{Four hidden-state signals} are computed after each image from the states of image $I$ already recorded while writing; only S1 needs one extra teacher-forced re-read with a uniform gray image $I^\circ$ of the same size.
Let $h^{(l)}_t$ be the hidden state of layer $l$ at position $t$, $t$ the first token of the mention and $t_0$ the position before the caption starts.
The signals are \textbf{S1 (image dependence)} and \textbf{S2 (novelty)} at a middle layer, our \textbf{$d$ (visual drift)} at a middle-to-late layer, and \textbf{$e$ (hesitation)} at a late layer:
\begin{gather}
S_1 = 1-\cos\big(h^{(l_1)}_t(I),\,h^{(l_1)}_t(I^\circ)\big),\nonumber\\
S_2 = -\tfrac{1}{10}\textstyle\sum_{j\in\mathcal N_{10}(m)}\cos\big(h^{(l_2)}_t,h_j\big),\nonumber\\
d = v\big(h^{(l_d)}_{t_0}\big)-v\big(h^{(l_d)}_{t-1}\big),\quad v(z)=\lVert B^\top z\rVert^2/\lVert z\rVert^2,\\
e = H\big(\mathrm{softmax}(W_U\,\mathrm{LN}(h^{(l_e)}_{t-1}))\big)/\log|V|,\nonumber
\end{gather}
where $\mathcal N_{10}(m)$ are the 10 nearest mentions of other images in memory by cosine, $B$ holds the top-6 principal directions of this image's image-token states, $W_U$ is the output layer, $\mathrm{LN}$ the final layer norm and $H$ the entropy.
Hallucinations have smaller S1 and larger S2, $d$ and $e$.

\paragraph{Pseudo-labels and self-training.}
The start evidence $z^0(m)=R\big(R(-S_1)+R(S_2)+R(d)+R(e)\big)$ sums the ranks of the four signals with equal weights.
The top $\beta{=}12.5\%$ of memory by $z^0$ are labeled hallucinated, the bottom $\gamma{=}50\%$ real, and the rest are excluded from training.
The scorer is linear in the standardized hidden state, $s(m)=w^\top x_m+b$, trained for $J$ rounds~\cite{xie2020selftraining}:
\begin{gather}
f_0=\LR_C\big(x,\tilde y(z^0)\big),\nonumber\\
f_j=\LR_C\Big(x,\tilde y\big(a\,R(f_{j-1}(x))+(1-a)\,z^0\big)\Big),
\end{gather}
where $\LR_C$ is class-balanced L2 logistic regression and the anchor $a$ retains a weight $1-a$ on the start evidence.
All models and datasets share $C{=}0.01$, $a{=}0.8$, $J{=}10$, set on development data.

\subsection{Quantile-fused judge}
\label{sec:judge}

\paragraph{Self-asking.}
When an object category first appears in a caption, a branch on the image-prefix KV cache receives ``Is there a \textit{X} in this image?'' and we read the yes probability
\begin{equation}
p(m)=\sigma\Big(\log\textstyle\sum_{y\in\mathcal Y}e^{\ell_y}-\log\sum_{n\in\mathcal N}e^{\ell_n}\Big),
\end{equation}
where $\ell$ are the output logits at the end of the question and $\mathcal Y,\mathcal N$ the spellings of yes and no.
The question omits the written caption (adding it lowers AUROC by 0.006--0.027, Sec.~\ref{sec:D.4}). On Qwen3-VL-8B, $p$ saturates into ties at 1 in single precision, so it is read in double precision (Tab.~\ref{tab:C2}).

With each score ranked by $R$ among values of the same type in $\M$, the fused risk and the decision are
\begin{gather}
r(m)=\lambda\,R\big(1-p(m)\big)+(1-\lambda)\,R\big(s(m)\big),\\
\begin{gathered}\text{refuse } m \iff r(m)>\tau_q,\\ \tau_q=Q_{1-q}\big(\{r(m')\}_{m'\in\M}\big),\end{gathered}
\end{gather}
with $\lambda=0.7$ and $Q_{1-q}$ the $(1-q)$ quantile.
The rate $q$ is the only operating parameter, chosen by the acceptable amount of deletion.

\subsection{Memory and online calibration}
\label{sec:memory}

After each image, the mentions of the final caption and every refused mention are stored with their signals, hidden states and $p$. The threshold is updated every 10 images and the scorer is re-trained every 100 images, always from earlier images only.
A cold start of 100--200 images suffices.
Re-training lets the scorer keep learning features underrepresented in the cold start (\cref{sec:components}).

If the fused risks of the $n$ memory mentions and a new mention are exchangeable, refusing above the $(1-q)$ quantile gives $q-\frac{1}{n+1}\le\Pr(\text{refuse})\le q+\frac{1}{n+1}$~\cite{angelopoulos2021a}.
In 18 test runs, the refusal rate remains within 3.3 points of the target $q$ (Tab.~\ref{tab:D2}).

\subsection{Intervention: laddered ban and recognition \mbox{table}}
\label{sec:intervention}

When a mention is refused, the KV cache is truncated before its first token, all official spellings of the refused category are banned as whole-word token sequences~\cite{wolf2020huggingfaces}, and greedy decoding resumes from that point.
In this laddered ban, the rollback grows with the number $k$ of catches in the current sentence: it returns to the mention $t$ for $k{=}1$ and to the start $t_{\text{NP}}$ of the mention's noun phrase for $2\le k\le 4$ (also banning the original first token there); for $k\ge 5$ it ends the sentence before the refused part if a full sentence remains, and otherwise deletes it.
A rewritten sentence must pass a rule-based fluency check; otherwise it backs off from the first faulty word. After 20 rollbacks per image, a further refusal deletes its sentence and ends the caption, except in the first sentence.

After a laddered ban, models often disguise the same object or write it back in other words (misspellings, split words, synonyms, hyponyms).
The \textbf{recognition table}, a code table derived from WordNet by fixed steps and revised through an error analysis on the development set, catches these forms. Each written form is normalized and looked up by its one-hot code in the recognition table; a form of a refused object triggers the same rollback and ban.

\section{Experiments}
\label{sec:experiments}

\begin{table*}[t]
\centering
\footnotesize
\caption{\textbf{End-to-end results} (greedy, at most 512 new tokens). Bold: best per column within each backbone. Tok.: mean number of output tokens per caption. Self-ask / scorer: one branch only. Parentheses: reachable. CHAIR, Cover, Hal, Cog and F1 in \%. Wall: same-machine wall-clock vs.\ the original.}
\label{tab:main}
\setlength{\tabcolsep}{4pt}
\begin{tabular}{llrrrcccccr}
\toprule
COCO 500 & Group & Real & Lost real & Halluc.\ (reach.) & CHAIRs / CHAIRi (reach.) & Cover & F1 & Wall & Tok. \\
\midrule
LLaVA-1.5-7B & Original  & 1232 & 0 & 431 (422) & 50.0 / 14.0 (49.4 / 13.7) & 80.1 & 82.9 & 1.00$\times$ & 117.9 \\
 & Scorer $q{=}20\%$  & 1151 & 90 & 223 (216) & 35.0 / 9.3 (34.4 / 9.1) & 74.8 & 82.0 & \textbf{1.06$\times$} & 112.5 \\
 & Self-ask $q{=}20\%$  & 1176 & 76 & 210 (204) & 30.6 / 7.8 (29.8 / 7.6) & 76.4 & 83.6 & 1.14$\times$ & 114.3 \\
 & \textbf{Ours $q{=}20\%$}  & \textbf{1198} & \textbf{50} & 183 (177) & 29.2 / 6.9 (28.2 / 6.7) & \textbf{77.8} & \textbf{84.8} & 1.18$\times$ & 113.2 \\
 & Ours $q{=}30\%$  & 1103 & 144 & 92 (86) & 16.2 / 4.3 (15.2 / 4.1) & 71.7 & 82.0 & 1.19$\times$ & 109.3 \\
 & \textbf{Ours $q{=}40\%$}  & 970 & 276 & 35 (29) & 6.6 / 1.8 (5.4 / 1.6) & 63.0 & 76.8 & 1.20$\times$ & 102.0 \\
 & Scorer $q{=}50\%$  & 800 & 432 & 35 (29) & 7.0 / 2.3 (5.8 / 2.1) & 52.0 & 67.8 & 1.13$\times$ & 93.3 \\
 & Self-ask $q{=}50\%$  & 821 & 425 & 23 (17) & 4.4 / 1.2 (3.2 / 0.9) & 53.3 & 69.3 & 1.25$\times$ & 96.6 \\
 & \textbf{Ours $q{=}50\%$}  & 826 & 412 & \textbf{19 (13)} & \textbf{3.6 / 0.9 (2.4 / 0.6)} & 53.7 & 69.6 & 1.20$\times$ & 95.7 \\
\midrule
Qwen3-VL-8B & Original  & 1212 & 0 & 440 (313) & 57.8 / 9.5 (46.0 / 7.2) & 78.8 & 84.2 & 1.00$\times$ & 402.3 \\
 & Scorer $q{=}20\%$  & 1128 & 89 & 250 (121) & 39.0 / 6.0 (19.8 / 3.3) & 73.3 & 82.4 & \textbf{1.06$\times$} & 394.7 \\
 & Self-ask $q{=}20\%$  & 1165 & 58 & 203 (84) & 33.8 / 4.7 (14.8 / 2.3) & 75.7 & 84.4 & 1.07$\times$ & 399.4 \\
 & \textbf{Ours $q{=}20\%$}  & \textbf{1173} & \textbf{47} & 192 (67) & 31.8 / 4.6 (12.4 / 1.9) & \textbf{76.2} & \textbf{84.7} & 1.07$\times$ & 400.4 \\
 & Ours $q{=}28\%$  & 1113 & 104 & 160 (38) & 27.8 / 3.9 (7.0 / 1.2) & 72.3 & 82.5 & 1.08$\times$ & 399.3 \\
 & Scorer $q{=}35\%$  & 987 & 227 & 158 (38) & 27.0 / 4.3 (6.8 / 1.5) & 64.1 & 76.8 & \textbf{1.06$\times$} & 383.3 \\
 & Self-ask $q{=}35\%$  & 1018 & 199 & 143 (21) & 25.4 / 3.6 (4.2 / 0.7) & 66.1 & 78.4 & 1.08$\times$ & 391.0 \\
 & \textbf{Ours $q{=}35\%$}  & 1027 & 189 & \textbf{137 (14)} & \textbf{24.4 / 3.4 (2.6 / 0.5)} & 66.7 & 78.9 & 1.07$\times$ & 391.6 \\
\midrule
Qwen2.5-VL-7B & Original  & 1083 & 0 & 214 (163) & 34.4 / 7.6 (26.6 / 6.1) & 70.4 & 79.9 & 1.00$\times$ & 187.6 \\
 & \textbf{Ours $q{=}35\%$}  & 843 & 246 & 49 (2) & 9.6 / 1.9 (0.4 / 0.1) & 54.8 & 70.3 & 1.16$\times$ & 186.8 \\
Qwen3.5-9B & Original  & 1224 & 0 & 491 (351) & 58.0 / 11.0 (46.4 / 8.6) & 79.5 & 84.0 & 1.00$\times$ & 436.8 \\
 & \textbf{Ours $q{=}35\%$}  & 1064 & 166 & 158 (21) & 27.6 / 3.7 (4.2 / 0.8) & 69.1 & 80.5 & 1.14$\times$ & 426.6 \\
\bottomrule
\end{tabular}

\vspace{4pt}
\begin{tabular}{llrrrccccccr}
\toprule
AMBER 1{,}004 & Group & Real & Lost real & Halluc. & CHAIR & Cover & Hal & Cog & F1 & Wall & Tok. \\
\midrule
LLaVA-1.5-7B & Original & 3334 & 0 & 498 & 7.7 & 51.7 & 36.0 & 4.2 & 66.3 & 1.00$\times$ & 97.4 \\
 & \textbf{Ours $q{=}50\%$} & 2105 & 1375 & 19 & 0.5 & 35.5 & 1.8 & 0.1 & 52.3 & 1.50$\times$ & 83.2 \\
Qwen3-VL-8B & Original & 5258 & 0 & 1034 & 7.6 & 74.5 & 60.1 & 3.5 & 82.5 & 1.00$\times$ & 296.1 \\
 & \textbf{Ours $q{=}35\%$} & 4106 & 1297 & 85 & 0.9 & 62.3 & 8.3 & 0.2 & 76.5 & 1.17$\times$ & 281.5 \\
\bottomrule
\end{tabular}
\end{table*}

\subsection{Setup}
\label{sec:setup}

\paragraph{Models and data.}
We use LLaVA-1.5-7B~\cite{liu2024improved} (fp16) and Qwen3-VL-8B~\cite{bai2025qwen3vl}, reported at $q{=}20$, $30$, $40$, $50\%$ and $20$, $28$, $35\%$, respectively. Qwen2.5-VL-7B~\cite{bai2025qwen25vl} and Qwen3.5-9B~\cite{qwen2026qwen35} run at $q{=}35\%$.
Decoding is greedy with at most 512 new tokens and the official prompts.
COCO uses 500 images sampled with seed 1994 from val2014 after removing the development and cold-start images; AMBER covers all 1{,}004 images, of which 260 serve as cold start and development, and the remaining 744 yield the same conclusions (Tab.~\ref{tab:E3}).
Labels are used only to select layers on the development set. The cold-start memory consists of 200 COCO val2017 and 100 AMBER images, disjoint from the development set.

\paragraph{Evaluation.}
Object mentions are identified with the official vocabularies of CHAIR~\cite{rohrbach2018object} or AMBER~\cite{wang2023amber}, counting only the first mention of each object per image.
The official CHAIR script matches words literally against a fixed vocabulary~\cite{kaul2024throne,al2024aloha,feng2024more} and counts some non-object words as objects. On COCO we therefore also report \textbf{reachable hallucinations}, the official ones minus three miscount types: the color word \emph{orange}, verb forms such as \emph{seated} and \emph{tied} (lemmatized to seat and tie), and objects in negations (\emph{no dog}; Sec.~\ref{sec:G.7}).
COCO F1 follows the standard definition (precision $1-$CHAIRi, recall = official Recall, i.e., Cover) and AMBER F1 is derived from CHAIR and Cover; \emph{lost real} counts real objects written by the original model but not by the method.
Differences are reported with 95\% intervals from 2{,}000 paired bootstrap resamples over images.

\subsection{Hallucinations drop sharply; the cost depends on the operating point}
\label{sec:main}

\begin{table*}[t]
\centering
\footnotesize
\caption{\textbf{Comparison with published methods} (same protocol; ours on their image list: Tab.~\ref{tab:G6}). Changes are relative to each paper's original model (orig.); COCO F1 of other methods as reported in their papers; parentheses: reachable, ours only; bold: best per column within each backbone; ``--'': not reported or not comparable.}
\label{tab:compare}
\setlength{\tabcolsep}{2.5pt}
\resizebox{\textwidth}{!}{%
\begin{tabular}{lrrrrrrrrr}
\toprule
Method & CHAIRs (reach.) $\downarrow$ & Change (reach.) & CHAIRi (reach.) $\downarrow$ & Change (reach.) & COCO F1 $\uparrow$ & AMBER CHAIR $\downarrow$ & Cover $\uparrow$ & Hal $\downarrow$ & AMBER F1 $\uparrow$ \\
\midrule
\multicolumn{10}{l}{\textit{LLaVA-1.5-7B}} \\
Original (unified table) & 50.0 & & 15.4 & & -- & 7.8 & 51.0 & 36.4 & 65.7 \\
REVERSE $\tau{=}0.003$ & 37.0 & $-$26\% & 10.3 & $-$33\% & -- & 6.0 & \textbf{52.2} & 30.4 & \textbf{67.1} \\
REVERSE $\tau{=}0.0003$ & 13.6 & $-$73\% & 6.1 & $-$60\% & -- & 4.0 & 26.9 & 10.2 & 42.0 \\
VGD @rec90 & 34.8 & $-$30\% & 9.7 & $-$37\% & -- & 4.4 & 50.8 & 19.1 & 66.3 \\
VGD @rec70 & 13.4 & $-$73\% & 4.6 & $-$70\% & -- & 2.8 & 38.2 & 8.2 & 54.8 \\
HaloProbe~\cite{zohrabi2026haloprobe} (orig.\ 51.6 / 15.2) & 17.6 & $-$66\% & 5.2 & $-$66\% & 75.2 & -- & -- & -- & -- \\
CAS~\cite{wu2026rethinking} $\beta{=}2$ (orig.\ 57.6 / 17.8) & 18.8 & $-$67\% & 9.2 & $-$48\% & -- & -- & -- & -- & -- \\
PAI~\cite{liu2024paying} $\alpha{=}0.4$ (orig.\ 46.2 / 13.8) & 42.4 & $-$8\% & 12.4 & $-$10\% & 76.9 & -- & -- & -- & -- \\
PAI~\cite{liu2024paying} $\alpha{=}0.6$ & 7.8 & $-$83\% & 4.8 & $-$65\% & 62.6 & -- & -- & -- & -- \\
Original (ours) & 50.0 (49.4) & & 14.0 (13.7) & & 82.9 & 7.7 & 51.7 & 36.0 & 66.3 \\
Ours $q{=}20\%$ & 29.2 (28.2) & $-$42\% & 6.9 (6.7) & $-$51\% & \textbf{84.8} & 2.2 & 50.3 & 9.6 & 66.4 \\
Ours $q{=}40\%$ & 6.6 (5.4) & $-$87\% & 1.8 (1.6) & $-$87\% & 76.8 & 0.9 & 41.8 & 3.0 & 58.8 \\
Ours $q{=}50\%$ & \textbf{3.6 (2.4)} & \textbf{$-$93\%} & \textbf{0.9 (0.6)} & \textbf{$-$94\%} & 69.6 & \textbf{0.5} & 35.5 & \textbf{1.8} & 52.3 \\
\midrule
\multicolumn{10}{l}{\textit{Qwen3-VL-8B}} \\
HaloProbe~\cite{zohrabi2026haloprobe} (orig.\ 24.8 / 8.0) & \textbf{12.2} & $-$51\% & 4.3 & $-$46\% & 74.6 & -- & -- & -- & -- \\
Original (ours) & 57.8 (46.0) & & 9.5 (7.2) & & 84.2 & 7.6 & 74.5 & 60.1 & 82.5 \\
Ours $q{=}20\%$ & 31.8 (12.4) & $-$45\% ($-$73\%) & 4.6 (1.9) & $-$52\% ($-$74\%) & \textbf{84.7} & 2.4 & \textbf{70.8} & 22.7 & \textbf{82.1} \\
Ours $q{=}28\%$ & 27.8 (7.0) & $-$52\% ($-$85\%) & 3.9 (1.2) & $-$59\% ($-$83\%) & 82.5 & 1.5 & 66.5 & 13.4 & 79.4 \\
Ours $q{=}35\%$ & 24.4 (2.6) & \textbf{$-$58\% ($-$94\%)} & \textbf{3.4 (0.5)} & \textbf{$-$64\% ($-$93\%)} & 78.9 & \textbf{0.9} & 62.3 & \textbf{8.3} & 76.5 \\
\bottomrule
\end{tabular}}
\end{table*}

\paragraph{Hallucinations and cost.}
At $q{=}20\%$, COCO reachable hallucinations fall by 58\% on LLaVA-1.5-7B and 79\% on Qwen3-VL-8B with only 2.8\% and 3.2\% fewer real objects (\cref{tab:main}); at $q{=}50\%$ / $35\%$ they fall from 422 to 13 and from 313 to 14, with 33\% and 15\% fewer real objects.
On AMBER, captions with a hallucination drop from 36.0\% to 1.8\% and from 60.1\% to 8.3\%, with 37\% and 22\% fewer real objects.
The self-ask rows involve no training: our quantile threshold and intervention alone reduce reachable hallucinations to 17 and 21 at $q{=}50\%$ / $35\%$, and the scorer lowers them further to 13 and 14. At $q{=}20\%$, the scorer removes 27 and 17 more reachable hallucinations on LLaVA-1.5-7B and Qwen3-VL-8B while losing 26 and 11 fewer real objects.
Wall-clock is 1.07--1.20$\times$ the original's on COCO and 1.14--1.50$\times$ on AMBER, the lowest (1.14$\times$) being Qwen3-VL-8B at $q{=}20\%$ (Tab.~\ref{tab:E2}).
The drop is not a length effect, although shorter captions also lower CHAIR~\cite{wang2023amber,kaul2024throne}: on COCO, the outputs of LLaVA-1.5-7B and Qwen3-VL-8B are only 4\% and 0.5\% shorter in tokens at $q{=}20\%$, and 19\% and 3\% at $q{=}50\%$ / $35\%$.

\paragraph{Miscounts and POPE.}
On Qwen3-VL-8B, 127 of the original's 440 official hallucinations and 123 of our 137 are CHAIR miscounts.
On POPE~\cite{li2023evaluating} and RePOPE~\cite{neuhaus2025repope}, yes/no answers name no object, so the judge never triggers and results equal the original's (Sec.~\ref{sec:E.2}).

\subsection{Label-free at deployment, with reductions comparable to labeled verifiers}
\label{sec:compare}

Published numbers come from the papers: REVERSE~\cite{wu2025generate} and VGD~\cite{yang2026hallucinations} from their unified-protocol tables (CHAIR 500 + AMBER), HaloProbe~\cite{zohrabi2026haloprobe} from its CHAIR results under the same protocol, PAI~\cite{liu2024paying} from its $\alpha$ ablation, and CAS~\cite{wu2026rethinking} from its layer ablation ($\beta{=}2$).
As image sets and original models differ, we compare relative reductions.
On LLaVA-1.5-7B, the two settings of REVERSE, VGD and PAI face ours at $q{=}40\%$ and $50\%$, with $q{=}20\%$ listed in addition; HaloProbe is the only method that also reports Qwen3-VL-8B. REVERSE fine-tunes the model, and VGD and HaloProbe train on labels (HaloProbe also with GPT-5 post-processing).

At similar coverage, our method hallucinates less than VGD: on AMBER with LLaVA-1.5-7B, ours at $q{=}20\%$ attains Cover 50.3, CHAIR 2.2 and Hal 9.6, versus 50.8, 4.4 and 19.1 for VGD @rec90 (\cref{tab:compare}).
On LLaVA-1.5-7B, ours at $q{=}50\%$ reduces CHAIRs / CHAIRi by 93\% / 94\%; on Qwen3-VL-8B at $q{=}35\%$, by 58\% / 64\% (94\% / 93\% reachable), more than HaloProbe (51\% / 46\%).

\subsection{What each component does}
\label{sec:components}
\begin{table}[!h]
\centering
\scriptsize
\caption{\textbf{From four signals to ours: detection AUROC} ($q{=}50\%$ / $35\%$). Each row adds one component; frozen: after the cold start; parentheses: change; $^{*}$: 95\% bootstrap interval excludes 0; bold: ours and ours $-$ plain.}
\label{tab:chain}
\setlength{\tabcolsep}{1.6pt}
\begin{tabular}{lcccc}
\toprule
 & \multicolumn{2}{c}{LLaVA-1.5-7B} & \multicolumn{2}{c}{Qwen3-VL-8B} \\
Component & COCO & AMBER & COCO & AMBER \\
\midrule
Four-signal start & .845 & .846 & .783 & .795 \\
+ scorer (frozen) & .901 (+.056$^{*}$) & .866 (+.020$^{*}$) & .905 (+.122$^{*}$) & .885 (+.090$^{*}$) \\
+ online re-train & .901 (.000) & .882 (+.016$^{*}$) & .918 (+.013$^{*}$) & .888 (+.003) \\
+ fuse self-ask (ours) & \textbf{.949 (+.048$^{*}$)} & \textbf{.947 (+.065$^{*}$)} & \textbf{.968 (+.050$^{*}$)} & \textbf{.949 (+.061$^{*}$)} \\
\midrule
Plain self-ask & .926 & .938 & .956 & .941 \\
Ours $-$ plain & \textbf{+.023$^{*}$} & \textbf{+.009$^{*}$} & \textbf{+.011$^{*}$} & \textbf{+.008$^{*}$} \\
\bottomrule
\end{tabular}
\end{table}

\begin{figure*}[t]
  \centering
  \includegraphics[width=\textwidth]{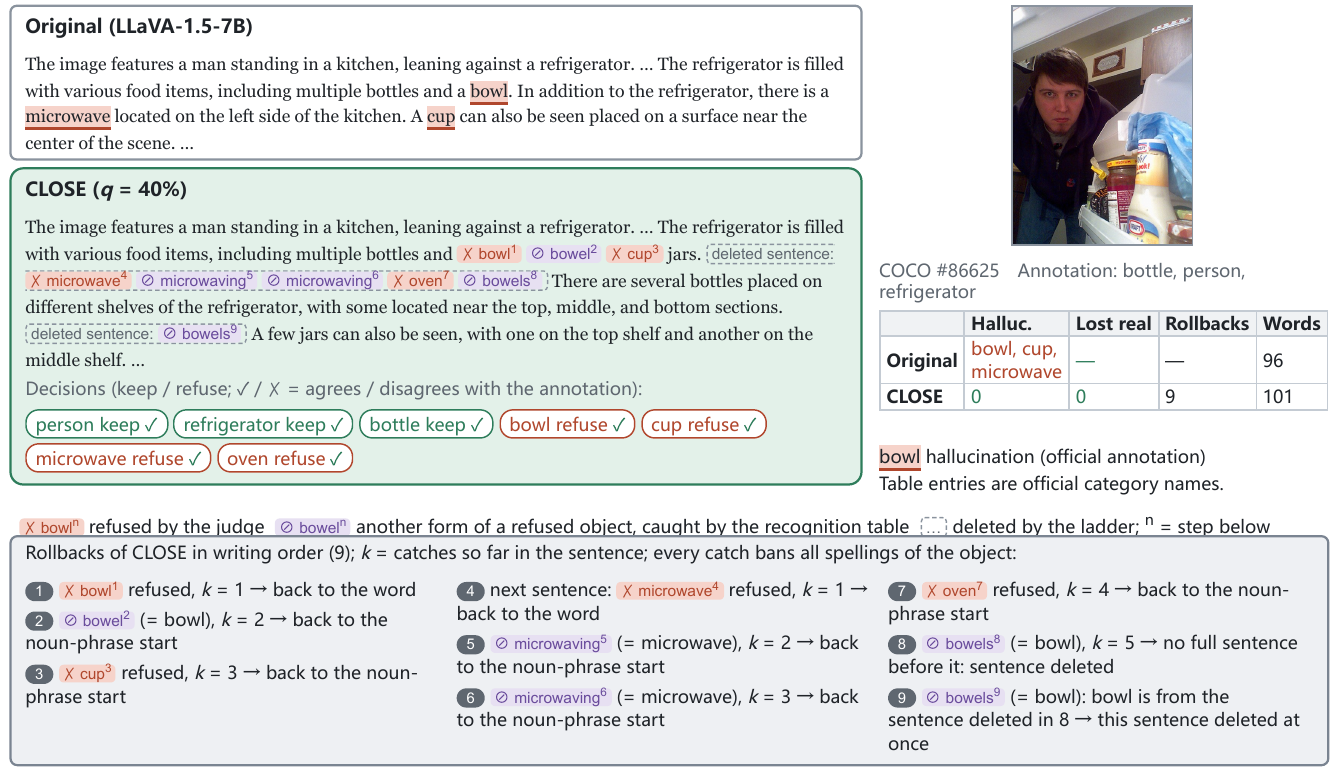}
  \caption{\textbf{An example} (LLaVA-1.5-7B, COCO, $q{=}40\%$). Left: the captions of the original model and of \ours (``\ldots'': unmarked sentences omitted); red highlight: hallucinations under the official annotation; crossed chips: mentions refused by the judge; purple chips: rewrites caught by the recognition table (\emph{bowel}, \emph{microwaving}, \emph{bowels}); dashed boxes: sentences deleted by the laddered ban; superscripts: the nine rollbacks listed at the bottom in writing order.}
  \label{fig:qual}
\end{figure*}

\paragraph{AUROC rises along the chain.}
The untrained four-signal start evidence reaches 0.78--0.85, and the scorer self-trained on its pseudo-labels improves it significantly in all four groups. Online re-training adds little to AUROC because a cold start of 100--200 images already largely suffices: with 50 images the scorer is 0.03--0.07 lower on COCO, and Qwen3-VL-8B reaches 0.92 only after about 300 images, the 200-image cold start plus 100 more (Tabs.~\ref{tab:D3} and~\ref{tab:D4}). Fusing with self-asking adds 0.048--0.065 and exceeds plain self-asking by 0.008--0.023 (\cref{tab:chain}).

\paragraph{S1 alone is insufficient, and the others are not redundant.}
As the sole teacher, S1 significantly lowers the fused AUROC in all eight groups, by 0.004--0.025; removing S1, S2, $d$ or $e$ alone lowers the fused AUROC significantly in 5, 4, 6 and 4 groups, and only removing S2 helps slightly in one (Tab.~\ref{tab:C1}).

\paragraph{The intervention enforces decisions at the cost of some real objects.}
With the judge fixed and only the intervention changed (development set, 100 images per group, Tab.~\ref{tab:G1}), backing off a single word leaves more hallucinations in all four groups (9 $\to$ 40 on LLaVA-1.5-7B COCO, 1--4 more in the other three) but retains 9--21 more real objects in three groups and one fewer on Qwen3-VL-8B COCO. Without the recognition table (\cref{fig:qual}), captions that write a refused object back rise from 0--3 to 4--26.

\subsection{Exploration: an external vocabulary}
\label{sec:external}

The judge checks only objects in the official vocabulary, and official metrics count only those, so out-of-vocabulary hallucinations escape both. We test an external vocabulary derived from WordNet nouns on 500 AMBER images unused in development (Qwen3-VL-8B, $q{=}35\%$, all nouns judged in two independent blind passes of Claude Opus 5.5; Tab.~\ref{tab:G3}). Checking only the official vocabulary removes all in-vocabulary hallucinations found by the blind judges, but the model rewrites some as out-of-vocabulary words, and out-of-vocabulary hallucinations rise from 153 to 210; with the external vocabulary, total hallucinations fall from 312 to 9 at a 24\% cost in real objects. On 100 MMBench~\cite{liu2024mmbench} scene images, with cold-start memory from AMBER, they fall from 77 to 3 (Tab.~\ref{tab:G4}). Combined, the external vocabulary and the recognition table leave the fewest captions that write a refused object back (Tab.~\ref{tab:G5}).

\section{Conclusion}
\label{sec:conclusion}

Without labels at deployment, the gate and the intervention execute the model's self-asked answer without training, and hidden states correct its ranking.
The model writes objects it denies when asked separately, in line with reports that open-ended generation relies more on language priors as it proceeds~\cite{chen2026online,favero2024multimodal,li2025the,sun2025exploring}; \ours asks the question at the moment the word is written and acts on the answer.

\paragraph{Limitations.}
\label{sec:limitations}
Quantile decisions control the refusal rate under exchangeability, not the error rate~\cite{angelopoulos2024conformal}; the same $q$ removes more real objects on backbones that hallucinate less, so $q$ is set per model.
The method only removes content and only addresses objects, not attribute or relation hallucinations.
The recognition table does not catch every rewrite: on development images it reduces the captions that write a refused object back by about 90\% (49 $\to$ 5, Tab.~\ref{tab:G1}), and a few still escape.

{
    \small
    \bibliographystyle{ieeenat_fullname_citeorder}
    \bibliography{main}

\begin{thebibliography}{66}
\providecommand{\natexlab}[1]{#1}
\providecommand{\url}[1]{\texttt{#1}}
\expandafter\ifx\csname urlstyle\endcsname\relax
  \providecommand{\doi}[1]{doi: #1}\else
  \providecommand{\doi}{doi: \begingroup \urlstyle{rm}\Url}\fi

\bibitem[Rohrbach et~al.(2018)Rohrbach, Hendricks, Burns, Darrell, and Saenko]{rohrbach2018object}
Anna Rohrbach, Lisa~Anne Hendricks, Kaylee Burns, Trevor Darrell, and Kate Saenko.
\newblock Object hallucination in image captioning.
\newblock In \emph{Proceedings of the 2018 Conference on Empirical Methods in Natural Language Processing}, pages 4035--4045, Brussels, Belgium, 2018. Association for Computational Linguistics.

\bibitem[Li et~al.(2023)Li, Du, Zhou, Wang, Zhao, and Wen]{li2023evaluating}
Yifan Li, Yifan Du, Kun Zhou, Jinpeng Wang, Wayne~Xin Zhao, and Ji-Rong Wen.
\newblock Evaluating object hallucination in large vision-language models.
\newblock In \emph{Proceedings of the 2023 Conference on Empirical Methods in Natural Language Processing}, pages 292--305, Singapore, 2023. Association for Computational Linguistics.

\bibitem[Liu et~al.(2024)Liu, Zheng, and Chen]{liu2024paying}
Shi Liu, Kecheng Zheng, and Wei Chen.
\newblock Paying more attention to image: A training-free method for alleviating hallucination in {LVLMs}.
\newblock In \emph{ECCV (83)}, pages 125--140, 2024.

\bibitem[Li et~al.(2025)Li, Shi, Gao, Liu, Wang, Chen, Liu, Zhao, Wang, and Metaxas]{li2025the}
Zhuowei Li, Haizhou Shi, Yunhe Gao, Di Liu, Zhenting Wang, Yuxiao Chen, Ting Liu, Long Zhao, Hao Wang, and Dimitris~N. Metaxas.
\newblock The hidden life of tokens: Reducing hallucination of large vision-language models via visual information steering.
\newblock In \emph{Forty-second International Conference on Machine Learning}, 2025.

\bibitem[Liu et~al.(2025)Liu, Ye, and Zou]{liu2025reducing}
Sheng Liu, Haotian Ye, and James Zou.
\newblock Reducing hallucinations in large vision-language models via latent space steering.
\newblock In \emph{The Thirteenth International Conference on Learning Representations}, 2025.

\bibitem[Wu et~al.(2026)Wu, Zhang, and Song]{wu2026rethinking}
Jingwen Wu, Xijun Zhang, and Ge Song.
\newblock Rethinking visual neglect: Steering via context-preference for {MLLM} hallucination mitigation.
\newblock \emph{arXiv preprint arXiv:2605.27993}, 2026.

\bibitem[Zou et~al.(2026)Zou, Gao, Guan, Li, and Marttinen]{zou2026adaptive}
Zhengtao Zou, Ya Gao, Jiarui Guan, Bin Li, and Pekka Marttinen.
\newblock Adaptive residual-update steering for low-overhead hallucination mitigation in large vision-language models.
\newblock In \emph{Forty-third International Conference on Machine Learning}, 2026.

\bibitem[Zou et~al.(2025)Zou, Wang, Yan, Lyu, Zheng, Huang, Chen, Jiang, Liu, Tang, and Hu]{zou2025look}
Xin Zou, Yizhou Wang, Yibo Yan, Yuanhuiyi Lyu, Kening Zheng, Sirui Huang, Junkai Chen, Peijie Jiang, Jia Liu, Chang Tang, and Xuming Hu.
\newblock Look twice before you answer: Memory-space visual retracing for hallucination mitigation in multimodal large language models.
\newblock In \emph{ICML}, 2025.

\bibitem[Zhong et~al.(2026)Zhong, He, Zheng, Li, Wang, and Kang]{zhong2026adaiat}
Li'an Zhong, Ziqiang He, Jibin Zheng, Jin Li, Z.~Jane Wang, and Xiangui Kang.
\newblock Ada{IAT}: Adaptively increasing attention to generated text to alleviate hallucinations in {LVLM}.
\newblock In \emph{Conference on Computer Vision and Pattern Recognition 2026}, 2026.

\bibitem[Zohrabi et~al.(2026)Zohrabi, Hasani, Gupta, Baghshah, Rohrbach, and Rohrbach]{zohrabi2026haloprobe}
Reihaneh Zohrabi, Hosein Hasani, Akshita Gupta, Mahdieh~Soleymani Baghshah, Anna Rohrbach, and Marcus Rohrbach.
\newblock {HaloProbe}: Bayesian detection and mitigation of object hallucinations in vision-language models.
\newblock In \emph{Forty-third International Conference on Machine Learning}, 2026.

\bibitem[Yang et~al.(2026)Yang, Liu, Wu, Wang, Zhu, Zhang, Hu, Wu, Fu, and Lin]{yang2026hallucinations}
Lei Yang, Xinze Liu, Dayan Wu, Ding Wang, Hengjie Zhu, Zihao Zhang, Tianzhu Hu, Hanqi Wu, Peng Fu, and Zheng Lin.
\newblock Hallucinations leave a grounding signature: Verifier-guided decoding for selective object correction.
\newblock \emph{arXiv preprint arXiv:2607.27823}, 2026.

\bibitem[Park et~al.(2025)Park, Kim, and Kim]{park2025halloc}
Eunkyu Park, Minyeong Kim, and Gunhee Kim.
\newblock {HalLoc}: Token-level localization of hallucinations for vision language models.
\newblock In \emph{CVPR}, pages 29893--29903, 2025.

\bibitem[Kogilathota et~al.(2026)Kogilathota, G, Sun, and Zhou]{kogilathota2026halp}
Sai~Akhil Kogilathota, Sripadha Vallabha~E G, Luzhe Sun, and Jiawei Zhou.
\newblock {HALP}: Detecting hallucinations in vision-language models without generating a single token.
\newblock In \emph{Proceedings of the 19th Conference of the {E}uropean Chapter of the {A}ssociation for {C}omputational {L}inguistics (Volume 1: Long Papers)}, pages 6067--6085, Rabat, Morocco, 2026. Association for Computational Linguistics.

\bibitem[Chen et~al.(2026)Chen, Yang, Zhu, Wu, Si, and Lin]{chen2026online}
Minghui Chen, Chenxu Yang, Hengjie Zhu, Dayan Wu, Qingyi Si, and Zheng Lin.
\newblock Online self-calibration against hallucination in vision-language models.
\newblock In \emph{IJCAI/ECAI}, pages 989--997, 2026.

\bibitem[Kadavath et~al.(2022)Kadavath, Conerly, Askell, Henighan, Drain, Perez, Schiefer, Hatfield-Dodds, DasSarma, Tran-Johnson, Johnston, Showk, Jones, Elhage, Hume, Chen, Bai, Bowman, Fort, Ganguli, Hernandez, Jacobson, Kernion, Kravec, Lovitt, Ndousse, Olsson, Ringer, Amodei, Brown, Clark, Joseph, Mann, McCandlish, Olah, and Kaplan]{kadavath2022language}
Saurav Kadavath, Tom Conerly, Amanda Askell, Tom Henighan, Dawn Drain, Ethan Perez, Nicholas Schiefer, Zac Hatfield-Dodds, Nova DasSarma, Eli Tran-Johnson, Scott Johnston, Sheer~El Showk, Andy Jones, Nelson Elhage, Tristan Hume, Anna Chen, Yuntao Bai, Sam Bowman, Stanislav Fort, Deep Ganguli, Danny Hernandez, Josh Jacobson, Jackson Kernion, Shauna Kravec, Liane Lovitt, Kamal Ndousse, Catherine Olsson, Sam Ringer, Dario Amodei, Tom Brown, Jack Clark, Nicholas Joseph, Ben Mann, Sam McCandlish, Chris Olah, and Jared Kaplan.
\newblock Language models (mostly) know what they know.
\newblock \emph{CoRR}, abs/2207.05221, 2022.

\bibitem[Leng et~al.(2024)Leng, Zhang, Chen, Li, Lu, Miao, and Bing]{leng2024mitigating}
Sicong Leng, Hang Zhang, Guanzheng Chen, Xin Li, Shijian Lu, Chunyan Miao, and Lidong Bing.
\newblock Mitigating object hallucinations in large vision-language models through visual contrastive decoding.
\newblock In \emph{CVPR}, pages 13872--13882, 2024.

\bibitem[Favero et~al.(2024)Favero, Zancato, Trager, Choudhary, Perera, Achille, Swaminathan, and Soatto]{favero2024multimodal}
Alessandro Favero, Luca Zancato, Matthew Trager, Siddharth Choudhary, Pramuditha Perera, Alessandro Achille, Ashwin Swaminathan, and Stefano Soatto.
\newblock Multi-modal hallucination control by visual information grounding.
\newblock In \emph{CVPR}, pages 14303--14312, 2024.

\bibitem[Sun et~al.(2025)Sun, Atarashi, Takeuchi, and Kashima]{sun2025exploring}
Yaqi Sun, Kyohei Atarashi, Koh Takeuchi, and Hisashi Kashima.
\newblock Exploring causes and mitigation of hallucinations in large vision language models.
\newblock \emph{CoRR}, abs/2502.16842, 2025.

\bibitem[Sun et~al.(2022)Sun, Ming, Zhu, and Li]{sun2022outofdistribution}
Yiyou Sun, Yifei Ming, Xiaojin Zhu, and Yixuan Li.
\newblock Out-of-distribution detection with deep nearest neighbors.
\newblock In \emph{Proceedings of the 39th International Conference on Machine Learning}, pages 20827--20840. PMLR, 2022.

\bibitem[Wang et~al.(2025)Wang, Chen, Zhang, Tian, Xu, Deng, and Chen]{wang2025mllm}
Chenxi Wang, Xiang Chen, Ningyu Zhang, Bozhong Tian, Haoming Xu, Shumin Deng, and Huajun Chen.
\newblock {MLLM} can see? dynamic correction decoding for hallucination mitigation.
\newblock In \emph{ICLR}, 2025.

\bibitem[Zhang et~al.(2025)Zhang, Yadav, Han, and Shutova]{zhang2025crossmodal}
Zhi Zhang, Srishti Yadav, Fengze Han, and Ekaterina Shutova.
\newblock Cross-modal information flow in multimodal large language models.
\newblock In \emph{2025 IEEE/CVF Conference on Computer Vision and Pattern Recognition (CVPR)}, pages 19781--19791. IEEE, 2025.

\bibitem[Jiang et~al.(2025)Jiang, Chen, Zhu, Luo, Shen, and Yang]{jiang2025devils}
Zhangqi Jiang, Junkai Chen, Beier Zhu, Tingjin Luo, Yankun Shen, and Xu Yang.
\newblock Devils in middle layers of large vision-language models: Interpreting, detecting and mitigating object hallucinations via attention lens.
\newblock In \emph{2025 IEEE/CVF Conference on Computer Vision and Pattern Recognition (CVPR)}, pages 25004--25014. IEEE, 2025.

\bibitem[Belrose et~al.(2023)Belrose, Ostrovsky, McKinney, Furman, Smith, Halawi, Biderman, and Steinhardt]{belrose2023eliciting}
Nora Belrose, Igor Ostrovsky, Lev McKinney, Zach Furman, Logan Smith, Danny Halawi, Stella Biderman, and Jacob Steinhardt.
\newblock Eliciting latent predictions from transformers with the tuned lens.
\newblock \emph{arXiv preprint arXiv:2303.08112}, 2023.

\bibitem[Chuang et~al.(2024)Chuang, Xie, Luo, Kim, Glass, and He]{chuang2024dola}
Yung-Sung Chuang, Yujia Xie, Hongyin Luo, Yoon Kim, James~R. Glass, and Pengcheng He.
\newblock {DoLa}: Decoding by contrasting layers improves factuality in large language models.
\newblock In \emph{ICLR}, 2024.

\bibitem[Hasanebrahimi et~al.(2026)Hasanebrahimi, Huang, Leckie, and Erfani]{hasanebrahimi2026viser}
Afsaneh Hasanebrahimi, Hanxun Huang, Christopher Leckie, and Sarah Erfani.
\newblock {V}is{ER}: Visual evidence and reliance for object hallucination detection in {LVLM}s.
\newblock In \emph{Proceedings of the 2026 Conference on Empirical Methods in Natural Language Processing}, pages 60186--60202, Budapest, Hungary, 2026. Association for Computational Linguistics.

\bibitem[Chen et~al.(2026)Chen, Yang, Magid, Williams, Russakovsky, and Tureci]{chen2026robustness}
Allison Chen, William Yang, Salma~Abdel Magid, Jonathan Williams, Olga Russakovsky, and Esin Tureci.
\newblock Robustness of internal signals for hallucination detection in vision-language models.
\newblock \emph{arXiv preprint arXiv:2610.27041}, 2026.

\bibitem[Atabuzzaman et~al.(2026)Atabuzzaman, Alexander, and Thomas]{atabuzzaman2026introconformal}
Md. Atabuzzaman, Christian Alexander, and Chris Thomas.
\newblock {I}ntro{C}onformal: Conformal factuality guarantees for large vision-language models via introspective signals.
\newblock In \emph{Proceedings of the 2026 Conference on Empirical Methods in Natural Language Processing}, pages 58747--58764, Budapest, Hungary, 2026. Association for Computational Linguistics.

\bibitem[Dhuliawala et~al.(2024)Dhuliawala, Komeili, Xu, Raileanu, Li, Celikyilmaz, and Weston]{dhuliawala2024chainofverification}
Shehzaad Dhuliawala, Mojtaba Komeili, Jing Xu, Roberta Raileanu, Xian Li, Asli Celikyilmaz, and Jason Weston.
\newblock Chain-of-verification reduces hallucination in large language models.
\newblock In \emph{Findings of the {A}ssociation for {C}omputational {L}inguistics: {ACL} 2024}, pages 3563--3578, Bangkok, Thailand, 2024. Association for Computational Linguistics.

\bibitem[Liu et~al.(2026)Liu, Wen, Lan, Wang, and Su]{liu2026countering}
Shiyu Liu, Xinyi Wen, Zhibin Lan, Ante Wang, and Jinsong Su.
\newblock Countering the over-reliance trap: Mitigating object hallucination for {LVLMs} via a self-validation framework.
\newblock \emph{arXiv preprint arXiv:2601.22451}, 2026.

\bibitem[Xie et~al.(2026)Xie, Tonioni, Rauschmayr, Tombari, and Schiele]{xie2026rcov}
Jiahao Xie, Alessio Tonioni, Nathalie Rauschmayr, Federico Tombari, and Bernt Schiele.
\newblock {R-CoV}: Region-aware chain-of-verification for alleviating object hallucinations in {LVLMs}.
\newblock \emph{arXiv preprint arXiv:2604.20696}, 2026.

\bibitem[Wu et~al.(2024)Wu, Liu, Wang, Zhang, Wu, Wang, and Tan]{wu2024logical}
Junfei Wu, Qiang Liu, Ding Wang, Jinghao Zhang, Shu Wu, Liang Wang, and Tieniu Tan.
\newblock Logical closed loop: Uncovering object hallucinations in large vision-language models.
\newblock In \emph{ACL (Findings)}, pages 6944--6962, 2024.

\bibitem[Samuel et~al.(2026)Samuel, Bar-Shalom, Frasca, Fetaya, Ziser, Chechik, and Maron]{samuel2026uniprobe}
Dvir Samuel, Guy Bar-Shalom, Fabrizio Frasca, Ethan Fetaya, Yftah Ziser, Gal Chechik, and Haggai Maron.
\newblock {UniProbe}: A learnable token-level hallucination detector for large {VLMs} using multi-structural internal representations.
\newblock \emph{arXiv preprint arXiv:2608.10835}, 2026.

\bibitem[Wu et~al.(2026)Wu, Lai, Lu, Chen, Zhou, Zhuang, Yu, and Wang]{wu2026truthlens}
Yanqi Wu, Runhe Lai, Xinhua Lu, Qichao Chen, Zhiping Zhou, Jia-Xin Zhuang, Weijiang Yu, and Ruixuan Wang.
\newblock {TruthLens}: Object hallucination detection via self-evaluating truthfulness scores in {LVLMs}.
\newblock In \emph{Computer Vision – ECCV 2026}, pages 537--554. Springer Nature Switzerland, 2026.

\bibitem[Wu et~al.(2025)Wu, Lee, Ge, Gonzalez, Darrell, and Chan]{wu2025generate}
Tsung-Han Wu, Heekyung Lee, Jiaxin Ge, Joseph~E. Gonzalez, Trevor Darrell, and David~M. Chan.
\newblock Generate, but verify: Reducing hallucination in vision-language models with retrospective resampling.
\newblock In \emph{The Thirty-ninth Annual Conference on Neural Information Processing Systems}, 2025.

\bibitem[Angelopoulos and Bates(2023)]{angelopoulos2021a}
Anastasios~N. Angelopoulos and Stephen Bates.
\newblock Conformal prediction: A gentle introduction.
\newblock \emph{Foundations and Trends in Machine Learning}, 16\penalty0 (4):\penalty0 494--591, 2023.

\bibitem[Angelopoulos et~al.(2024)Angelopoulos, Bates, Fisch, Lei, and Schuster]{angelopoulos2024conformal}
Anastasios~Nikolas Angelopoulos, Stephen Bates, Adam Fisch, Lihua Lei, and Tal Schuster.
\newblock Conformal risk control.
\newblock In \emph{ICLR}, 2024.

\bibitem[Gibbs and Candes(2021)]{gibbs2021adaptive}
Isaac Gibbs and Emmanuel Candes.
\newblock Adaptive conformal inference under distribution shift.
\newblock In \emph{Advances in Neural Information Processing Systems}, pages 1660--1672. Curran Associates, Inc., 2021.

\bibitem[Khanmohammadi et~al.(2026)Khanmohammadi, Miahi, Kaur, Smiley, Brugere, Thind, and Ghassemi]{khanmohammadi2026grounded}
Reza Khanmohammadi, Erfan Miahi, Simerjot Kaur, Charese Smiley, Ivan Brugere, Kundan~S Thind, and Mohammad~M. Ghassemi.
\newblock Grounded or guessing? {LVLM} confidence estimation via blind-image contrastive ranking.
\newblock In \emph{Proceedings of the 15th International Joint Conference on Natural Language Processing and the 5th Conference of the {A}sia-Pacific Chapter of the {A}ssociation for {C}omputational {L}inguistics (Volume 1: Long Papers)}, pages 4645--4721, Hengqin, China, 2026. Association for Computational Linguistics.

\bibitem[Liu et~al.(2026)Liu, Tian, and Xie]{liu2026mitigating}
Chang Liu, Yu Tian, and Rui Xie.
\newblock Mitigating object hallucination in large vision-language models via false discovery controlled visual data splitting.
\newblock \emph{arXiv preprint arXiv:2609.38979}, 2026.

\bibitem[Du et~al.(2024)Du, Xiao, and Li]{du2024haloscope}
Xuefeng Du, Chaowei Xiao, and Sharon Li.
\newblock {HaloScope}: Harnessing unlabeled {LLM} generations for hallucination detection.
\newblock In \emph{NeurIPS}, 2024.

\bibitem[Park et~al.(2025)Park, Du, Yeh, Wang, and Li]{park2025steer}
Seongheon Park, Xuefeng Du, Min-Hsuan Yeh, Haobo Wang, and Yixuan Li.
\newblock Steer {LLM} latents for hallucination detection.
\newblock In \emph{ICML}, 2025.

\bibitem[Candussio et~al.(2026)Candussio, Padovani, Scalena, and Nissim]{candussio2026dont}
Sara Candussio, Francesca Padovani, Daniel Scalena, and Malvina Nissim.
\newblock {``Don't Say It!''}: Constraints, compliance, and communication when language models play taboo.
\newblock \emph{arXiv preprint arXiv:2607.00601}, 2026.

\bibitem[Petryk et~al.(2024)Petryk, Chan, Kachinthaya, Zou, Canny, Gonzalez, and Darrell]{al2024aloha}
Suzanne Petryk, David~M. Chan, Anish Kachinthaya, Haodi Zou, John~F. Canny, Joseph Gonzalez, and Trevor Darrell.
\newblock {ALOHa}: A new measure for hallucination in captioning models.
\newblock In \emph{NAACL (Short Papers)}, pages 342--357, 2024.

\bibitem[Xie et~al.(2020)Xie, Luong, Hovy, and Le]{xie2020selftraining}
Qizhe Xie, Minh-Thang Luong, Eduard Hovy, and Quoc~V. Le.
\newblock Self-training with noisy student improves {ImageNet} classification.
\newblock In \emph{2020 IEEE/CVF Conference on Computer Vision and Pattern Recognition (CVPR)}, pages 10684--10695. IEEE, 2020.

\bibitem[Wolf et~al.(2020)Wolf, Debut, Sanh, Chaumond, Delangue, Moi, Cistac, Rault, Louf, Funtowicz, Davison, Shleifer, von Platen, Ma, Jernite, Plu, Xu, Le~Scao, Gugger, Drame, Lhoest, and Rush]{wolf2020huggingfaces}
Thomas Wolf, Lysandre Debut, Victor Sanh, Julien Chaumond, Clement Delangue, Anthony Moi, Pierric Cistac, Tim Rault, Remi Louf, Morgan Funtowicz, Joe Davison, Sam Shleifer, Patrick von Platen, Clara Ma, Yacine Jernite, Julien Plu, Canwen Xu, Teven Le~Scao, Sylvain Gugger, Mariama Drame, Quentin Lhoest, and Alexander Rush.
\newblock Transformers: State-of-the-art natural language processing.
\newblock In \emph{Proceedings of the 2020 Conference on Empirical Methods in Natural Language Processing: System Demonstrations}, pages 38--45, Online, 2020. Association for Computational Linguistics.

\bibitem[Liu et~al.(2024)Liu, Li, Li, and Lee]{liu2024improved}
Haotian Liu, Chunyuan Li, Yuheng Li, and Yong~Jae Lee.
\newblock Improved baselines with visual instruction tuning.
\newblock In \emph{2024 IEEE/CVF Conference on Computer Vision and Pattern Recognition (CVPR)}, pages 26286--26296. IEEE, 2024.

\bibitem[Bai et~al.(2025{\natexlab{a}})Bai, Cai, Chen, Chen, Chen, Cheng, Deng, Ding, Gao, Ge, Ge, Guo, Huang, Huang, Huang, Hui, Jiang, Li, Li, Li, Li, Lin, Lin, Liu, Liu, Liu, Liu, Liu, Liu, Lu, Luo, Lv, Men, Meng, Ren, Ren, Song, Sun, Tang, Tu, Wan, Wang, Wang, Wang, Wang, Xie, Xu, Xu, Xu, Yang, Yang, Yang, Yang, Yu, Zhang, Zhang, Zhang, Zheng, Zhong, Zhou, Zhou, Zhou, Zhu, and Zhu]{bai2025qwen3vl}
Shuai Bai, Yuxuan Cai, Ruizhe Chen, Keqin Chen, Xionghui Chen, Zesen Cheng, Lianghao Deng, Wei Ding, Chang Gao, Chunjiang Ge, Wenbin Ge, Zhifang Guo, Qidong Huang, Jie Huang, Fei Huang, Binyuan Hui, Shutong Jiang, Zhaohai Li, Mingsheng Li, Mei Li, Kaixin Li, Zicheng Lin, Junyang Lin, Xuejing Liu, Jiawei Liu, Chenglong Liu, Yang Liu, Dayiheng Liu, Shixuan Liu, Dunjie Lu, Ruilin Luo, Chenxu Lv, Rui Men, Lingchen Meng, Xuancheng Ren, Xingzhang Ren, Sibo Song, Yuchong Sun, Jun Tang, Jianhong Tu, Jianqiang Wan, Peng Wang, Pengfei Wang, Qiuyue Wang, Yuxuan Wang, Tianbao Xie, Yiheng Xu, Haiyang Xu, Jin Xu, Zhibo Yang, Mingkun Yang, Jianxin Yang, An Yang, Bowen Yu, Fei Zhang, Hang Zhang, Xi Zhang, Bo Zheng, Humen Zhong, Jingren Zhou, Fan Zhou, Jing Zhou, Yuanzhi Zhu, and Ke Zhu.
\newblock {Qwen3-VL} technical report.
\newblock \emph{CoRR}, abs/2511.21631, 2025{\natexlab{a}}.

\bibitem[Bai et~al.(2025{\natexlab{b}})Bai, Chen, Liu, Wang, Ge, Song, Dang, Wang, Wang, Tang, Zhong, Zhu, Yang, Li, Wan, Wang, Ding, Fu, Xu, Ye, Zhang, Xie, Cheng, Zhang, Yang, Xu, and Lin]{bai2025qwen25vl}
Shuai Bai, Keqin Chen, Xuejing Liu, Jialin Wang, Wenbin Ge, Sibo Song, Kai Dang, Peng Wang, Shijie Wang, Jun Tang, Humen Zhong, Yuanzhi Zhu, Ming-Hsuan Yang, Zhaohai Li, Jianqiang Wan, Pengfei Wang, Wei Ding, Zheren Fu, Yiheng Xu, Jiabo Ye, Xi Zhang, Tianbao Xie, Zesen Cheng, Hang Zhang, Zhibo Yang, Haiyang Xu, and Junyang Lin.
\newblock {Qwen2.5-VL} technical report.
\newblock \emph{CoRR}, abs/2502.13923, 2025{\natexlab{b}}.

\bibitem[{Qwen Team}(2026)]{qwen2026qwen35}
{Qwen Team}.
\newblock {Qwen3.5}: Towards native multimodal agents.
\newblock \url{https://qwen.ai/blog?id=qwen3.5}, 2026.

\bibitem[Wang et~al.(2023)Wang, Wang, Xu, Zhang, Gu, Jia, Wang, Xu, Yan, Zhang, and Sang]{wang2023amber}
Junyang Wang, Yuhang Wang, Guohai Xu, Jing Zhang, Yukai Gu, Haitao Jia, Jiaqi Wang, Haiyang Xu, Ming Yan, Ji Zhang, and Jitao Sang.
\newblock {AMBER}: An {LLM-free} multi-dimensional benchmark for {MLLMs} hallucination evaluation.
\newblock \emph{arXiv preprint arXiv:2311.07397}, 2023.

\bibitem[Kaul et~al.(2024)Kaul, Li, Yang, Dukler, Swaminathan, Taylor, and Soatto]{kaul2024throne}
Prannay Kaul, Zhizhong Li, Hao Yang, Yonatan Dukler, Ashwin Swaminathan, C.~J. Taylor, and Stefano Soatto.
\newblock {THRONE}: An object-based hallucination benchmark for the free-form generations of large vision-language models.
\newblock In \emph{CVPR}, pages 27218--27228, 2024.

\bibitem[Feng et~al.(2024)Feng, Tang, Zhang, and Xu]{feng2024more}
Mingqian Feng, Yunlong Tang, Zeliang Zhang, and Chenliang Xu.
\newblock Do more details always introduce more hallucinations in {LVLM-based} image captioning?
\newblock \emph{CoRR}, abs/2406.12663, 2024.

\bibitem[Neuhaus and Hein(2025)]{neuhaus2025repope}
Yannic Neuhaus and Matthias Hein.
\newblock {RePOPE}: Impact of annotation errors on the {POPE} benchmark.
\newblock \emph{arXiv preprint arXiv:2504.15707}, 2025.

\bibitem[Liu et~al.(2024)Liu, Duan, Zhang, Li, Zhang, Zhao, Yuan, Wang, He, Liu, Chen, and Lin]{liu2024mmbench}
Yuan Liu, Haodong Duan, Yuanhan Zhang, Bo Li, Songyang Zhang, Wangbo Zhao, Yike Yuan, Jiaqi Wang, Conghui He, Ziwei Liu, Kai Chen, and Dahua Lin.
\newblock {MMBench}: Is your multi-modal model an all-around player?
\newblock In \emph{ECCV (6)}, pages 216--233, 2024.

\bibitem[Duan et~al.(2025)Duan, Kong, Cheng, et~al.]{duan2025truthprint}
Jinhao Duan, Fei Kong, Hao Cheng, et~al.
\newblock {TruthPrInt: Mitigating Large Vision-Language Models Object Hallucination Via Latent Truthful-Guided Pre-Intervention}.
\newblock In \emph{Proc. ICCV}, 2025.
\newblock arXiv:2503.10602.

\bibitem[Burns et~al.(2023)Burns, Ye, Klein, et~al.]{burns2023discovering}
Collin Burns, Haotian Ye, Dan Klein, et~al.
\newblock {Discovering Latent Knowledge in Language Models Without Supervision}.
\newblock In \emph{Proc. ICLR}, 2023.
\newblock arXiv:2212.03827.

\bibitem[Kossen et~al.(2024)Kossen, Han, Razzak, et~al.]{kossen2024semantic}
Jannik Kossen, Jiatong Han, Muhammed Razzak, et~al.
\newblock {Semantic Entropy Probes: Robust and Cheap Hallucination Detection in LLMs}.
\newblock \emph{arXiv preprint arXiv:2406.15927}, 2024.

\bibitem[Azaria and Mitchell(2023)]{azaria2023the}
Amos Azaria and Tom Mitchell.
\newblock {The Internal State of an LLM Knows When It's Lying}.
\newblock In \emph{Proc. EMNLP Findings}, 2023.
\newblock arXiv:2304.13734.

\bibitem[Li et~al.(2023)Li, Patel, Viégas, et~al.]{li2023inferencetime}
Kenneth Li, Oam Patel, Fernanda Viégas, et~al.
\newblock {Inference-Time Intervention: Eliciting Truthful Answers from a Language Model}.
\newblock In \emph{Proc. NeurIPS}, 2023.
\newblock arXiv:2306.03341.

\bibitem[Sarkar et~al.(2025)Sarkar, Che, Gavin, Beerel, and Kundu]{sarkar2025mitigating}
Sreetama Sarkar, Yue Che, Alex Gavin, Peter~Anthony Beerel, and Souvik Kundu.
\newblock Mitigating hallucinations in vision-language models through image-guided head suppression.
\newblock In \emph{Proceedings of the 2025 Conference on Empirical Methods in Natural Language Processing}, pages 12481--12500, Suzhou, China, 2025. Association for Computational Linguistics.

\bibitem[Shi et~al.(2026)Shi, Yang, and Liu]{shi2026oneshot}
Youxu Shi, Suorong Yang, and Dong Liu.
\newblock One-shot optimized steering vector for hallucination mitigation for {VLMs}.
\newblock \emph{arXiv preprint arXiv:2601.23041}, 2026.

\bibitem[Yu et~al.(2025{\natexlab{a}})Yu, Zhang, Li, Xu, Yao, Chen, Lu, Cui, Dang, He, Feng, Song, Zheng, Liu, Chua, and Sun]{yu2024rlaifv}
Tianyu Yu, Haoye Zhang, Qiming Li, Qixin Xu, Yuan Yao, Da Chen, Xiaoman Lu, Ganqu Cui, Yunkai Dang, Taiwen He, Xiaocheng Feng, Jun Song, Bo Zheng, Zhiyuan Liu, Tat-Seng Chua, and Maosong Sun.
\newblock {RLAIF-V}: Open-source {AI} feedback leads to super {GPT-4V} trustworthiness.
\newblock In \emph{Proceedings of the IEEE/CVF Conference on Computer Vision and Pattern Recognition (CVPR)}, 2025{\natexlab{a}}.

\bibitem[Yu et~al.(2025{\natexlab{b}})Yu, Chen, Kuang, Feng, Zhou, Wang, and Dobbie]{yu2025causally}
Liu Yu, Zhonghao Chen, Ping Kuang, Zhikun Feng, Fan Zhou, Lan Wang, and Gillian Dobbie.
\newblock Causally-grounded dual-path attention intervention for object hallucination mitigation in {LVLMs}.
\newblock \emph{arXiv preprint arXiv:2511.09018}, 2025{\natexlab{b}}.

\bibitem[Akl et~al.(2026)Akl, Khamis, Cheraghian, et~al.]{akl2026hime}
Ahmed Akl, Abdelwahed Khamis, Ali Cheraghian, et~al.
\newblock {HIME: Mitigating Object Hallucinations in LVLMs via Hallucination Insensitivity Model Editing}.
\newblock \emph{arXiv preprint arXiv:2602.18711}, 2026.

\bibitem[Yang et~al.(2025)Yang, Zheng, Chen, et~al.]{yang2025nullu}
Le Yang, Ziwei Zheng, Boxu Chen, et~al.
\newblock {Nullu: Mitigating Object Hallucinations in Large Vision-Language Models via HalluSpace Projection}.
\newblock In \emph{Proc. CVPR}, 2025.
\newblock arXiv:2412.13817.

\bibitem[Yue et~al.(2024)Yue, Zhang, and Jin]{yue2024less}
Zihao Yue, Liang Zhang, and Qin Jin.
\newblock Less is more: Mitigating multimodal hallucination from an {EOS} decision perspective.
\newblock In \emph{Proceedings of the 62nd Annual Meeting of the Association for Computational Linguistics (Volume 1: Long Papers)}, pages 11766--11781, 2024.

\end{thebibliography}
}

\clearpage
\appendix
\input{appendix_setup}
\onecolumn
\begin{multicols}{2}
\section*{Appendix Overview}
\addcontentsline{toc}{section}{Appendix Overview}

Tabs.~\ref{tab:main} and~\ref{tab:compare} use the standard F1 ($P=1-\text{CHAIRi}$, $R$ the official Recall). In this appendix, the COCO F1 on the development set (Tab.~\ref{tab:G1}) is the harmonic mean of $1-\text{CHAIRi}$ and Cover of that table and is compared only within it. AMBER F1 is derived from CHAIR and Cover.
The claims of this paper rest on the test runs: the 500 COCO images left after removing the development and cold-start images, all 1{,}004 AMBER images (including the 100 cold-start and 160 development images), and the remaining 744 AMBER images. Apart from each backbone's layers and operating point $q$, the four backbones and both datasets share one set of values.
Neither fitting the verifier nor online calibration reads ground truth; labels are used only to select layers on the development set, the other shared values are also set on development data, and the test runs serve only to report results.

The appendix comprises \ref{sec:A}, qualitative examples; \ref{sec:B}, algorithms; \ref{sec:C}, signal design and layer selection; \ref{sec:D}, ablations of the full memory pipeline; \ref{sec:E}, full end-to-end results; \ref{sec:F}, comparison with published methods; \ref{sec:G}, implementation details, configuration and cost.

\paragraph{Configuration and data.}
Unless stated otherwise, all tables and figures use the configuration of \ours in Sec.~\ref{sec:method}. Results on the development set (1{,}000 COCO and 160 AMBER images) serve to select layers, set values and analyze the mechanism, and are not evidence of generalization.

\paragraph{Correspondence with the main text.}
The analysis of Sec.~\ref{sec:analysis} corresponds to \cref{sec:C} and Sec.~\ref{sec:D.1}. For the method of Sec.~\ref{sec:method}, the algorithms, signals and values, the judge of \ours{} and memory, and the intervention correspond to \cref{sec:B}, \cref{sec:C}, Secs.~\ref{sec:D.2}--\ref{sec:D.4}, and Secs.~\ref{sec:G.3} and~\ref{sec:G.4}, respectively. For Sec.~\ref{sec:experiments}, the setup, full results, comparison and external vocabulary correspond to \cref{sec:G.1}, \cref{sec:E}, \cref{sec:F} and \cref{sec:G.3}, respectively, and the component ablations to Secs.~\ref{sec:C.2}, \ref{sec:D.3} and~\ref{sec:G.3}.

\section{Qualitative Examples}
\label{sec:A}

\begin{figure}[!b]
  \centering
  \includegraphics[width=5in]{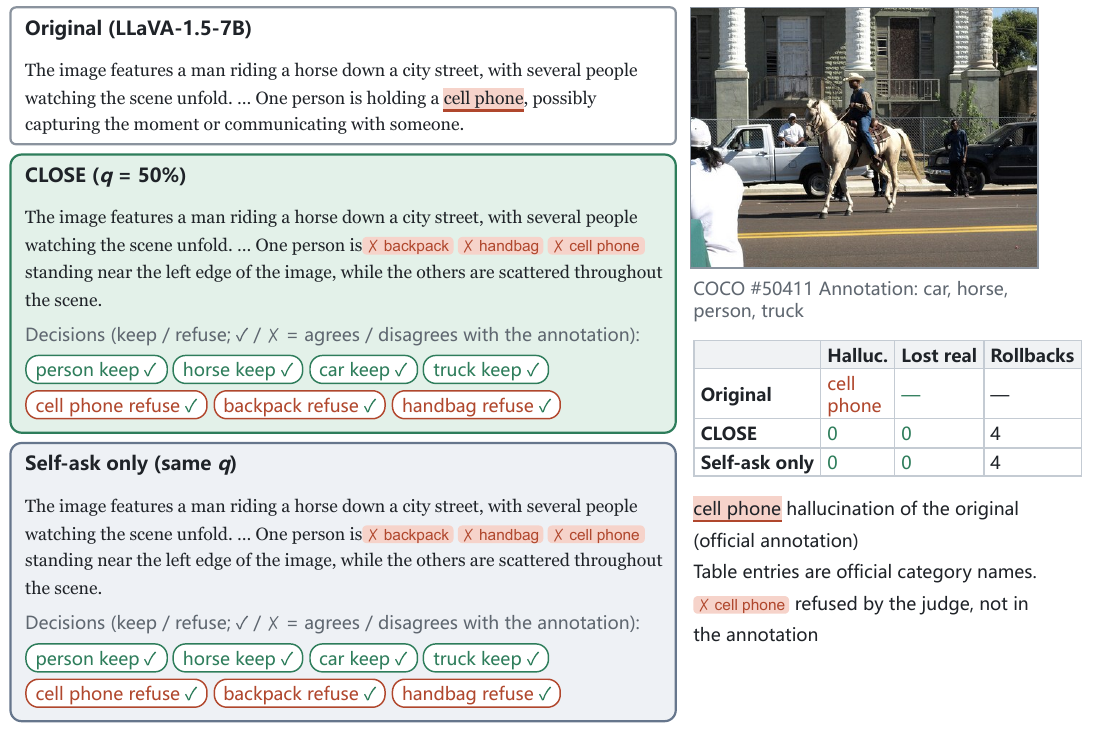}
  \caption{LLaVA-1.5-7B, COCO, $q=50\%$. \ours makes 7 decisions (keeps 4, refuses 3), all agreeing with the annotation; it removes the cell phone written by the original model and loses no real object. Self-asking alone also loses no real object on this image.}
  \label{fig:A1}
\end{figure}

\begin{figure*}[!htbp]
  \centering
  \includegraphics[width=5in]{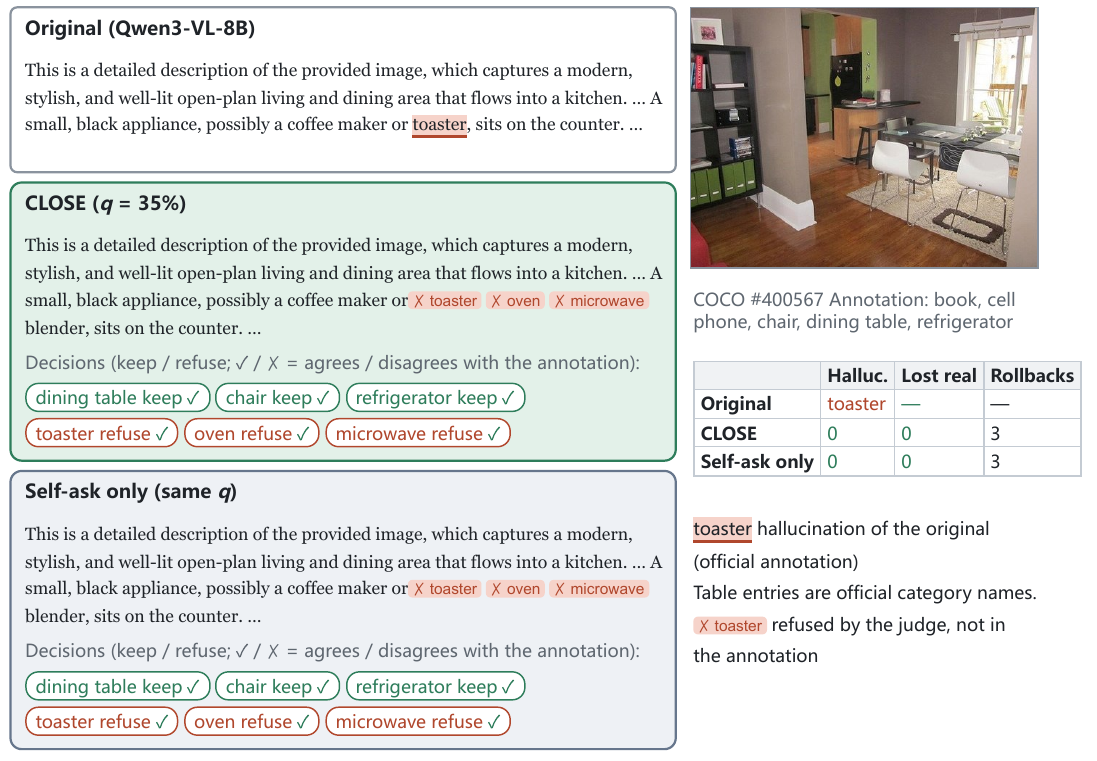}
  \caption{Qwen3-VL-8B, COCO, $q=35\%$. \ours makes 6 decisions (keeps 3, refuses 3), all agreeing with the annotation; it removes the toaster written by the original model and loses no real object. Self-asking alone also loses no real object on this image. A limitation: after the toaster is refused, both \ours and self-asking alone rewrite it as ``coffee maker or blender''. The coffee maker was already written by the original model and the blender appears in the rewrite; neither is in the image. Both lie outside the official vocabulary, so neither the official count nor the judge sees them; only a check with the external vocabulary finds them (\cref{sec:G.3}).}
  \label{fig:A2}
  \vspace{8pt}
  \includegraphics[width=5in]{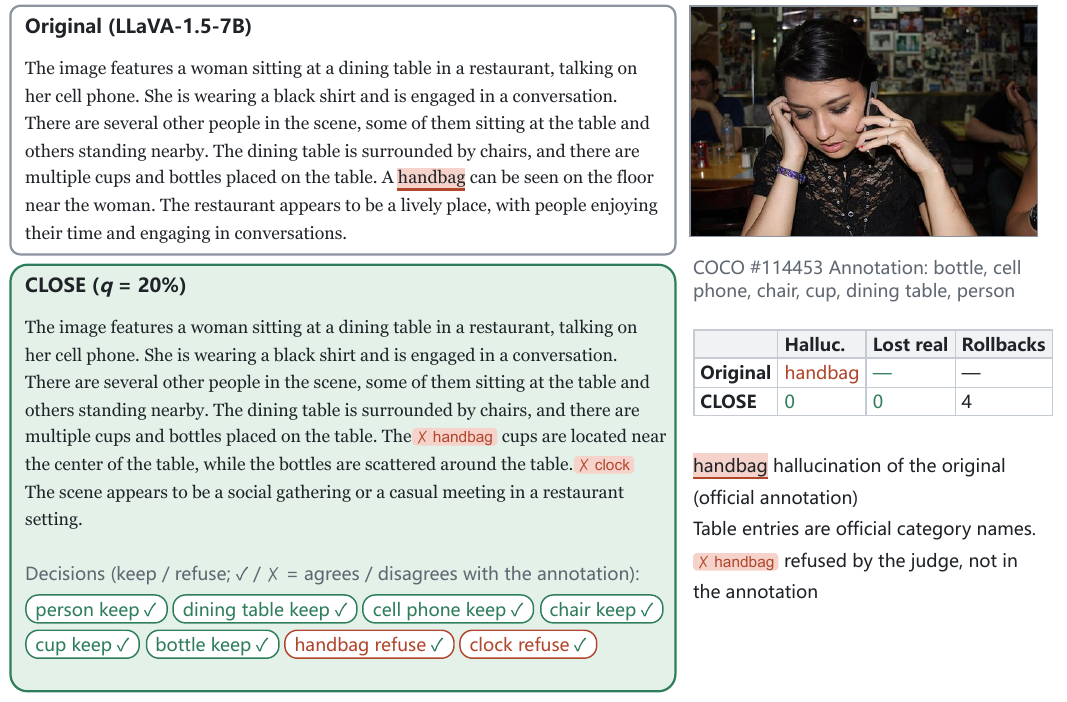}
  \caption{LLaVA-1.5-7B, COCO, $q=20\%$. \ours makes 8 decisions (keeps 6, refuses 2: the handbag and a clock that appears in the rewrite), all agreeing with the annotation; it removes the handbag written by the original model and loses no real object. The caption changes from 95 to 101 words. The rewrite's ``the bottles are scattered around the table'' overstates the image, in which one bottle is visible, but the bottle is in the annotation, so category-level counting does not see it.}
  \label{fig:A3}
\end{figure*}

\begin{figure*}[!htbp]
  \centering
  \includegraphics[width=5in]{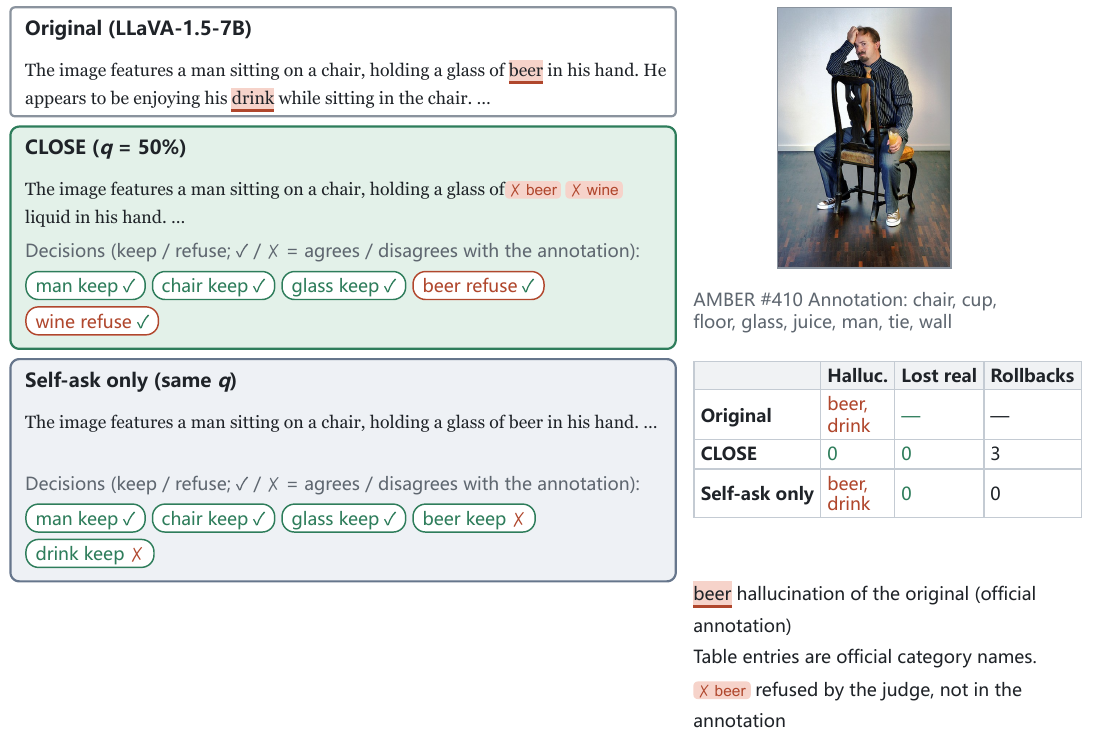}
  \caption{LLaVA-1.5-7B, AMBER, $q=50\%$. \ours makes 5 decisions (keeps 3, refuses 2: the beer and a wine that appears in the rewrite), all agreeing with the annotation; neither the beer nor the drink written by the original model remains, and no real object is lost. Self-asking alone keeps both the beer and the drink on this image.}
  \label{fig:A5}
  \vspace{8pt}
  \includegraphics[width=5in]{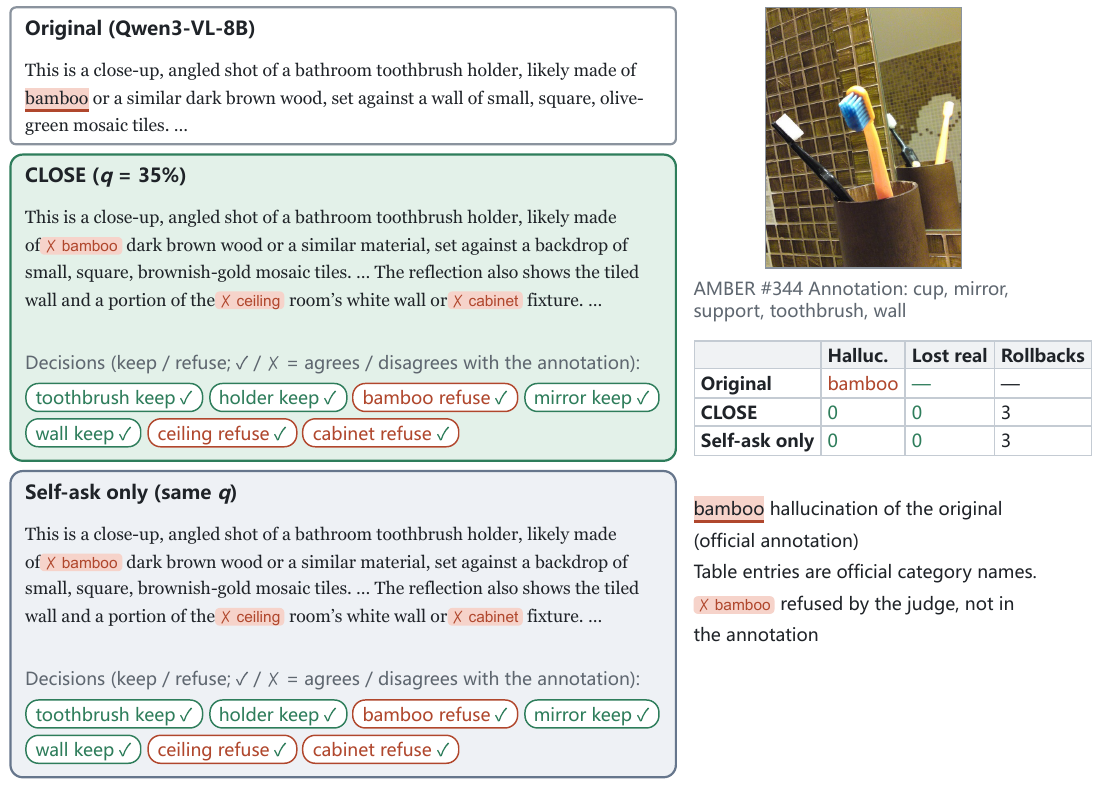}
  \caption{Qwen3-VL-8B, AMBER, $q=35\%$. \ours makes 7 decisions (keeps 4, refuses 3), all agreeing with the annotation; it removes the bamboo written by the original model and loses no real object. Self-asking alone makes the same decisions on this image.}
  \label{fig:A6}
\end{figure*}

All examples come from the test runs, and all texts are verbatim model outputs, with ``\ldots'' marking omitted sentences. A cross ($\times$) marks each mention refused by the judge; red denotes objects absent from the annotation and orange objects present in it. Figs.~\ref{fig:A1}, \ref{fig:A2}, \ref{fig:A5} and~\ref{fig:A6} also show self-asking alone at the same $q$.
\end{multicols}

\onecolumn
\FloatBarrier
\section{Algorithms}
\label{sec:B}

\providecommand{\Cmt}[1]{\ \ $\triangleright$~#1}
\algrenewcommand\algorithmicrequire{\textbf{Input:}}

The method consists of three procedures: the outer loop over the whole test stream (Algorithm~\ref{alg:B1}), verification during generation for one image (Algorithm~\ref{alg:B2}), and self-training of the scorer on the memory (Algorithm~\ref{alg:B3}). Notation follows Sec.~\ref{sec:method}; $R(\cdot)$ is the rank in the memory divided by the number of samples, and $F_{\M}(\cdot)$ is the empirical quantile among values of the same type in the memory.

The model weights are unchanged. The method reads the hidden states at the layers of the scorer and of the four signals and, when self-asking, the yes / no logits at the last layer. It writes back only the truncated KV cache and the ban mask at the output. Generation, self-asking and re-reading are forward passes on the GPU; the nearest-neighbor search for S2, the quantile threshold and the self-training of the scorer are computed on the CPU.

\begin{algorithm}[H]
\small
\caption{Outer loop of \ours over the test stream}
\label{alg:B1}
\begin{algorithmic}[1]
\Require frozen VLM; unlabeled cold-start images $\mathcal C_0$; test stream $I_1,I_2,\dots$; refusal rate $q$
\Statex \textbf{Values:} $\lambda=0.7$, $\beta=12.5\%$, $\gamma=50\%$, $C=0.01$, $a=0.8$, $J=10$; threshold refreshed every $T=10$ images, scorer re-trained every $T_h=100$ images; at most 20 rollbacks per image
\State $\M\gets\emptyset$
\For{$I\in\mathcal C_0$} \Cmt{cold start: generation only, no intervention}
  \State $y\gets\mathrm{Generate}(I)$; self-ask $p$ for the first mention of each object category in $y$ as in line~\ref{alg:ask} of Algorithm~\ref{alg:B2}; $\M\gets\M\cup\mathrm{Reread}(I,y,\emptyset)$ \label{alg:coldask}
\EndFor
\State $\theta\gets\mathrm{SelfTrain}(\M)$ \Cmt{Algorithm~\ref{alg:B3}}
\For{$n=1,2,\dots$}
  \State $(y_n,\mathcal P_n)\gets\mathrm{Verify}(I_n,\M,\theta,q)$ \Cmt{Algorithm~\ref{alg:B2}; $\mathcal P_n$ holds the prefixes of refused mentions}
  \State $\M\gets\M\cup\mathrm{Reread}(I_n,y_n,\mathcal P_n)$ \Cmt{re-read after this image is written; refused mentions also enter the memory}
  \State \textbf{if} $n\bmod T=0$ \textbf{then} refresh the threshold $\tau_q$ and the memory values used for ranking \textbf{end if} \Cmt{earlier images only}
  \State \textbf{if} $n\bmod T_h=0$ \textbf{then} $\theta\gets\mathrm{SelfTrain}(\M)$ \textbf{end if}
  \State \textbf{output} $y_n$
\EndFor
\Statex
\Function{Reread}{$I,y,\mathcal P$} \Cmt{teacher-forced forward pass, no generation}
  \State $I^\circ\gets$ uniform gray image of the same size as $I$
  \For{$x\in\{y\}\cup\mathcal P$}
    \State one forward pass under $I^\circ$ \Cmt{the original-image side reuses the hidden states from generation}
    \For{each first mention $m$ of an object category in $x$ (first token at position $t$)}
      \State compute $S_1,d,e$ as in Sec.~\ref{sec:head}; record $\big(h^{(l_h)}_t,h^{(l_2)}_t,S_1,d,e,p\big)$
      \Statex \hspace*{4.5em}\parbox[t]{\dimexpr\linewidth-4.5em}{$\triangleright$~$p$ is the self-asked probability at generation (for cold-start images, from line~\ref{alg:coldask}); $S_2$ is computed from the memory during self-training}
    \EndFor
  \EndFor
  \State \Return all records
\EndFunction
\end{algorithmic}
\end{algorithm}

\par\addvspace{\intextsep}
\ifdefined\nolinenumbers\nolinenumbers\fi
\begingroup
\small
\noindent\hrule height.8pt depth0pt\kern2pt
\refstepcounter{algorithm}\noindent\textbf{Algorithm~\thealgorithm}\ Verification during generation for one image, Verify\label{alg:B2}\par
\kern2pt\hrule\kern2pt
\begin{algorithmic}[1]
\Require image $I$; memory $\M$; scorer $\theta=(w,b,\mu,\sigma)$; refusal rate $q$
\State $\mathcal B,\mathcal R,\mathcal A,\mathcal P\gets\emptyset$; $n_{\mathrm{rb}}\gets0$ \Cmt{ban list, refused categories, accepted categories, refused prefixes, number of rollbacks}
\State $\tau_q$ and $F_{\M}$ take the values of the latest refresh in Algorithm~\ref{alg:B1} ($\tau_q=Q_{1-q}\big(\{r(m'):m'\in\M\}\big)$)
\While{the caption is not finished}
  \State generate the next token, setting to $-\infty$ the logit of the token that would complete any whole-word token sequence in $\mathcal B$ (same semantics as \texttt{bad\_words\_ids}); once $\mathcal R\neq\emptyset$, the logits of tokens containing non-ASCII letters are also set to $-\infty$
  \If{a word $u$ is completed (first token at position $t$)}
    \State \textbf{if} $\mathrm{Recognize}(u)=c'\in\mathcal R$ \textbf{then} $\mathrm{Caught}(c',t)$; \textbf{continue} \Cmt{the refused object is written in other words}
    \State $c\gets$ category of $u$ in the official vocabulary; \textbf{if} $c$ does not exist or $c\in\mathcal A$ \textbf{then continue}
    \State $p\gets P(\text{yes}\mid I,\ \text{Is there a } c \text{ in this image?})$ \Cmt{reuses the KV cache of the image prefix} \label{alg:ask}
    \State $s\gets w^\top(h^{(l_h)}_t-\mu)/\sigma+b$; $r\gets\lambda F_{\M}(1-p)+(1-\lambda)F_{\M}(s)$
    \Statex \hspace*{3em}\parbox[t]{\dimexpr\linewidth-3em}{$\triangleright$~$h$ is taken by a forward hook from this generation step, with no extra forward pass}
    \If{$r>\tau_q$}
      \State $\mathcal R\gets\mathcal R\cup\{c\}$; add the prefix up to $u$ to $\mathcal P$; $\mathrm{Caught}(c,t)$
    \Else
      \State $\mathcal A\gets\mathcal A\cup\{c\}$
    \EndIf
  \EndIf
  \State \textbf{if} a rollback has occurred, a sentence is completed and $\neg\,\mathrm{Fluent}(\text{sentence})$ \textbf{then} back off one segment from the first faulty word; $n_{\mathrm{rb}}\gets n_{\mathrm{rb}}+1$
\EndWhile
\State \Return $y,\mathcal P$
\Statex
\Function{Caught}{$c,t$} \Cmt{laddered ban}
  \State $\mathcal B\gets\mathcal B\cup\{$all official spellings of $c$ and the spelling used this time$\}$; $n_{\mathrm{rb}}\gets n_{\mathrm{rb}}+1$
  \State \textbf{if} $n_{\mathrm{rb}}=20$ \textbf{then} back up to the start of the current segment, delete the segment, and continue generating with checks as usual; \textbf{return}; \textbf{else if} $n_{\mathrm{rb}}>20$ \textbf{then} delete the sentence and end this image; in the first sentence, end the sentence before the refused noun phrase instead
  \State $k\gets$ number of catches so far in this sentence
  \State \textbf{if} $k=1$ \textbf{then} roll back to $t$ \textbf{else if} $k\le4$ \textbf{then} roll back to the start $t_{\mathrm{NP}}$ of the noun phrase and ban the original first token there
  \State \textbf{else} end the sentence or delete it (the first sentence is not deleted), and lock the objects of this sentence
\EndFunction
\Statex
\Function{Recognize}{$u$} \Cmt{any hit returns the refused category $c'$}
  \State after normalization, $u$ shares a stem with a spelling of $c'$, is an inflectional variant, is within edit distance $\le1$, is a truncation, or forms the name of $c'$ together with the previous word
  \State look up the one-hot encoding of the recognition table: $u$ belongs to $c'$ in the recognition table derived from WordNet
  \State the token just blocked by the decoder, joined with the letters already written, is the beginning of a name of $c'$
\EndFunction
\Statex Rolling back to $t$: the KV cache is truncated before $t$ and generation continues; on hybrid linear-attention backbones, the latest state snapshot before $t$ is restored (\cref{sec:G.4}).
\end{algorithmic}
\kern2pt\hrule\relax
\endgroup
\par\addvspace{\intextsep}
\ifdefined\linenumbers\linenumbers\fi

\begin{algorithm}[H]
\small
\caption{Self-training of the scorer on the memory, SelfTrain}
\label{alg:B3}
\begin{algorithmic}[1]
\Require memory $\M$, each mention with $h,S_1,d,e,p$
\State compute $S_2$ as in Sec.~\ref{sec:head} from the mentions of other images in $\M$; $z^0\gets R\big(R(-S_1)+R(S_2)+R(d)+R(e)\big)$ \Cmt{four-signal start evidence, equal weights}
\State $\mu,\sigma\gets$ mean and standard deviation of $h$ over $\M$; $x\gets(h-\mu)/\sigma$
\State $\tilde y(z)$: mentions with $R(z)\ge1-\beta$ are labeled hallucinated, those with $R(z)\le\gamma$ real, and the rest are excluded
\State $f_0\gets\LR_C\big(x,\tilde y(z^0)\big)$ \Cmt{class-balanced, L2-regularized logistic regression}
\For{$j=1,\dots,J$}
  \State $f_j\gets\LR_C\big(x,\tilde y(a\,R(f_{j-1}(x))+(1-a)\,z^0)\big)$
\EndFor
\State recompute $s$ and $r$ of all mentions in $\M$ with $(w,b)$ of $f_J$ and $(\mu,\sigma)$ \Cmt{old scores are not kept}
\State \Return $\theta=(w,b,\mu,\sigma)$
\end{algorithmic}
\end{algorithm}

\twocolumn

\twocolumn
\FloatBarrier
\setcounter{section}{2}
\section{Signal Design and Layer Selection}
\label{sec:C}

\subsection{Evaluation protocol}
\label{sec:C.1}

With a 200-image cold start, the standard deviation of self-training itself is about 0.009, larger than the effect of most signal changes, so a single comparison is unreliable. The design conclusions of this section are judged under one protocol:
\begin{itemize}[leftmargin=*,itemsep=1pt,topsep=2pt]
\item the scorer is self-trained only on the unlabeled cold-start memory and is frozen before scoring;
\item one component is changed at a time, with a paired comparison over 20 resamples with replacement of the cold-start memory (10 for the coordinate-ascent search), and one run of the online stage is added;
\item a conclusion must be reproduced on both COCO (cold start 200 images $\rightarrow$ development set 1{,}000 images) and AMBER (cold start 100 images $\rightarrow$ development set 160 images).
\end{itemize}

Tabs.~\ref{tab:C1}, \ref{tab:C3}, \ref{tab:C4} and \ref{tab:C5} use the \textbf{test-stream protocol}: the scorer is self-trained on the cold-start memory and then frozen, and the AUROC is computed on the mention stream of the test run. Intervals of differences in Tabs.~\ref{tab:C1} and \ref{tab:C5} are 95\% intervals of a paired bootstrap over images (300 resamples), and $^*$ marks an interval that excludes 0.

\subsection{Choice of the four signals}
\label{sec:C.2}

\begin{table*}[t]
\centering
\caption{\textbf{Changes of the scorer and \ours{} when only one signal is used or one signal is removed.} ``w/o'' columns: AUROC after removing that signal minus that with all four signals. ``Only'' columns: in the scorer rows, the scorer AUROC when only that signal is the teacher; in the \ours{} rows, ``Only S1'' gives the AUROC of \ours{} with S1 as the only teacher minus that with all four signals. AUROC of the scorer / \ours{} with all four signals: on COCO, LLaVA-1.5-7B 0.902 / 0.949, Qwen3-VL-8B 0.905 / 0.967, Qwen3.5-9B 0.917 / 0.968, Qwen2.5-VL-7B 0.875 / 0.958; on AMBER, 0.866 / 0.945, 0.886 / 0.948, 0.895 / 0.950 and 0.832 / 0.936 in the same order. Bold: in each row, the largest drop among the four ``w/o'' columns and the highest AUROC among the four ``Only'' columns.}
\label{tab:C1}
\small
\setlength{\tabcolsep}{4pt}
\renewcommand{\arraystretch}{1.15}
\begin{tabular}{llcccccccc}
\toprule
 & & \multicolumn{4}{c}{Signal removed (difference)} & \multicolumn{4}{c}{Only teacher} \\
\cmidrule(lr){3-6}\cmidrule(lr){7-10}
Backbone $\cdot$ data & & w/o S1 & w/o S2 & w/o $d$ & w/o $e$ & S1 & S2 & $d$ & $e$ \\
\midrule
\multirow{2}{*}{LLaVA-1.5-7B $\cdot$ COCO} & Scorer & $-$0.028$^*$ & $-$0.015$^*$ & \textbf{$-$0.031}$^*$ & $-$0.004 & \textbf{0.875} & 0.392 & 0.859 & 0.822 \\
 & \ours{} & $-$0.000 & $-$0.003$^*$ & \textbf{$-$0.006}$^*$ & $-$0.002$^*$ & $-$0.007$^*$ & -- & -- & -- \\
\addlinespace[2pt]
\multirow{2}{*}{Qwen3-VL-8B $\cdot$ COCO} & Scorer & \textbf{$-$0.052}$^*$ & $-$0.035$^*$ & $-$0.018$^*$ & $-$0.030$^*$ & \textbf{0.868} & 0.847 & 0.554 & 0.778 \\
 & \ours{} & $-$0.003 & \textbf{$-$0.005}$^*$ & \textbf{$-$0.005}$^*$ & $-$0.001 & $-$0.005$^*$ & -- & -- & -- \\
\addlinespace[2pt]
\multirow{2}{*}{Qwen3.5-9B $\cdot$ COCO} & Scorer & \textbf{$-$0.036}$^*$ & $-$0.028$^*$ & +0.002 & $-$0.016$^*$ & \textbf{0.885} & 0.794 & 0.523 & 0.833 \\
 & \ours{} & \textbf{$-$0.004}$^*$ & $-$0.003$^*$ & $-$0.002 & $-$0.002$^*$ & $-$0.004$^*$ & -- & -- & -- \\
\addlinespace[2pt]
\multirow{2}{*}{Qwen2.5-VL-7B $\cdot$ COCO} & Scorer & \textbf{$-$0.074}$^*$ & $-$0.030$^*$ & $-$0.023$^*$ & $-$0.012 & \textbf{0.809} & 0.731 & 0.776 & 0.572 \\
 & \ours{} & \textbf{$-$0.008}$^*$ & $-$0.002 & $-$0.005$^*$ & $-$0.002 & $-$0.009$^*$ & -- & -- & -- \\
\midrule
\multirow{2}{*}{LLaVA-1.5-7B $\cdot$ AMBER} & Scorer & $-$0.001 & +0.005$^*$ & \textbf{$-$0.043}$^*$ & $-$0.006$^*$ & 0.800 & 0.677 & 0.837 & \textbf{0.842} \\
 & \ours{} & +0.001 & +0.000 & \textbf{$-$0.002}$^*$ & $-$0.000 & $-$0.006$^*$ & -- & -- & -- \\
\addlinespace[2pt]
\multirow{2}{*}{Qwen3-VL-8B $\cdot$ AMBER} & Scorer & \textbf{$-$0.032}$^*$ & +0.002 & $-$0.026$^*$ & $-$0.010$^*$ & 0.746 & \textbf{0.822} & 0.728 & 0.795 \\
 & \ours{} & \textbf{$-$0.005}$^*$ & +0.002$^*$ & $-$0.003$^*$ & $-$0.001$^*$ & $-$0.012$^*$ & -- & -- & -- \\
\addlinespace[2pt]
\multirow{2}{*}{Qwen3.5-9B $\cdot$ AMBER} & Scorer & \textbf{$-$0.051}$^*$ & +0.007$^*$ & $-$0.015$^*$ & $-$0.010$^*$ & 0.833 & 0.776 & 0.656 & \textbf{0.848} \\
 & \ours{} & \textbf{$-$0.005}$^*$ & +0.001 & $-$0.002$^*$ & $-$0.001$^*$ & $-$0.006$^*$ & -- & -- & -- \\
\addlinespace[2pt]
\multirow{2}{*}{Qwen2.5-VL-7B $\cdot$ AMBER} & Scorer & +0.003 & \textbf{$-$0.084}$^*$ & +0.011$^*$ & +0.021$^*$ & 0.635 & 0.790 & \textbf{0.818} & 0.553 \\
 & \ours{} & $-$0.004$^*$ & \textbf{$-$0.009}$^*$ & $-$0.000 & +0.000 & $-$0.025$^*$ & -- & -- & -- \\
\bottomrule
\end{tabular}
\end{table*}

\begin{itemize}[leftmargin=*,itemsep=1pt,topsep=2pt]
\item Scorer: with all four signals as teachers, the scorer is stronger in all eight groups than with any single teacher (all 32 comparisons significant), and each signal is the strongest single teacher in some group. Removing S1, S2, $d$ or $e$ alone makes the scorer significantly worse in 6, 5, 6 and 5 groups, respectively; some removals improve it (S2 in 2 groups, $d$ and $e$ in 1 group each, all on AMBER).
\item \ours{}: removing any one signal changes it by $-$0.009 to +0.002, less than for the scorer, because self-asking carries a weight of 0.7 in the fusion.
\item Removing any one signal saves no GPU computation: the four signals come from the hidden states of the original image already recorded during generation and the same single gray-image re-read at the end of each image, and S2 only computes nearest neighbors on the CPU.
\end{itemize}

\subsection{Layer selection}
\label{sec:C.3}

\begin{figure*}[t]
\centering
\includegraphics[width=\textwidth]{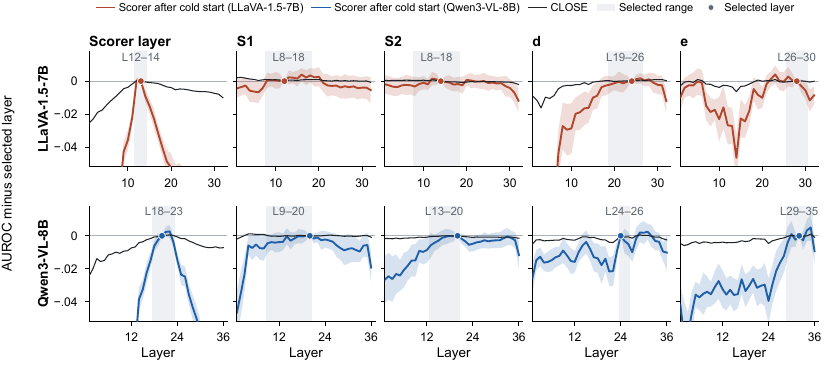}
\caption{When the layer of only one component moves, the change of the scorer-alone AUROC after the cold start relative to the selected layer (colored lines; bands: $\pm$1 standard deviation over 20 resamples; shading: layer range; thin black lines: the same change of the AUROC of \ours{}). The y-axis is cut at $-$0.05.}
\label{fig:C1}
\end{figure*}

\paragraph{Criterion and procedure (the same for all four backbones).}
\begin{itemize}[leftmargin=*,itemsep=1pt,topsep=2pt]
\item \textbf{Criterion: the AUROC of \ours{} (0.7 self-ask + 0.3 scorer)} on the labeled development set (1{,}000 COCO + 160 AMBER images); for every change of layers, the scorer is self-trained again from scratch on the unlabeled cold-start memory (200 COCO + 100 AMBER images). The scorer-alone AUROC is not the criterion: the two criteria choose largely the same scorer layer (Sec.~\ref{sec:C.6}), but some components affect only the scorer-alone AUROC.
\item \textbf{Range}: each component starts from the layer span given by prior work, which is narrowed to one contiguous span by the scorer-alone AUROC after the cold start (Fig.~\ref{fig:C1}).
\item \textbf{Search}: every backbone keeps the same structure (scorer, S1 and S2 in the middle layers, $d$ in the middle-to-late layers, $e$ in the late layers); within it, each component is scanned layer by layer with two rounds of coordinate ascent (10 resamples), and the top 3 layers of each component form a joint grid (243 combinations).
\item \textbf{Layer-switching rule}: the grid optimum is compared with the selected layers in 20 paired comparisons, and the layers are switched only if the optimum wins at least 19 times on both COCO and AMBER with a positive mean gain.
\end{itemize}

\paragraph{Why each component sits at its depth.}
The structure follows the stages of image--text processing that prior work located on LLaVA-1.5:

\par\noindent\rule{\columnwidth}{0.6pt}\par
\noindent{\small\textbf{Stage (LLaVA-1.5, 32 layers)}: findings of prior work $\Rightarrow$ our components placed here (LLaVA-1.5 layer / relative depth)}\par
\noindent\rule{\columnwidth}{0.4pt}\par
\begin{itemize}[leftmargin=*,itemsep=2pt,topsep=1pt]\small
\item \textbf{Shallow, about layers 1--4}: coarse features of the whole image are first written into the text positions~\cite{zhang2025crossmodal}. $\Rightarrow$ None.
\item \textbf{Middle, about layers 5--18 (``visual information enrichment'')}: visual information about specific objects is written into the corresponding text positions a second time~\cite{zhang2025crossmodal}; real objects receive more image attention than hallucinated ones~\cite{jiang2025devils}. $\Rightarrow$ Scorer 13, S1 12, S2 14 (0.38--0.44).
\item \textbf{Middle-to-late, about layers 19--26 (``semantic refinement'')}: through the logit lens, image tokens already read out words consistent with their patches~\cite{jiang2025devils}; the language prior dilutes visual information as generation proceeds~\cite{li2025the}. $\Rightarrow$ $d$ 24 (0.75).
\item \textbf{Window before the last layers}: semantic words such as objects are most prominent in the few layers before the last one, while the last layer favors function words~\cite{li2025the}. $\Rightarrow$ $e$ 28 (0.88).
\end{itemize}
\par\noindent\rule{\columnwidth}{0.6pt}\par

\begin{figure*}[t]
\centering
\includegraphics[width=0.75\textwidth]{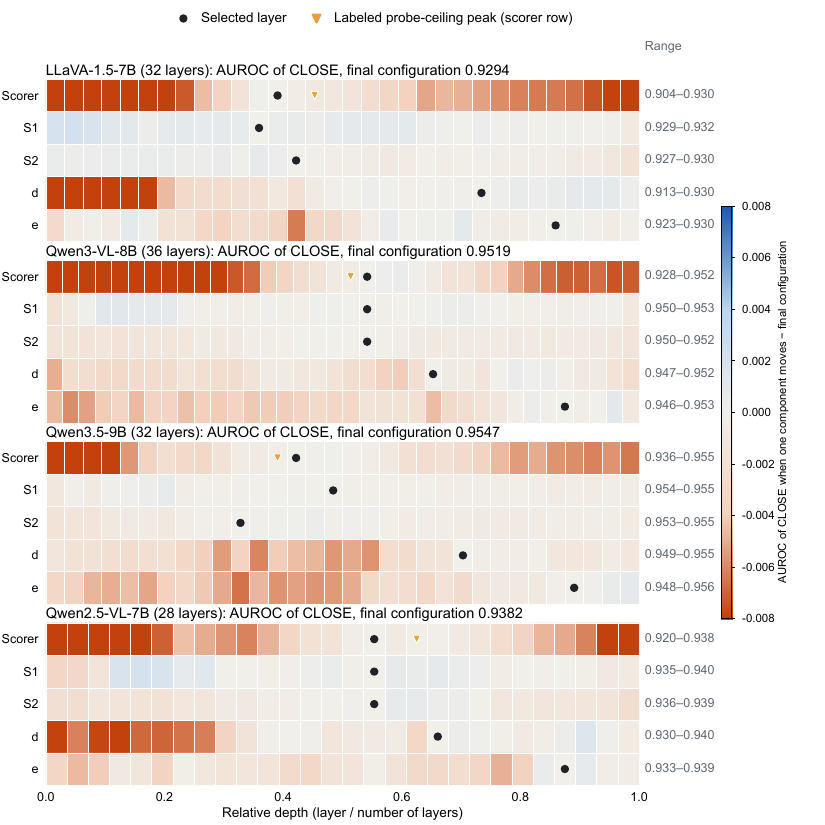}
\caption{Change of the AUROC of \ours{} relative to the final layers when each of the five components of the four backbones moves layer by layer (shared color scale; blue: better, orange: worse, gray: equal). Mean of the 1{,}000-image COCO and 160-image AMBER development sets, 20 resamples; four signals with equal weights. Solid dots: selected layers; triangles: the peak layer of the labeled probe ceiling.}
\label{fig:C4}
\end{figure*}

\begin{itemize}[leftmargin=*,itemsep=1pt,topsep=2pt]
\item \textbf{S2} shares the span the scorer reads (Fig.~\ref{fig:layer}; \cite{duan2025truthprint,burns2023discovering,kogilathota2026halp,du2024haloscope,kossen2024semantic}\bref{burns2023discovering}).
\item \textbf{$d$} is placed in the semantic-refinement span. Placing it in the first 0.2 of the depth on LLaVA-1.5-7B or the first 0.3 on Qwen2.5-VL-7B lowers the AUROC of \ours{} by 0.006--0.016, and placing it at depths 0.3--0.56 on Qwen3.5-9B lowers it by 0.004--0.006; beyond a depth of about 0.65, all four backbones are within 0.003 of the final layers (Fig.~\ref{fig:C4}).
\item \textbf{$e$} is the logit-lens entropy, meaningful only in late layers where the states are aligned with the vocabulary; following \cite{li2025the} it is placed in the window before the last layer (depths 0.88--0.91 on all four backbones).
\item \textbf{Qwen family}: \cite{zhang2025crossmodal,jiang2025devils} analyze only the LLaVA family; the Qwen family is mapped to the same stages by relative depth, and the layers are then selected on each backbone's development set. Moving S1 or S2 within the structural range changes \ours{} by at most 0.002 (Fig.~\ref{fig:C4}).
\item \textbf{Shallow layers}: they carry only coarse whole-image features and are excluded from the search in advance. Moving S1 alone to a very shallow layer (LLaVA-1.5-7B layer 2, Qwen2.5-VL-7B layer 5) raises the AUROC of \ours{} by about 0.002, because S1 is then less correlated with self-asking; this isolated case is not adopted.
\end{itemize}

\paragraph{Layer ranges.}
The cold-start images are resampled 20 times, and the scorer-alone AUROC after the cold start is computed on COCO and on AMBER. A contiguous span of layers is kept in which no layer is significantly worse than the selected layer on both datasets (the selected layer winning at least 19 of 20 times on each) and the mean difference does not exceed one resampling standard deviation (LLaVA-1.5-7B 0.0098, Qwen3-VL-8B 0.0085). No layer in the span beats the selected layer on both datasets either, so the choice within the range is governed mainly by data and noise.

\begin{figure*}[t]
\centering
\includegraphics[width=0.9\textwidth]{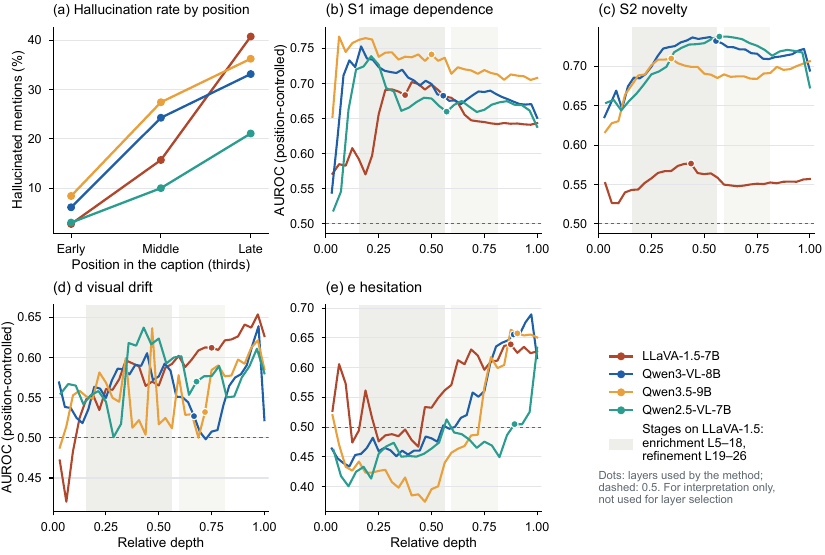}
\caption{Discriminative power of each signal itself across depth (following \cite{li2025the}; mean of the COCO and AMBER development sets). (a) Share of hallucinations in each third of the caption by character position; (b)--(e) AUROC of each signal itself (hallucination as the positive class; below 0.5 means the opposite direction), computed within each third and then averaged, so that position effects do not enter. Dots: layers used; light bands: the enrichment and refinement spans measured on LLaVA-1.5 by \cite{jiang2025devils}.}
\label{fig:C2}
\end{figure*}

\begin{figure*}[t]
\centering
\includegraphics[width=0.7\textwidth]{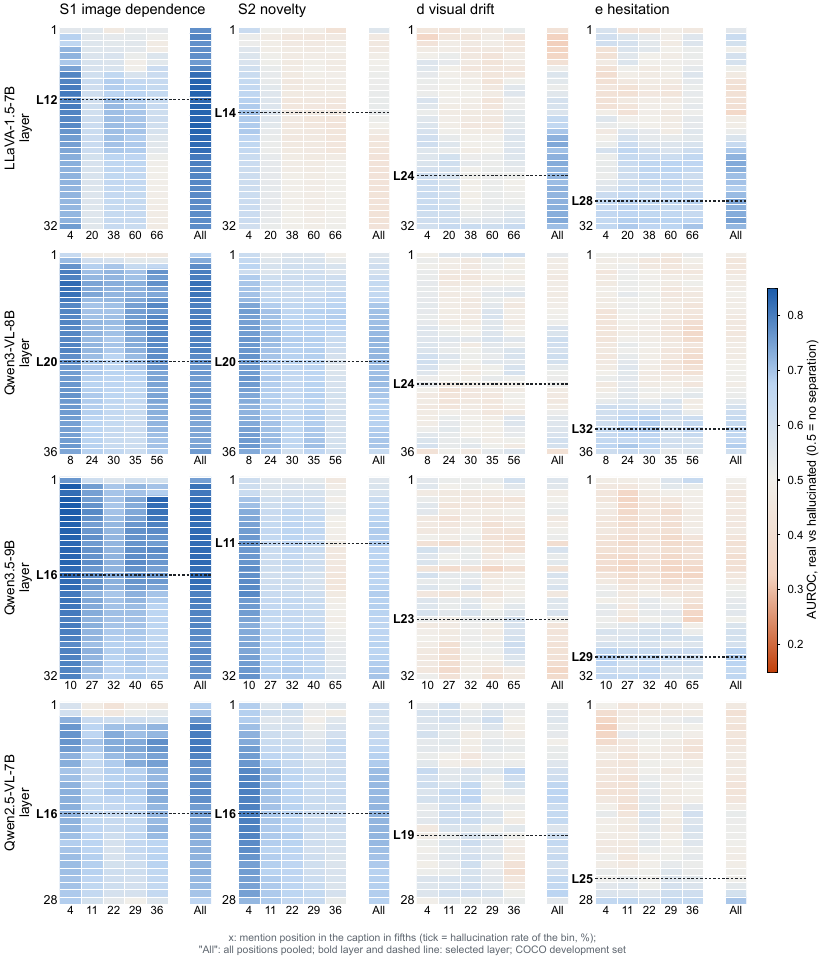}
\caption{The third-averaged curves of Fig.~\ref{fig:C2} expanded into a ``layer $\times$ position'' grid, following Fig.~2 of \cite{li2025the} (COCO development set; bold layer numbers and dashed lines: selected layers; x-axis: position in fifths, ticks giving the hallucination rate of each bin; rightmost column: all positions pooled). On LLaVA-1.5-7B, \textbf{$d$ hardly separates within a position bin and discriminates only when all positions are pooled} (pooled AUROC of the shallow, middle and deep thirds of depth: 0.43 / 0.65 / 0.72): $d$ mainly captures ``the later a word is written, the farther its state is from the image subspace'', so it is weak alone yet indispensable in the fusion (Sec.~\ref{sec:C.2}).}
\label{fig:C3}
\end{figure*}

\begin{itemize}[leftmargin=*,itemsep=1pt,topsep=2pt]
\item By the scorer AUROC, the selected layer differs from the best layer in its span by at most 0.005, and all 95\% intervals contain 0.
\item Moving one component within its range (LLaVA-1.5-7B and Qwen3-VL-8B; mean of COCO and AMBER) changes the scorer-alone AUROC after the cold start by $-$0.0090 to +0.0050 and the AUROC of \ours{} by only $-$0.0007 to +0.0011 relative to the selected layer.
\end{itemize}

\paragraph{Layer selection is insensitive to the self-ask precision.}
Under the same procedure, no backbone switches layers. Qwen3-VL-8B uses double-precision self-asking and the development-set self-asking of Qwen2.5-VL-7B and Qwen3.5-9B is in single precision; precision changes only the self-ask readout, and on Qwen3-VL-8B the AUROC of \ours{} at the selected layers is COCO 0.9515 and AMBER 0.9491 in single precision and 0.9516 and 0.9521 in double precision. On LLaVA-1.5-7B self-asking does not saturate, so precision does not affect the ranking.

\paragraph{Effect of double precision on the refusal rate.}
In single precision, 55\% of the AMBER mentions in memory tie at $\pyes=1$ and are refused in a batch once the threshold reaches this value, so the actual refusal rate at high $q$ far exceeds $q$; in double precision it essentially equals $q$ (Tab.~\ref{tab:C2}). For $q\le 30\%$ the two differ by at most 0.2 percentage points.

\begin{table}[t]
\centering
\caption{\textbf{Target $q$ and actual refusal rate} (mention stream of the Qwen3-VL-8B test run, causal replay with memory; the single-precision columns round the same answers to float32).}
\label{tab:C2}
\small
\renewcommand{\arraystretch}{1.12}
\begin{tabular}{ccccc}
\toprule
Target $q$ & \multicolumn{2}{c}{AMBER} & \multicolumn{2}{c}{COCO} \\
\cmidrule(lr){2-3}\cmidrule(lr){4-5}
 & Single & Double & Single & Double \\
\midrule
20\% & 21.1\% & 21.1\% & 22.1\% & 22.1\% \\
28\% & 28.5\% & 28.4\% & 29.0\% & 29.0\% \\
30\% & 30.5\% & 30.3\% & 30.0\% & 30.0\% \\
35\% & 36.5\% & 35.0\% & 35.5\% & 35.5\% \\
40\% & 46.4\% & 39.8\% & 40.9\% & 40.9\% \\
45\% & 68.1\% & 44.7\% & 45.2\% & 44.9\% \\
50\% & 90.7\% & 49.5\% & 51.5\% & 50.4\% \\
55\% & 100.0\% & 54.4\% & 59.2\% & 54.9\% \\
\bottomrule
\end{tabular}
\end{table}

\begin{table*}[!tp]
\centering
\caption{\textbf{Grid of the pseudo-label fractions $\beta$ and $\gamma$.} Mean of the four backbones, COCO and AMBER separately; each cell is scorer / \ours{} AUROC; bold: our values.}
\label{tab:C3}
\small
\renewcommand{\arraystretch}{1.15}
\begin{tabular}{lccccc}
\toprule
COCO: $\beta$ $\backslash$ $\gamma$ & 30\% & 40\% & 50\% & 60\% & 70\% \\
\midrule
5\% & 0.888 / 0.959 & 0.895 / 0.959 & 0.892 / 0.959 & 0.892 / 0.958 & 0.879 / 0.957 \\
7.5\% & 0.888 / 0.960 & 0.897 / 0.960 & 0.897 / 0.960 & 0.892 / 0.959 & 0.885 / 0.958 \\
10\% & 0.889 / 0.960 & 0.898 / 0.960 & 0.899 / 0.960 & 0.894 / 0.959 & 0.883 / 0.957 \\
12.5\% & 0.888 / 0.960 & 0.896 / 0.961 & \textbf{0.899 / 0.961} & 0.896 / 0.959 & 0.889 / 0.958 \\
15\% & 0.882 / 0.959 & 0.894 / 0.961 & 0.899 / 0.961 & 0.893 / 0.959 & 0.883 / 0.957 \\
20\% & 0.881 / 0.960 & 0.891 / 0.960 & 0.899 / 0.961 & 0.889 / 0.958 & 0.872 / 0.956 \\
25\% & 0.876 / 0.959 & 0.890 / 0.960 & 0.889 / 0.959 & 0.880 / 0.958 & 0.865 / 0.955 \\
\midrule
AMBER: $\beta$ $\backslash$ $\gamma$ & 30\% & 40\% & 50\% & 60\% & 70\% \\
\midrule
5\% & 0.874 / 0.945 & 0.879 / 0.945 & 0.878 / 0.945 & 0.878 / 0.945 & 0.876 / 0.944 \\
7.5\% & 0.875 / 0.945 & 0.877 / 0.945 & 0.875 / 0.945 & 0.875 / 0.945 & 0.872 / 0.944 \\
10\% & 0.871 / 0.945 & 0.873 / 0.945 & 0.875 / 0.945 & 0.873 / 0.945 & 0.872 / 0.944 \\
12.5\% & 0.865 / 0.944 & 0.867 / 0.945 & \textbf{0.870 / 0.945} & 0.869 / 0.945 & 0.871 / 0.945 \\
15\% & 0.861 / 0.944 & 0.868 / 0.945 & 0.866 / 0.944 & 0.865 / 0.944 & 0.867 / 0.944 \\
20\% & 0.863 / 0.944 & 0.865 / 0.944 & 0.862 / 0.944 & 0.861 / 0.943 & 0.863 / 0.943 \\
25\% & 0.861 / 0.944 & 0.863 / 0.944 & 0.859 / 0.944 & 0.863 / 0.944 & 0.860 / 0.943 \\
\bottomrule
\end{tabular}
\end{table*}

\begin{table*}[!tp]
\centering
\caption{\textbf{Self-training parameters and fusion weight.} Mean of the four backbones; each cell is scorer / \ours{} AUROC, and the fusion-weight rows give only the AUROC of \ours{} ($\lambda=1$: self-asking only; $\lambda=0$: scorer only); bold: our values.}
\label{tab:C4}
\small
\setlength{\tabcolsep}{5pt}
\renewcommand{\arraystretch}{1.1}
\begin{tabular}[t]{llcc}
\toprule
Parameter & Value & COCO & AMBER \\
\midrule
Anchor $a$ & 0.0 & 0.886 / 0.958 & 0.864 / 0.944 \\
 & 0.5 & 0.895 / 0.959 & 0.865 / 0.945 \\
 & 0.6 & 0.899 / 0.960 & 0.868 / 0.945 \\
 & 0.7 & 0.896 / 0.960 & 0.866 / 0.945 \\
 & 0.8 & \textbf{0.899 / 0.961} & \textbf{0.870 / 0.945} \\
 & 0.9 & 0.897 / 0.960 & 0.870 / 0.945 \\
 & 1.0 & 0.892 / 0.959 & 0.872 / 0.945 \\
\midrule
Rounds $J$ & 1 & 0.886 / 0.958 & 0.864 / 0.944 \\
 & 2 & 0.896 / 0.959 & 0.869 / 0.945 \\
 & 5 & 0.900 / 0.961 & 0.870 / 0.945 \\
 & 10 & \textbf{0.899 / 0.961} & \textbf{0.870 / 0.945} \\
 & 15 & 0.899 / 0.961 & 0.870 / 0.945 \\
\bottomrule
\end{tabular}
\hspace{14pt}
\begin{tabular}[t]{llcc}
\toprule
Parameter & Value & COCO & AMBER \\
\midrule
Regularization $C$ & 0.001 & 0.899 / 0.961 & 0.871 / 0.945 \\
 & 0.003 & 0.898 / 0.960 & 0.869 / 0.945 \\
 & 0.01 & \textbf{0.899 / 0.961} & \textbf{0.870 / 0.945} \\
 & 0.03 & 0.898 / 0.960 & 0.869 / 0.945 \\
 & 0.1 & 0.898 / 0.960 & 0.867 / 0.945 \\
\midrule
Fusion weight $\lambda$ & 0.0 & 0.899 & 0.870 \\
 & 0.3 & 0.943 & 0.922 \\
 & 0.5 & 0.957 & 0.939 \\
 & 0.6 & 0.960 & 0.943 \\
 & 0.7 & \textbf{0.961} & \textbf{0.945} \\
 & 0.8 & 0.958 & 0.944 \\
 & 1.0 & 0.946 & 0.937 \\
\bottomrule
\end{tabular}
\end{table*}

\begin{table*}[t]
\centering
\caption{\textbf{Fusion scale, scorer classifier and number of S2 neighbors.} All columns give the AUROC of \ours{} after the change minus that with our setting. Scale: ours fuses ranks; the alternatives apply the same 0.7 / 0.3 weights to raw probabilities ($1-\pyes$ and the sigmoid of the scorer logit), z-scores (standardized by the mean and standard deviation of the cold-start memory) or logits. Classifier: ours is logistic regression; the alternative is a one-hidden-layer MLP (64 / 256 units $\times$ L2 coefficient $10^{-3}$ / $10^{-2}$, early stopping, the same pseudo-labels and self-training procedure), listing the range over the four configurations, with the number of significantly worse configurations in parentheses. Neighbors: ours $k=10$. The baselines are the AUROC of \ours{} in the caption of Tab.~\ref{tab:C1} (Qwen3.5-9B $\cdot$ COCO: 0.969, from another cold-start self-training).}
\label{tab:C5}
\small
\setlength{\tabcolsep}{4pt}
\renewcommand{\arraystretch}{1.15}
\begin{tabular}{lcccccc}
\toprule
 & \multicolumn{3}{c}{Fusion scale} & Classifier & \multicolumn{2}{c}{S2 neighbors} \\
\cmidrule(lr){2-4}\cmidrule(lr){5-5}\cmidrule(lr){6-7}
Backbone $\cdot$ data & Raw probability & z-score & Logit & Four MLPs & $k=3$ & $k=30$ \\
\midrule
LLaVA-1.5-7B $\cdot$ COCO & $-$0.002 & $-$0.003$^*$ & +0.001 & $-$0.006 to $-$0.005 (4) & $-$0.001 & $-$0.001 \\
Qwen3-VL-8B $\cdot$ COCO & $-$0.011$^*$ & $-$0.011$^*$ & $-$0.006$^*$ & $-$0.003 to 0.000 (2) & 0.000 & $-$0.002$^*$ \\
Qwen3.5-9B $\cdot$ COCO & $-$0.004$^*$ & $-$0.004$^*$ & $-$0.004$^*$ & $-$0.002 to $-$0.001 (2) & $-$0.001 & $-$0.002$^*$ \\
Qwen2.5-VL-7B $\cdot$ COCO & $-$0.005 & $-$0.005$^*$ & $-$0.002 & $-$0.007 to $-$0.003 (3) & +0.001 & $-$0.001 \\
\midrule
LLaVA-1.5-7B $\cdot$ AMBER & $-$0.008$^*$ & $-$0.005$^*$ & $-$0.003$^*$ & $-$0.005 to $-$0.003 (4) & $-$0.001$^*$ & +0.001$^*$ \\
Qwen3-VL-8B $\cdot$ AMBER & $-$0.022$^*$ & $-$0.022$^*$ & $-$0.002$^*$ & $-$0.001 to $-$0.001 (0) & 0.000 & 0.000 \\
Qwen3.5-9B $\cdot$ AMBER & $-$0.004$^*$ & $-$0.003$^*$ & +0.001 & $-$0.004 to +0.001 (3) & 0.000 & 0.000 \\
Qwen2.5-VL-7B $\cdot$ AMBER & $-$0.026$^*$ & $-$0.020$^*$ & 0.000 & $-$0.008 to $-$0.004 (4) & $-$0.0004$^*$ & 0.000 \\
\bottomrule
\end{tabular}
\end{table*}

\paragraph{Layer-wise evidence on the four backbones (Figs.~\ref{fig:C2} and \ref{fig:C3}).}
The stage division comes from LLaVA-1.5; the discriminative power of each untrained signal was measured layer by layer on all four backbones.
\begin{itemize}[leftmargin=*,itemsep=1pt,topsep=2pt]
\item \textbf{Hallucinations increase as the caption proceeds} (Fig.~\ref{fig:C2}(a)): the share of hallucinated mentions rises from about 3\%--8\% in the first third to 21\%--41\% in the last third (thirds by character position; each backbone averaged over the COCO and AMBER development sets; 3\%--11\% and 16\%--56\% over the eight groups). The loss of visual information during generation~\cite{li2025the}, which $d$ measures, is reproduced on all four backbones.
\item \textbf{S1} (Fig.~\ref{fig:C2}(b)): close to chance in the first few layers, it rises steeply at depths 0.06--0.3 (layers 8--10 on LLaVA-1.5-7B, inside the object-level writing span of \cite{zhang2025crossmodal}; earlier on the Qwen family) and then slowly declines into a wide plateau that contains all selected layers. The most accurate layer of S1 itself is at the steep rise or right after it, 0.02--0.08 higher than the selected layer.
\item \textbf{S2} (Fig.~\ref{fig:C2}(c)): it peaks at depths 0.34--0.57 on all four backbones, with the selected layer at or next to the peak, direct evidence that S2 shares the scorer's span; on Qwen3.5-9B and Qwen3-VL-8B it rises again to near the peak in the last few layers.
\item \textbf{$e$} (Fig.~\ref{fig:C2}(e)): close to or below chance before a depth of about 0.6, it rises only beyond 0.7--0.8. The peak lies before the last layer on Qwen3-VL-8B, Qwen3.5-9B and LLaVA-1.5-7B; Qwen2.5-VL-7B discriminates only at the last layer, and \ours{} on it is also the least sensitive to the layer of $e$ (Fig.~\ref{fig:C4}).
\end{itemize}

\subsection{Setting values and weights}
\label{sec:C.4}

A group below is one backbone on one dataset (eight groups; Tabs.~\ref{tab:C3} and \ref{tab:C4} list the mean of the four backbones).

\begin{itemize}[leftmargin=*,itemsep=1pt,topsep=2pt]
\item Pseudo-label fractions (Tab.~\ref{tab:C3}): per group, the best cell is higher than our values by at most 0.029 for the scorer and 0.002 for \ours{}, and the AUROC of \ours{} varies over the grid by at most 0.011. $\beta=12.5\%$ and $\gamma=50\%$ are set on development data; the test stream only checks the plateau.
\item Self-training parameters (Tab.~\ref{tab:C4}): per group, the AUROC of \ours{} changes by $-$0.005 to +0.001.
\item Fusion weight (Tab.~\ref{tab:C4}): in each group, the fusion is higher than either route alone, and $\lambda=0.7$ is within 0.002 of the best value.
\item Scale (Tab.~\ref{tab:C5}): 18 of 24 comparisons are significantly lower than rank fusion, and none is significantly higher; the largest drops are on AMBER for Qwen3-VL-8B and Qwen2.5-VL-7B (raw probability $-$0.022 and $-$0.026).
\item Classifier (Tab.~\ref{tab:C5}): with the MLP, the AUROC of \ours{} changes by $-$0.008 to +0.001, with 22 of 32 comparisons significantly worse; for the scorer alone, the change is $-$0.065 to +0.001.
\item Neighbors (Tab.~\ref{tab:C5}): changing $k$ to 3 or 30 changes the AUROC of \ours{} by $-$0.002 to +0.001.
\end{itemize}

\subsection{Differences between COCO and AMBER}
\label{sec:C.5}

Removing S2 makes \ours{} significantly worse in three groups on COCO, but on AMBER only for Qwen2.5-VL-7B (Tab.~\ref{tab:C1}). This is related to rare categories, those with at most 5 mentions in the mention stream of the test runs.

The AUROC of S2 alone is 0.52--0.71 on COCO and 0.62--0.79 on AMBER, and S2 is biased toward rare categories: its rank correlation with category rarity is 0.28--0.66, against $-$0.23 to +0.32 for the other three signals. AMBER has 273--337 categories, and rare categories account for only 5\%--8\% of mentions, yet for 29\%--50\% of the most suspicious 12.5\% by S2 (the fraction that the start evidence labels ``hallucinated''), of which 52\%--67\% are objects actually present in the image; COCO has only 78--79 categories, and rare categories account for 2\%--4\% of mentions and for 6\%--16\% of the most suspicious 12.5\% by S2. Combined with S1, $d$ and $e$ at equal weights into the start evidence, this bias is diluted: rare categories in the most suspicious 12.5\% fall to 13\%--23\% on AMBER and to 1\%--10\% on COCO.

\subsection{Across models: LLaVA-1.5-7B, Qwen2.5-VL-7B, Qwen3-VL-8B, Qwen3.5-9B}
\label{sec:C.6}

\begin{figure*}[!tp]
\centering
\includegraphics[width=0.75\textwidth]{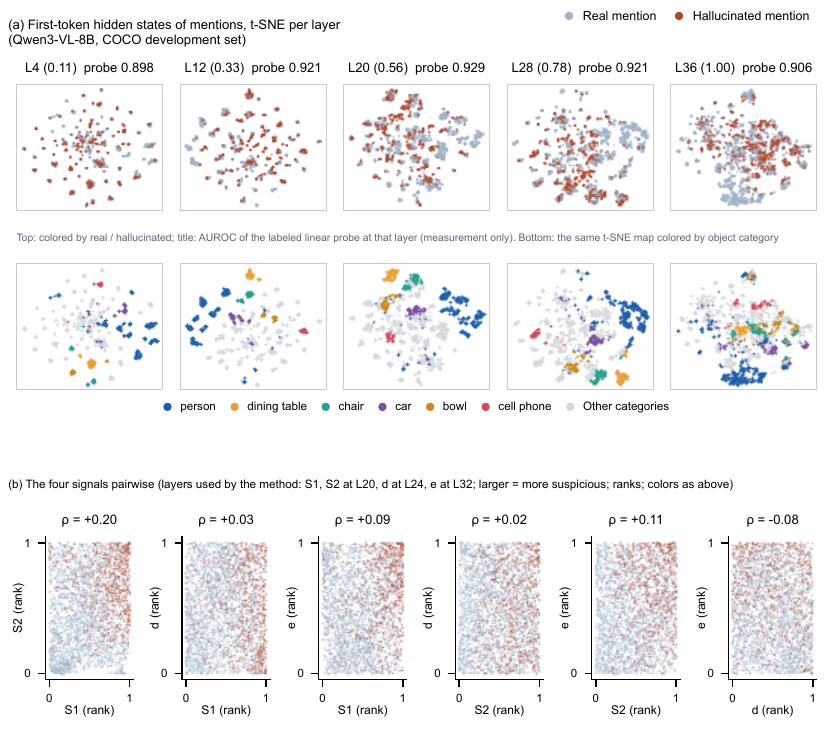}
\caption{Hidden-state structure beyond Fig.~\ref{fig:signal} (Qwen3-VL-8B, COCO development set, 3{,}053 mentions, 651 of them hallucinated). (a) t-SNE of the hidden state at the first token of each mention at L4 / L12 / L20 / L28 / L36, colored real versus hallucinated in the top row (title: AUROC of the labeled linear probe of that layer) and by object category in the bottom row: shallow layers cluster by object category, and at L20 and L28 hallucinations gather into small clusters next to objects of the same category. (b) Pairwise rank scatter plots of the four signals at the layers used, Spearman $\lvert\rho\rvert\le 0.20$: the signals are nearly uncorrelated, and hallucinations lean toward the high end of each, the basis for summing the four signals in the start evidence.}
\label{fig:C5}
\end{figure*}

\begin{figure*}[!tp]
\centering
\includegraphics[width=0.85\textwidth]{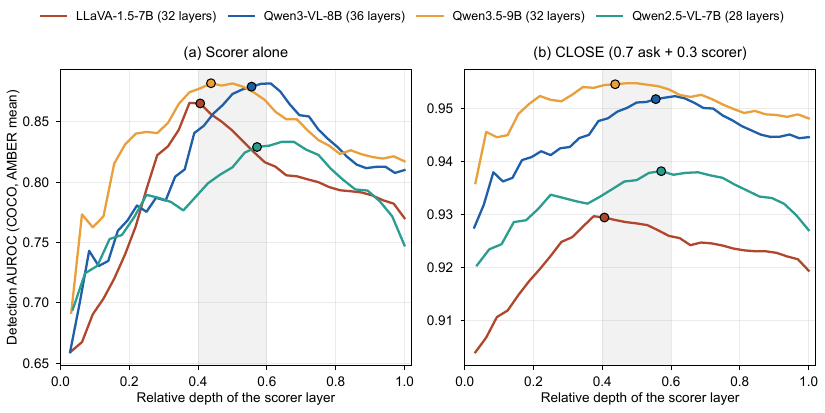}
\caption{Full-depth scan of the scorer layer (other components fixed at their final layers; mean of COCO and AMBER). (a) Scorer-alone AUROC, the solid lines of Fig.~\ref{fig:layer}; (b) AUROC of \ours{}. Gray band: depth 0.4--0.6; dots: selected layers.}
\label{fig:C6}
\end{figure*}

\paragraph{Which layer the scorer reads.}
The scorer layer is the most influential choice and is selected once per backbone on the labeled development set. With the other components fixed at their final layers, the scorer layer is scanned from the first to the last layer; for each layer the scorer is self-trained from scratch on the cold-start memory, and the AUROC of the scorer alone and of \ours{} are computed on the 1{,}160 development images with 20 resamples (Figs.~\ref{fig:C6} and \ref{fig:C4}); the highest point is then confirmed by the layer-switching rule of Sec.~\ref{sec:C.3}.

\begin{table}[H]
\centering
\footnotesize
\setlength{\tabcolsep}{2.2pt}
\renewcommand{\arraystretch}{1.12}
\begin{tabular}{@{}lcccc@{}}
\toprule
 & \shortstack{LLaVA\\(32 layers)} & \shortstack{Qwen3\\(36 layers)} & \shortstack{Qwen2.5\\(28 layers)} & \shortstack{Qwen3.5\\(32 layers)} \\
\midrule
Probe-ceiling peak & L15 (0.47) & L19 (0.53) & L18 (0.64) & L13 (0.41) \\
Best scorer alone & L12 (0.38) & L22 (0.61) & L19 (0.68) & L14 (0.44) \\
Best \ours{} & L12 (0.38) & L22 (0.61) & L16 (0.57) & L16 (0.50) \\
\ours{} plateau$^{\dagger}$ & L11--17 & L18--24 & L13--21 & L11--19 \\
Selected layer & L13 & L20 & L16 & L14 \\
\quad gap to \ours{} best & $-$.0002 & $-$.0006 & 0 & $-$.0002 \\
Scorer share of ceiling & 94\% & 94\% & 91\% & 93\% \\
Start evidence (COCO) & 0.834 & 0.802 & 0.767 & 0.805 \\
Start evidence (AMBER) & 0.852 & 0.820 & 0.788 & 0.814 \\
\bottomrule
\end{tabular}\\[2pt]
\parbox{\columnwidth}{\scriptsize LLaVA, Qwen3, Qwen2.5 and Qwen3.5 denote LLaVA-1.5-7B, Qwen3-VL-8B, Qwen2.5-VL-7B~\cite{bai2025qwen25vl} and Qwen3.5-9B. $^{\dagger}$Within 0.002 of the best.}
\end{table}

In the layer table of this subsection, LLaVA-1.5-7B uses fp16 data, Qwen3-VL-8B double-precision self-asking, and the other two backbones single precision (Tab.~\ref{tab:D1}). The probe ceiling is a labeled linear probe trained per layer (5-fold by image); the scorer share of the ceiling is the AUROC of the best scorer-alone layer divided by the probe peak; the start evidence is the AUROC of the four equally weighted untrained signals on the development images.

\begin{itemize}[leftmargin=*,itemsep=1pt,topsep=2pt]
\item \textbf{The scorer-alone AUROC is an inverted U on all four backbones} (Fig.~\ref{fig:C6}(a)). Depths of the peaks: probe 0.41--0.64, scorer alone 0.38--0.68, deepest on Qwen2.5-VL-7B.
\item \textbf{The layers chosen by the two criteria largely coincide}: the same layer on Qwen3-VL-8B and LLaVA-1.5-7B; 2 layers apart on Qwen3.5-9B and 3 on Qwen2.5-VL-7B (scorer alone ties for highest at L18 and L19), where the best scorer-alone layer still lies within the \ours{} plateau.
\item \textbf{\ours{} is insensitive to the scorer layer}: any layer on the wide plateau around the optimum is within 0.002 in the AUROC of \ours{}; on Qwen3-VL-8B, the selected layer L20 and the \ours{} optimum L22 both lie on the plateau, so the layer is not switched. With the scorer at any layer of relative depth 0.4--0.6, the AUROC of \ours{} is at most 0.004 / 0.004 / 0.003 / 0.001 below the optimum (Qwen3-VL-8B / LLaVA-1.5-7B / Qwen2.5-VL-7B / Qwen3.5-9B), so \textbf{placing it at depth 0.4--0.6 without labels costs at most 0.004}; in the first or last third the loss is larger (up to 0.025 and 0.008 on Qwen3-VL-8B).
\item \textbf{One explanation}: the early layers carry the most image information (on the three Qwen backbones, S1 itself is most accurate at depths 0.06--0.21) but have not yet formed object-level judgments; the late layers serve the choice of the next word ($e$ discriminates only beyond a depth of about 0.75); the middle layers retain both image-related information and object identity, consistent with truthfulness probes in text models working best in the middle layers~\cite{azaria2023the,li2023inferencetime}.
\end{itemize}

\paragraph{The scorer layer follows where hallucinations are most linearly readable.}
\begin{itemize}[leftmargin=*,itemsep=1pt,topsep=2pt]
\item The best layer of the label-free scorer is at most 3 layers from the probe peak on every backbone (3 layers on Qwen3-VL-8B and LLaVA-1.5-7B, 1 on the others). The scorer layer of Qwen3.5-9B is shallower than that of Qwen3-VL-8B (depth 0.44 versus 0.56), consistent with its probe ceiling peaking at depth 0.41; the other components sit at similar relative positions ($d$ 0.72 versus 0.67, $e$ 0.91 versus 0.89), and moving S1 or S2 of Qwen3.5-9B to any layer changes \ours{} by at most 0.003 (Fig.~\ref{fig:C4}).
\item The scorer of Qwen2.5-VL-7B is weaker because of the pseudo-labels, not the representation: its probe ceiling at the selected layer (0.919) equals that of LLaVA-1.5-7B (0.919), but its start evidence is the weakest, and the scorer reaches only 91\% of the ceiling (93\%--94\% for the other three backbones); reducing the bad pseudo-label fraction according to the hallucination rate does not help (on COCO the scorer-alone change is within resampling noise, and on AMBER it is worse). The layer table of this subsection compares the best layer of each backbone, whereas the 90\%--94\% of Analysis~1 in Sec.~\ref{sec:analysis} is the share at the selected layers. The mention sets differ across backbones; by \ours{} at the selected layers (mean of the COCO and AMBER development sets, Fig.~\ref{fig:C4}), the 0.938 of Qwen2.5-VL-7B is higher than the 0.929 of LLaVA-1.5-7B.
\end{itemize}

\paragraph{What transfers across models.}
The layer structure transfers, and all four backbones follow it: the scorer layer is in the middle span (selected layers at depths 0.41--0.57; optimum by the \ours{} criterion at 0.38--0.61 and by the scorer criterion at 0.38--0.68); S1 and S2 work about equally well at any layer within the structural range (Sec.~\ref{sec:C.3}); $d$ is in the middle-to-late span (0.67--0.75); $e$ is in the window before the last layer (0.88--0.91). The exact layer does not transfer: the optimal depth of the scorer layer varies with the backbone, even within the Qwen family (layer table of this subsection). \textbf{Recommendation}: select the layers once for each new model on a small labeled development set; the scorer layer must be scanned, while the other components can be placed by the structure above and then fine-tuned by the \ours{} criterion.

\paragraph{Selection by the \ours{} criterion.}
All four backbones select layers by the procedure of Sec.~\ref{sec:C.3}, with the search restricted to the structural range of each component (Qwen2.5-VL-7B: scorer 10--22, S1 and S2 9--19, $d$ 17--24, $e$ 23--28; Qwen3.5-9B: scorer 10--26, S1 and S2 10--22, $d$ 19--27, $e$ 26--32; for LLaVA-1.5-7B and Qwen3-VL-8B, the ranges are marked in Fig.~\ref{fig:C1}); layers are written as scorer / S1 / S2 / $d$ / $e$. No backbone switches layers.
\begin{itemize}[leftmargin=*,itemsep=1pt,topsep=2pt]
\item \textbf{Qwen2.5-VL-7B}: grid optimum 16 / 18 / 18 / 21 / 26: COCO +0.0005 [$-$0.0056, +0.0051] (11/20), AMBER +0.0046 [+0.0014, +0.0093] (20/20); only one dataset meets the condition.
\item \textbf{Qwen3.5-9B}: grid optimum 14 / 10 / 10 / 23 / 32: COCO $-$0.0001 [$-$0.0018, +0.0020] (6/20), AMBER +0.0021 [+0.0002, +0.0035] (19/20); only one dataset meets the condition.
\item The grid optima of all four backbones are significantly better only on AMBER, and their scorer layers either equal the selected layer or lie within the \ours{} plateau (layer table of this subsection).
\item The chat template of Qwen3.5-9B enables the thinking mode by default; self-asking always disables it (the AUROC of self-asking alone is about 0.80 with it enabled and 0.93--0.95 with it disabled).
\end{itemize}

\subsection{Sources of the four signals}
\label{sec:C.7}

Sources of the ideas: S1 is a contrast with and without the image~\cite{leng2024mitigating,favero2024multimodal,sun2025exploring}, S2 is deep kNN out-of-distribution detection~\cite{sun2022outofdistribution}, and $e$ follows the logit lens, the tuned lens~\cite{belrose2023eliciting} and contrast across layers~\cite{chuang2024dola}; the measure $d$ is proposed in this paper, and the phenomenon it builds on, the loss of visual information during generation, is reported in \cite{li2025the,sun2025exploring}. Fig.~\ref{fig:C5} shows the hidden-state structure at these layers and the pairwise correlation of the four signals.

\FloatBarrier
\setcounter{section}{3}
\providecommand{\Dcell}[2][c]{\begin{tabular}[c]{@{}#1@{}}#2\end{tabular}}
\section{Ablations of the Full Memory Pipeline}
\label{sec:D}

This section ablates the detection and memory pipeline step by step: the contribution of each step, the division of labor between self-asking and the scorer, the threshold, the cold start, and re-training.

\begin{figure*}[!t]
\centering
\includegraphics[width=\textwidth]{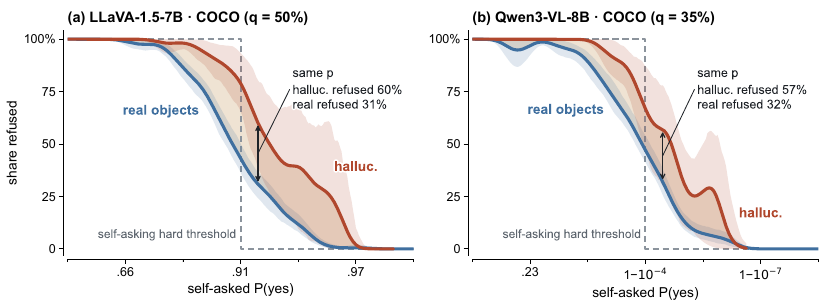}
\caption{\textbf{The correction concentrates near the self-asking threshold} (original captions, COCO development set). The horizontal axis is the self-asked \pyes{}, equally spaced by the rank of the mentions and centered on the self-asking-only threshold. The dashed line is the hard threshold of self-asking alone; the two curves are the refusal rates of hallucinations and of real objects after adding hidden states, with the same number of refusals for both gates; light bands are 95\% intervals from a per-image bootstrap. Over the eight development groups, the decisions changed by fusion account for 4.5\%--6.6\% of the mentions in the four COCO groups and 5.8\%--10.6\% in the four AMBER groups.}
\label{fig:D3}
\end{figure*}

\paragraph{Protocol.} Labels are the official annotations. The \emph{mention stream} denotes the mentions stored in the memory by the test runs of \ours on the four backbones at $q=50\%$ (LLaVA-1.5-7B) and $q=35\%$ (the other backbones): the mentions of the final captions plus every refused mention; mentions accepted but absent from the final caption are excluded, so the stream is at most the number of decisions in Tab.~\ref{tab:D2}. \emph{Replay} denotes a causal, image-by-image pass over the mention stream with the memory, without regenerating captions. Brackets are 95\% intervals from a per-image bootstrap, and $^*$ means that the interval excludes 0; resamples: 2{,}000 for Tab.~\ref{tab:D1}, 400 for Sec.~\ref{sec:D.4}, 300 for Fig.~\ref{fig:D3}.

\subsection{From single signals to fusion: detection \mbox{AUROC}}
\label{sec:D.1}

The figures and Tab.~\ref{tab:D1} in this subsection correspond to Sec.~\ref{sec:analysis}.

\begin{table*}[!t]
\centering
\caption{\textbf{Detection AUROC on the original captions} (development set, 1{,}000 COCO and 160 AMBER images; the scorer is self-trained on the unlabeled cold-start memory and then frozen; self-asking on Qwen3.5-9B and Qwen2.5-VL-7B is read in single precision).}
\label{tab:D1}
\small\setlength{\tabcolsep}{4pt}\renewcommand{\arraystretch}{1.15}
\resizebox{\textwidth}{!}{%
\begin{tabular}{lccccccc}
\toprule
 & \multicolumn{3}{c}{COCO} & & \multicolumn{3}{c}{AMBER} \\
\cmidrule{2-4}\cmidrule{6-8}
Backbone & Scorer & \Dcell{Self-ask alone\\$\to$ \textbf{ours}} & Diff. & & Scorer & \Dcell{Self-ask alone\\$\to$ \textbf{ours}} & Diff. \\
\midrule
LLaVA-1.5-7B & 0.873 & 0.896 $\to$ \textbf{0.919} & +0.023\sig{} [+0.018, +0.027] & & 0.867 & 0.936 $\to$ \textbf{0.942} & +0.005 [$-$0.006, +0.018] \\
Qwen3-VL-8B & 0.893 & 0.941 $\to$ \textbf{0.952} & +0.011\sig{} [+0.007, +0.014] & & 0.870 & 0.946 $\to$ \textbf{0.953} & +0.007\sig{} [+0.000, +0.013] \\
Qwen3.5-9B & 0.890 & 0.949 $\to$ \textbf{0.961} & +0.012\sig{} [+0.008, +0.016] & & 0.893 & 0.939 $\to$ \textbf{0.951} & +0.011\sig{} [+0.006, +0.017] \\
Qwen2.5-VL-7B & 0.845 & 0.934 $\to$ \textbf{0.943} & +0.009\sig{} [+0.002, +0.015] & & 0.834 & 0.926 $\to$ \textbf{0.936} & +0.010 [$-$0.003, +0.023] \\
\bottomrule
\end{tabular}}
\end{table*}

\begin{figure*}[!t]
\centering
\includegraphics[width=0.9\textwidth]{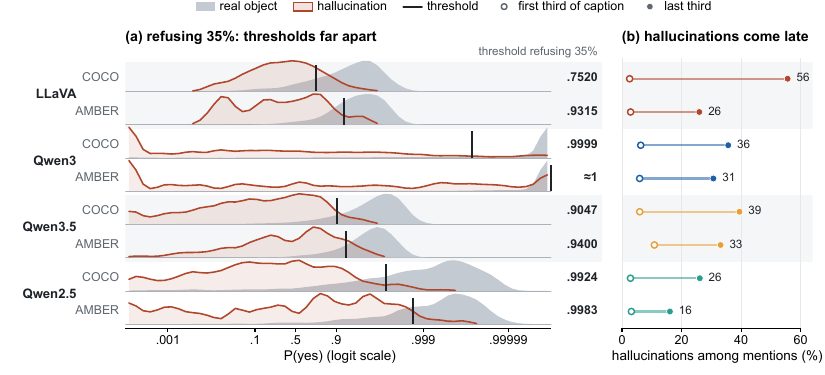}
\caption{(a) Four-backbone version of Fig.~\ref{fig:scale}, adding Qwen3.5-9B and Qwen2.5-VL-7B; the two distributions in each row are normalized separately. (b) Share of hallucinations in the first third (open dots) and the last third (filled dots) of the captions of each group: hallucinations concentrate late. Row labels: LLaVA, Qwen3, Qwen3.5, Qwen2.5 denote LLaVA-1.5-7B, Qwen3-VL-8B, Qwen3.5-9B, Qwen2.5-VL-7B.}
\label{fig:D1}
\end{figure*}

\begin{figure*}[!t]
\centering
\includegraphics[width=\textwidth]{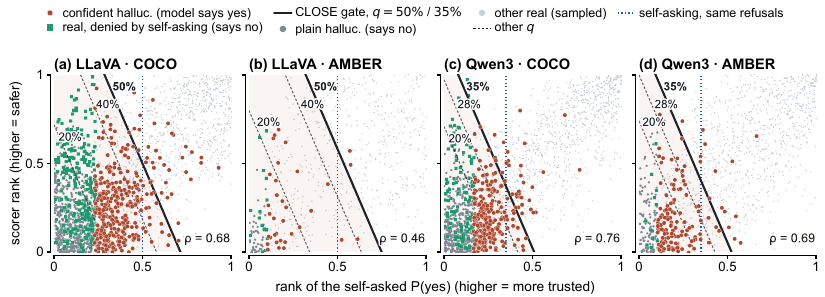}
\caption{Four-group version of Fig.~\ref{fig:ask}, adding the two AMBER groups (original captions, development set). It is read as Fig.~\ref{fig:ask}; $\rho$ is the rank correlation between the self-asking risk and the scorer, and the gate lines are computed from a static ranking.}
\label{fig:D2}
\end{figure*}

\begin{figure*}[!t]
\centering
\includegraphics[width=\textwidth]{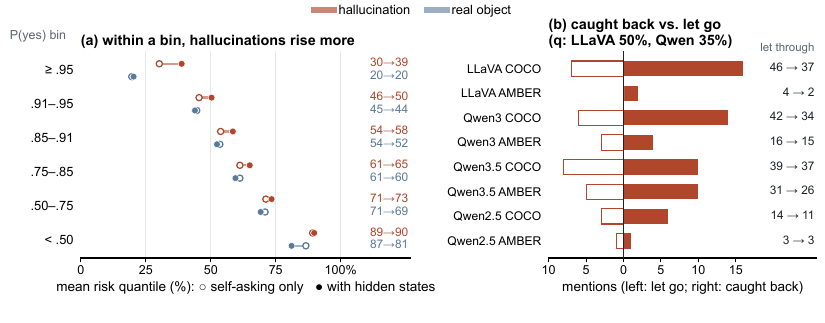}
\caption{\textbf{Hidden states correct self-asking} (original captions, development set). (a) Mean risk quantile (\%) in each bin of Fig.~\ref{fig:stage} (LLaVA-1.5-7B, COCO): after fusion, the risk rank of hallucinations rises more than that of real objects. (b) Hallucinations newly caught and newly let through at the same number of refusals ($q{=}50\%$ for LLaVA-1.5-7B, $35\%$ for the Qwen backbones) in the eight groups; the right column gives the hallucinations let through (self-asking alone $\to$ ours). With a paired per-image bootstrap of 2{,}000 resamples, the net reduction in hallucinations let through is 22 on COCO [6, 39] and 8 on AMBER [$-$2, 19], the latter not significant.}
\label{fig:D4}
\end{figure*}

\paragraph{Further numbers for Analysis~3 (Sec.~\ref{sec:analysis}).} Over the eight development groups, the scorer ranks 68\%--82\% of the confident hallucinations ($\pyes\ge 0.5$) among the riskiest 35\%, but only 17\%--29\% of the real objects with $\pyes\ge 0.5$, and its rank correlation with self-asking is 0.46--0.76. The model answers no to 1\%--9\% of the real objects. At the same number of refusals, 324 real objects are rescued and 302 newly refused on COCO, against 134 and 126 on AMBER. On Qwen3-VL-8B COCO ($q=35\%$), with mentions binned by the self-asked \pyes{} into six bins, the within-bin AUROC rises from 0.49--0.87 with self-asking alone to 0.60--0.90 with hidden states. Near the threshold (within 5 quantile points), on the side that self-asking alone keeps, the refusal rate of hallucinations on Qwen3-VL-8B COCO rises by 64 percentage points and that of real objects by 33; on the side that self-asking alone refuses, the keep rate of real objects rises by 46 (LLaVA-1.5-7B) and 38 (Qwen3-VL-8B) percentage points, against only 15 and 14 for hallucinations.

\subsection{Threshold: memory quantile and fixed thresholds}
\label{sec:D.2}

\begin{table*}[!t]
\centering
\caption{\textbf{Actual refusal rate of \ours} (18 test runs; actual refusal rate = refusals / decisions; quarters follow the order of the images in the test stream).}
\label{tab:D2}
\small\setlength{\tabcolsep}{4pt}\renewcommand{\arraystretch}{1.15}
\begin{tabular}{lcccc}
\toprule
Backbone · data & Target $q$ & Decisions & Actual refusal rate & By quarter of the test stream \\
\midrule
LLaVA-1.5-7B · COCO & 20\% & 1{,}781 & 22.7\% & 22.3 / 24.3 / 22.9 / 21.2\% \\
LLaVA-1.5-7B · COCO & 30\% & 1{,}797 & 33.3\% & 33.5 / 35.4 / 33.5 / 30.8\% \\
LLaVA-1.5-7B · COCO & 40\% & 1{,}757 & 43.3\% & 45.8 / 46.4 / 40.7 / 40.1\% \\
LLaVA-1.5-7B · COCO & 50\% & 1{,}710 & 51.3\% & 54.7 / 50.9 / 50.6 / 48.8\% \\
LLaVA-1.5-7B · AMBER & 20\% & 4{,}186 & 20.5\% & 23.3 / 23.2 / 15.5 / 19.6\% \\
LLaVA-1.5-7B · AMBER & 30\% & 4{,}263 & 30.2\% & 32.8 / 34.3 / 24.4 / 28.8\% \\
LLaVA-1.5-7B · AMBER & 40\% & 4{,}343 & 40.1\% & 42.8 / 42.7 / 35.6 / 39.0\% \\
LLaVA-1.5-7B · AMBER & 50\% & 4{,}366 & 50.6\% & 53.5 / 53.0 / 46.0 / 49.8\% \\
Qwen3-VL-8B · COCO & 20\% & 1{,}569 & 21.5\% & 18.6 / 23.2 / 20.8 / 23.9\% \\
Qwen3-VL-8B · COCO & 28\% & 1{,}589 & 28.6\% & 28.1 / 28.5 / 29.2 / 28.5\% \\
Qwen3-VL-8B · COCO & 35\% & 1{,}581 & 35.2\% & 36.0 / 36.0 / 34.8 / 34.2\% \\
Qwen3-VL-8B · AMBER & 20\% & 6{,}532 & 20.3\% & 18.8 / 22.0 / 19.6 / 20.9\% \\
Qwen3-VL-8B · AMBER & 28\% & 6{,}574 & 28.1\% & 28.1 / 29.0 / 27.4 / 27.8\% \\
Qwen3-VL-8B · AMBER & 35\% & 6{,}531 & 34.5\% & 35.8 / 36.6 / 33.6 / 31.7\% \\
Qwen3.5-9B · COCO & 35\% & 1{,}692 & 36.5\% & 36.1 / 33.2 / 38.0 / 38.3\% \\
Qwen3.5-9B · AMBER & 35\% & 7{,}243 & 35.9\% & 38.3 / 37.0 / 33.3 / 35.0\% \\
Qwen2.5-VL-7B · COCO & 35\% & 1{,}286 & 36.0\% & 39.1 / 33.9 / 38.0 / 32.6\% \\
Qwen2.5-VL-7B · AMBER & 35\% & 4{,}788 & 35.8\% & 39.0 / 35.3 / 31.3 / 37.6\% \\
\bottomrule
\end{tabular}
\end{table*}

Every quarter stays within 6.4 points of the target $q$, with no steady rise across quarters (Tab.~\ref{tab:D2}).

\paragraph{A fixed threshold does not transfer across backbones.} The \pyes{} threshold needed to refuse the same 35\% of mentions differs widely (Fig.~\ref{fig:scale} and Fig.~\ref{fig:D1}(a)).

\subsection{Cold-start size and online re-training of the scorer}
\label{sec:D.3}

\begin{table*}[!t]
\centering
\caption{\textbf{Cold-start size} (the scorer is self-trained on a random subset of the unlabeled cold-start memory and then frozen; AUROC on the mention stream, mean of 5 random seeds; each cell is scorer / \ours{} AUROC; the AMBER cold-start memory has 100 images, so ``all'' is 100 images there).}
\label{tab:D3}
\small\setlength{\tabcolsep}{4pt}\renewcommand{\arraystretch}{1.15}
\begin{tabular}{lccccc}
\toprule
Backbone · data & 25 images & 50 images & 100 images & 150 images & All \\
\midrule
LLaVA-1.5-7B · COCO & 0.865 / 0.942 & 0.870 / 0.944 & 0.889 / 0.947 & 0.895 / 0.948 & 0.902 / 0.949 \\
Qwen3-VL-8B · COCO & 0.850 / 0.961 & 0.833 / 0.961 & 0.883 / 0.965 & 0.901 / 0.966 & 0.905 / 0.967 \\
Qwen3.5-9B · COCO & 0.856 / 0.963 & 0.886 / 0.966 & 0.901 / 0.968 & 0.905 / 0.968 & 0.917 / 0.968 \\
Qwen2.5-VL-7B · COCO & 0.815 / 0.954 & 0.821 / 0.955 & 0.858 / 0.956 & 0.865 / 0.958 & 0.875 / 0.958 \\
LLaVA-1.5-7B · AMBER & 0.847 / 0.943 & 0.861 / 0.945 & -- & -- & 0.866 / 0.945 \\
Qwen3-VL-8B · AMBER & 0.871 / 0.946 & 0.888 / 0.949 & -- & -- & 0.886 / 0.948 \\
Qwen3.5-9B · AMBER & 0.869 / 0.948 & 0.890 / 0.951 & -- & -- & 0.895 / 0.950 \\
Qwen2.5-VL-7B · AMBER & 0.830 / 0.935 & 0.834 / 0.936 & -- & -- & 0.832 / 0.936 \\
\bottomrule
\end{tabular}
\end{table*}

Comparing re-training frequencies by replay on the mention stream, the AUROC of \ours{} with re-training every 100 images differs from that without re-training by $-$0.001 to +0.003, and the actual refusal rate by at most 0.2 points; the AUROC of \ours{} differs by at most 0.001 among re-training every 50, 100 and 200 images.

\begin{table*}[!t]
\centering
\caption{\textbf{The scorer as the number of unlabeled images grows} (the memory is the cold start plus the first $k$ images of the test stream; the scorer is self-trained on it without labels, and AUROC is computed on the held-out end of the test stream: the last 200 images on COCO and the last 300 on AMBER; each cell is scorer / \ours{} AUROC; one pass in the original order).}
\label{tab:D4}
\small\setlength{\tabcolsep}{4pt}\renewcommand{\arraystretch}{1.15}
\begin{tabular}{lcccccc}
\toprule
Backbone · data & Cold start only & +50 images & +100 images & +150 images & +200 images & +250 images \\
\midrule
LLaVA-1.5-7B · COCO & 0.905 / 0.956 & 0.905 / 0.957 & 0.904 / 0.956 & 0.903 / 0.956 & 0.900 / 0.956 & 0.906 / 0.957 \\
Qwen3-VL-8B · COCO & 0.902 / 0.975 & 0.909 / 0.976 & 0.921 / 0.976 & 0.926 / 0.977 & 0.919 / 0.977 & 0.922 / 0.976 \\
\bottomrule
\end{tabular}

\vspace{4pt}
\resizebox{\textwidth}{!}{%
\begin{tabular}{lccccccc}
\toprule
Backbone · data & Cold start only & +100 images & +200 images & +300 images & +400 images & +500 images & +600 images \\
\midrule
LLaVA-1.5-7B · AMBER & 0.863 / 0.949 & 0.888 / 0.951 & 0.884 / 0.950 & 0.876 / 0.949 & 0.881 / 0.949 & 0.880 / 0.949 & 0.873 / 0.948 \\
Qwen3-VL-8B · AMBER & 0.893 / 0.957 & 0.891 / 0.956 & 0.894 / 0.957 & 0.895 / 0.957 & 0.891 / 0.957 & 0.891 / 0.957 & 0.893 / 0.957 \\
\bottomrule
\end{tabular}}
\end{table*}

The scorer improves with the number of images only on Qwen3-VL-8B COCO (reaching about 0.92 after 100 more images) and on LLaVA-1.5-7B AMBER (rising after 100 more images and not further afterwards); for the other two groups the cold start already suffices. \ours{} changes by at most 0.002 relative to the cold start alone: more images improve the scorer, while the ranking of \ours{} is determined mainly by self-asking (Tab.~\ref{tab:D4}).

\subsection{Late hallucinations and the effect of the written caption}
\label{sec:D.4}

This subsection supplements Finding~1 and Sec.~\ref{sec:judge}: why the model writes many hallucinations although self-asking has a high AUROC.

\begin{figure*}[!t]
\centering
\includegraphics[width=\textwidth]{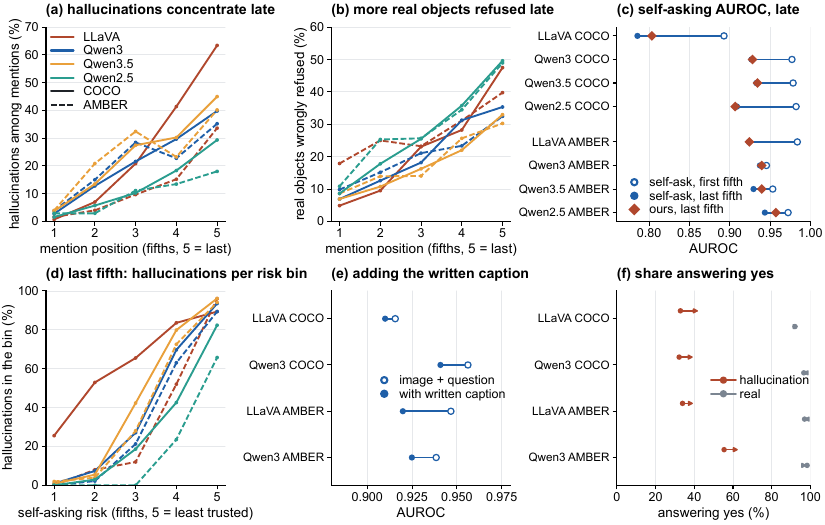}
\caption{\textbf{Late hallucinations and the effect of the written caption.} Solid lines: COCO; dashed lines: AMBER; colors denote backbones. (a)(b) Hallucination share and share of real objects wrongly refused under the global threshold, by position fifth of the mentions. (c) Self-asking AUROC from the first to the last fifth; diamonds mark ours in the last fifth. (d) Hallucination share by self-asking risk fifth within the last fifth. (e)(f) The same question with only the image and the question versus with the written caption added; in (f), red denotes hallucinations, gray real objects, and arrows point to the value with the written caption. Short labels LLaVA, Qwen3, Qwen3.5, Qwen2.5 denote LLaVA-1.5-7B, Qwen3-VL-8B, Qwen3.5-9B, Qwen2.5-VL-7B.}
\label{fig:D5}
\end{figure*}

Fig.~\ref{fig:D5}(a)--(d) uses the original captions on the development set (1{,}000 COCO and 160 AMBER images): the mentions of each group are split into five equal parts by token position, and the global threshold is the self-asking threshold that refuses 35\% of all mentions.

\begin{itemize}[leftmargin=*,itemsep=1pt,topsep=2pt]
\item \textbf{Self-asking weakens late in the caption but does not fail:} in all eight groups the AUROC in the last fifth is lower than in the first.
\item \textbf{What becomes harder late is the threshold:} under the same global threshold, the share of real objects wrongly refused rises from 5\%--18\% in the first fifth to 30\%--50\% in the last, so in the hallucination-rich late part, more real objects are refused along with the hallucinations.
\item \textbf{What cannot be separated lies in the middle risk bins:} except for LLaVA-1.5-7B and Qwen2.5-VL-7B on AMBER (where bin 3 contains only 12\% and 0\% hallucinations), both hallucinations and real objects are numerous in bins 3 and 4 of the last fifth; only LLaVA-1.5-7B COCO is mixed in every bin (Fig.~\ref{fig:D5}(d)).
\end{itemize}

Fig.~\ref{fig:D5}(e)(f) asks the same question twice on the captions of the original model on the test sets (500 COCO and 1{,}004 AMBER images; LLaVA-1.5-7B and Qwen3-VL-8B; the first mention of each object category): ``without'' is image + question, i.e., our self-asking; ``with'' is image + the prompt of the dataset + the caption written up to this mention + the same question.

\begin{itemize}[leftmargin=*,itemsep=1pt,topsep=2pt]
\item \textbf{The written caption pushes hallucinations toward yes, but only partly:} the share of hallucinations answered yes rises by 5.8--9.5 points, while that of real objects changes by at most 2 points; of the hallucinations denied when asked alone, only 14\%--24\% flip to yes when the written caption is added. Hence hallucinations that self-asking detects but the model still writes are not mainly caused by the written text (an observation: generation is not intervened on, and the format and the content of the dialogue change together).
\item \textbf{This is consistent with asking without the written caption:} adding it lowers the self-asking AUROC by 0.006--0.027 in the four groups (from 0.916--0.956 to 0.910--0.941), with 95\% intervals excluding 0 in 3 groups.
\end{itemize}

\paragraph{Numbers of Findings~1 and~2 (Sec.~\ref{sec:intro}).} With $\pyes<0.5$ counted as a denial on the original development captions, LLaVA-1.5-7B denies 62\% / 58\% of its hallucinations on COCO / AMBER and Qwen3-VL-8B 62\% / 44\%, with a \pyes{} AUROC of 0.90 / 0.94 and 0.94 / 0.95. The test sets lead to the same conclusion (32\%--55\% of the hallucinations answered yes, AUROC 0.916--0.956).

\FloatBarrier
\section{Full End-to-End Results}
\label{sec:E}

\subsection{Full metrics of \ours}
\label{sec:E.1}

All results in this section are from test runs. Beyond Tab.~\ref{tab:main}, Tabs.~\ref{tab:E1} and~\ref{tab:E2} add the total rollbacks of each group, the images whose rollbacks reach the per-image limit of 20 (cap hits) and the self-asks per image; Tab.~\ref{tab:E2} also adds, on AMBER, the remaining $q$ and backbones and the rows that use self-asking or the scorer alone, and Tab.~\ref{tab:E3} repeats it on the 744 AMBER images used neither for the cold start nor for development. CHAIR, Cover, Hal, Cog and F1 are in \%; the wall-clock is the complete run time relative to the original model on the same machine, excluding model loading and the cold start.

\begin{table*}[t]
\centering
\caption{\textbf{COCO, 500 images: rollbacks and self-asking} (ours and the comparison rows at $q{=}50\%$ / $35\%$ of Tab.~\ref{tab:main}).}
\label{tab:E1}
\footnotesize
\setlength{\tabcolsep}{4pt}
\renewcommand{\arraystretch}{1.15}
\begin{tabular}[t]{llrrr}
\toprule
Backbone & Group & \shortstack{Roll-\\backs} & \shortstack{Cap\\hits} & \shortstack{Asks /\\image} \\
\midrule
LLaVA-1.5-7B & Ours $q{=}20\%$ & 777 & 0 & 3.56 \\
 & Ours $q{=}30\%$ & 1184 & 1 & 3.59 \\
 & Ours $q{=}40\%$ & 1750 & 6 & 3.51 \\
 & Ours $q{=}50\%$ & 2283 & 16 & 3.42 \\
 & Self-ask $q{=}50\%$ & 2512 & 24 & 3.43 \\
 & Scorer $q{=}50\%$ & 2243 & 3 & 0.00 \\
\bottomrule
\end{tabular}\hspace{2em}%
\begin{tabular}[t]{llrrr}
\toprule
Backbone & Group & \shortstack{Roll-\\backs} & \shortstack{Cap\\hits} & \shortstack{Asks /\\image} \\
\midrule
Qwen3-VL-8B & Ours $q{=}20\%$ & 561 & 1 & 3.14 \\
 & Ours $q{=}28\%$ & 768 & 1 & 3.18 \\
 & Ours $q{=}35\%$ & 1000 & 1 & 3.16 \\
 & Self-ask $q{=}35\%$ & 1049 & 2 & 3.14 \\
 & Scorer $q{=}35\%$ & 1244 & 4 & 0.00 \\
\midrule
Qwen2.5-VL-7B & Ours $q{=}35\%$ & 751 & 1 & 2.57 \\
\midrule
Qwen3.5-9B & Ours $q{=}35\%$ & 1059 & 1 & 3.38 \\
\bottomrule
\end{tabular}
\end{table*}

\begin{table*}[t]
\centering
\caption{\textbf{AMBER, 1{,}004 images} (Qwen2.5-VL-7B and Qwen3.5-9B: only the original model and ours at $q{=}35\%$ are run).}
\label{tab:E2}
\footnotesize
\setlength{\tabcolsep}{4pt}
\renewcommand{\arraystretch}{1.15}
\begin{tabular}{llrrrrrrrrrrrr}
\toprule
Backbone & Group & Real & \shortstack{Lost\\real} & Halluc. & CHAIR & Cover & Hal & Cog & F1 & \shortstack{Roll-\\backs} & \shortstack{Cap\\hits} & \shortstack{Asks /\\image} & Wall \\
\midrule
LLaVA-1.5-7B & Original & 3334 & 0 & 498 & 7.7 & 51.7 & 36.0 & 4.2 & 66.3 & 0 & 0 & 0.00 & 1.00$\times$ \\
 & Ours $q{=}20\%$ & 3180 & 308 & 106 & 2.2 & 50.3 & 9.6 & 0.6 & 66.4 & 1512 & 3 & 4.17 & 1.31$\times$ \\
 & Ours $q{=}30\%$ & 2887 & 629 & 56 & 1.3 & 46.9 & 5.4 & 0.3 & 63.6 & 2710 & 19 & 4.25 & 1.37$\times$ \\
 & Ours $q{=}40\%$ & 2531 & 981 & 31 & 0.9 & 41.8 & 3.0 & 0.1 & 58.8 & 4080 & 41 & 4.33 & 1.44$\times$ \\
 & Ours $q{=}50\%$ & 2105 & 1375 & 19 & 0.5 & 35.5 & 1.8 & 0.1 & 52.3 & 5588 & 80 & 4.35 & 1.50$\times$ \\
 & Self-ask $q{=}50\%$ & 2077 & 1457 & 24 & 0.7 & 34.6 & 2.3 & 0.1 & 51.3 & 5865 & 88 & 4.29 & 1.51$\times$ \\
 & Scorer $q{=}50\%$ & 2067 & 1319 & 42 & 1.6 & 35.4 & 4.1 & 0.2 & 52.1 & 5398 & 66 & 0.00 & 1.40$\times$ \\
\midrule
Qwen3-VL-8B & Original & 5258 & 0 & 1034 & 7.6 & 74.5 & 60.1 & 3.5 & 82.5 & 0 & 0 & 0.00 & 1.00$\times$ \\
 & Ours $q{=}20\%$ & 4885 & 512 & 262 & 2.4 & 70.8 & 22.7 & 0.7 & 82.1 & 1864 & 2 & 6.51 & 1.14$\times$ \\
 & Ours $q{=}28\%$ & 4512 & 898 & 148 & 1.5 & 66.5 & 13.4 & 0.4 & 79.4 & 2770 & 6 & 6.55 & 1.16$\times$ \\
 & Ours $q{=}35\%$ & 4106 & 1297 & 85 & 0.9 & 62.3 & 8.3 & 0.2 & 76.5 & 3620 & 12 & 6.50 & 1.17$\times$ \\
 & Self-ask $q{=}35\%$ & 4050 & 1360 & 97 & 1.1 & 60.7 & 9.3 & 0.2 & 75.2 & 3703 & 15 & 6.42 & 1.15$\times$ \\
 & Scorer $q{=}35\%$ & 3834 & 1525 & 174 & 1.9 & 60.0 & 16.1 & 0.9 & 74.5 & 4212 & 24 & 0.00 & 1.14$\times$ \\
\midrule
Qwen2.5-VL-7B & Original & 4063 & 0 & 357 & 5.0 & 63.8 & 28.5 & 1.6 & 76.3 & 0 & 0 & 0.00 & 1.00$\times$ \\
 & Ours $q{=}35\%$ & 3006 & 1160 & 32 & 0.7 & 50.2 & 3.2 & 0.1 & 66.7 & 2507 & 2 & 4.77 & 1.41$\times$ \\
\midrule
Qwen3.5-9B & Original & 5555 & 0 & 1360 & 9.7 & 75.7 & 70.5 & 5.2 & 82.4 & 0 & 0 & 0.00 & 1.00$\times$ \\
 & Ours $q{=}35\%$ & 4413 & 1310 & 120 & 1.2 & 63.8 & 11.2 & 0.2 & 77.5 & 4018 & 11 & 7.21 & 1.28$\times$ \\
\bottomrule
\end{tabular}
\end{table*}

\begin{table*}[t]
\centering
\caption{\textbf{The 744 AMBER images used neither for the cold start nor for development} (selected per image from the same runs as Tab.~\ref{tab:E2}; no separate wall-clock, marked ``--'').}
\label{tab:E3}
\footnotesize
\setlength{\tabcolsep}{4pt}
\renewcommand{\arraystretch}{1.15}
\begin{tabular}{llrrrrrrrrrrrr}
\toprule
Backbone & Group & Real & \shortstack{Lost\\real} & Halluc. & CHAIR & Cover & Hal & Cog & F1 & \shortstack{Roll-\\backs} & \shortstack{Cap\\hits} & \shortstack{Asks /\\image} & Wall \\
\midrule
LLaVA-1.5-7B & Original & 2454 & 0 & 365 & 7.6 & 51.6 & 36.2 & 4.4 & 66.2 & 0 & 0 & 0.00 & -- \\
 & Ours $q{=}20\%$ & 2335 & 238 & 76 & 2.1 & 50.3 & 9.3 & 0.6 & 66.5 & 1201 & 3 & 4.17 & -- \\
 & Ours $q{=}30\%$ & 2124 & 474 & 43 & 1.5 & 46.9 & 5.5 & 0.3 & 63.5 & 2064 & 15 & 4.25 & -- \\
 & Ours $q{=}40\%$ & 1854 & 737 & 21 & 0.9 & 41.6 & 2.7 & 0.1 & 58.6 & 3139 & 34 & 4.33 & -- \\
 & Ours $q{=}50\%$ & 1540 & 1030 & 12 & 0.5 & 35.3 & 1.6 & 0.1 & 52.1 & 4192 & 65 & 4.32 & -- \\
 & Self-ask $q{=}50\%$ & 1526 & 1085 & 14 & 0.5 & 34.5 & 1.7 & 0.1 & 51.2 & 4372 & 68 & 4.27 & -- \\
 & Scorer $q{=}50\%$ & 1520 & 978 & 32 & 1.6 & 35.4 & 4.2 & 0.3 & 52.1 & 3947 & 43 & 0.00 & -- \\
\midrule
Qwen3-VL-8B & Original & 3879 & 0 & 715 & 7.0 & 74.3 & 58.3 & 3.3 & 82.6 & 0 & 0 & 0.00 & -- \\
 & Ours $q{=}20\%$ & 3585 & 398 & 185 & 2.2 & 70.3 & 22.2 & 0.7 & 81.8 & 1410 & 2 & 6.43 & -- \\
 & Ours $q{=}28\%$ & 3298 & 697 & 106 & 1.4 & 65.9 & 13.2 & 0.4 & 79.0 & 2062 & 5 & 6.47 & -- \\
 & Ours $q{=}35\%$ & 3004 & 987 & 64 & 0.9 & 61.8 & 8.2 & 0.2 & 76.1 & 2686 & 11 & 6.41 & -- \\
 & Self-ask $q{=}35\%$ & 2976 & 1024 & 74 & 1.0 & 60.3 & 9.4 & 0.2 & 74.9 & 2754 & 12 & 6.36 & -- \\
 & Scorer $q{=}35\%$ & 2816 & 1139 & 119 & 1.7 & 59.8 & 14.9 & 0.8 & 74.4 & 3059 & 15 & 0.00 & -- \\
\midrule
Qwen2.5-VL-7B & Original & 2970 & 0 & 272 & 5.1 & 63.1 & 28.8 & 1.7 & 75.8 & 0 & 0 & 0.00 & -- \\
 & Ours $q{=}35\%$ & 2203 & 847 & 21 & 0.5 & 49.8 & 2.8 & 0.1 & 66.4 & 1911 & 2 & 4.77 & -- \\
\midrule
Qwen3.5-9B & Original & 4083 & 0 & 1013 & 9.8 & 75.2 & 70.2 & 5.4 & 82.0 & 0 & 0 & 0.00 & -- \\
 & Ours $q{=}35\%$ & 3241 & 969 & 83 & 1.1 & 63.4 & 10.5 & 0.2 & 77.3 & 3009 & 9 & 7.19 & -- \\
\bottomrule
\end{tabular}
\end{table*}

\subsection{General capability}
\label{sec:E.2}

\begin{table*}[t]
\centering
\caption{\textbf{POPE and RePOPE: \ours{} equals the original model.} All 8{,}910 POPE questions, original model and \ours{} run on the same machine (means over the three original COCO splits; LLaVA official prompts, greedy; scored with the rule of LLaVA's official eval\_pope, yes as the positive class; RePOPE: questions and labels corrected by RePOPE~\cite{neuhaus2025repope}, of which 82 in the random split have no corresponding answer and are not scored).}
\label{tab:E4}
\footnotesize
\setlength{\tabcolsep}{4pt}
\renewcommand{\arraystretch}{1.15}
\begin{tabular}{llcccc}
\toprule
 & & \multicolumn{2}{c}{POPE} & \multicolumn{2}{c}{RePOPE} \\
\cmidrule(lr){3-4}\cmidrule(lr){5-6}
Backbone & Method & Acc & F1 & Acc & F1 \\
\midrule
LLaVA-1.5-7B & Original & 86.78 & 85.79 & 92.04 & 90.63 \\
 & \ours{} ($q{=}50\%$) & 86.78 & 85.79 & 92.04 & 90.63 \\
\midrule
Qwen3-VL-8B & Original & 89.12 & 88.58 & 94.56 & 93.76 \\
 & \ours{} ($q{=}35\%$) & 89.12 & 88.58 & 94.56 & 93.76 \\
\bottomrule
\end{tabular}
\end{table*}

\ours does not act on yes/no benchmarks such as POPE and RePOPE. The answer is only Yes or No and names no object, so the judge never triggers and no self-asking or rollback occurs; our decoding is then token-for-token identical to standard decoding (Sec.~\ref{sec:G.5}). A run on all 8{,}910 POPE questions confirms this: with \ours{} at $q=50\%$ on LLaVA-1.5-7B and $q=35\%$ on Qwen3-VL-8B, the judge is never triggered, and all 8{,}910 answers of each backbone equal those of the original model. The outputs of \ours are therefore those of the original model, and its scores are the same at every $q$, with no loss (Tab.~\ref{tab:E4}).

\subsection{Beam-5 decoding (LLaVA-1.5-7B)}
\label{sec:E.3}

\paragraph{Nominal $q$ and the actual operating point.}
The quantile bound of Sec.~\ref{sec:memory} controls the refusal rate among the memory entries, not the deletion rate of the final text, so the same nominal $q$ corresponds to different operating points under greedy decoding and beam search.

\begin{itemize}[leftmargin=*,itemsep=2pt]
\item \textbf{Setup}: beam size 5, the configuration of Sec.~\ref{sec:method}, the same 500 COCO images, $q{=}30\%$ and $q{=}50\%$ (Tab.~\ref{tab:E5}). The cold-start memory is reused from the greedy runs and is not counted in the cost.
\item \textbf{Enforcing decisions}: a category judged a hallucination prunes the beam that is writing it, mostly without a rollback, so there are far fewer rollbacks than with greedy decoding ($q{=}30\%$: 101 vs.\ 1184; $q{=}50\%$: 305 vs.\ 2283).
\end{itemize}

\begin{table*}[t]
\centering
\caption{\textbf{Beam-5 decoding} (LLaVA-1.5-7B, COCO, 500 images; official CHAIR, reachable in parentheses; F1 as defined in Sec.~\ref{sec:setup}; the original model also uses beam size 5; ``--'': not applicable).}
\label{tab:E5}
\footnotesize
\setlength{\tabcolsep}{4pt}
\renewcommand{\arraystretch}{1.15}
\begin{tabular}{lrrrcccccr}
\toprule
Group & Real & Lost real & Halluc.\ (reach.) & CHAIRs (reach.) & CHAIRi (reach.) & Cover & F1 & Rollbacks & Wall \\
\midrule
Original & 1246 & -- & 434 (422) & 53.8 (52.8) & 14.4 (14.2) & 81.0 & 83.2 & -- & -- \\
Ours $q{=}30\%$ & 1174 & 90 & 163 (152) & 26.8 (25.2) & 6.6 (6.3) & 76.3 & 84.0 & 101 & 1.22$\times$ \\
Ours $q{=}50\%$ & 939 & 320 & 38 (27) & 6.8 (4.8) & 1.7 (1.4) & 61.0 & 75.3 & 305 & 1.23$\times$ \\
\bottomrule
\end{tabular}
\end{table*}

\begin{itemize}[leftmargin=*,itemsep=2pt]
\item Relative to the original model with beam search, CHAIRs drops by 50\% at $q{=}30\%$ and by 87\% at $q{=}50\%$, and CHAIRi by 54\% and 88\%.
\item Removal under beam search is milder: with greedy decoding, CHAIRs drops by 68\% and 93\%; lost real as a share of the original model's real objects is 7.2\% and 25.7\% with beam search and 11.7\% and 33.4\% with greedy decoding.
\item Beam search by itself does not reduce hallucinations: when the original model switches to beam search, CHAIRs rises from 50.0 to 53.8.
\end{itemize}

\FloatBarrier
\onecolumn
\section{Comparison with Published Methods\texorpdfstring{\\}{ }(Reported Values)}
\label{sec:F}

\begin{figure*}[!htbp]
\centering
\includegraphics[width=0.72\textwidth]{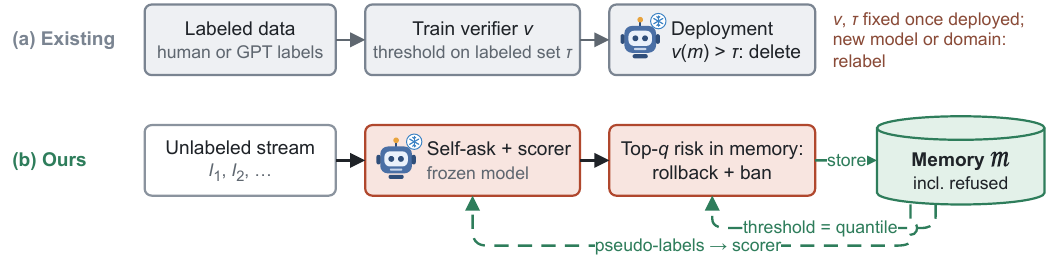}
\caption{\textbf{Alignment with existing in-generation verification along the same pipeline.} The verifier and threshold of existing methods are trained and calibrated on labeled data and fixed after deployment; our judge and threshold use no labels at deployment, the memory is updated after each image, and the threshold is refreshed from the memory quantile every 10 images.}
\label{fig:F1}
\end{figure*}

\begin{figure*}[!htbp]
\centering
\includegraphics[width=0.66\textwidth]{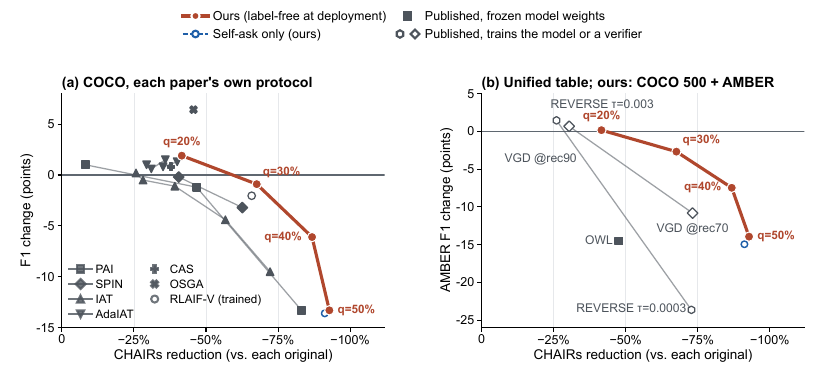}
\caption{\textbf{CHAIRs reduction and F1 change} (LLaVA-1.5-7B). The x-axis is the CHAIRs reduction on COCO relative to each method's original model, and the y-axis the F1 change relative to each original model: (a) COCO F1, with published methods at the values reported in their papers; (b) AMBER F1, for the methods of the unified table, with ours on the 500 COCO images of Sec.~\ref{sec:setup} and all 1{,}004 AMBER images. The red line shows our four settings $q{=}20\%$, $30\%$, $40\%$ and $50\%$, and the blue open circle self-ask only at $q{=}50\%$. The methods in (a) are PAI~\cite{liu2024paying}, SPIN~\cite{sarkar2025mitigating}, IAT and AdaIAT~\cite{zhong2026adaiat}, CAS~\cite{wu2026rethinking}, OSGA~\cite{shi2026oneshot} and RLAIF-V~\cite{yu2024rlaifv} (taken from the OSGA table); in (b), open hexagons are REVERSE, open diamonds VGD and the filled square OWL~\cite{yu2025causally}, which reports under the same protocol. Image sets and F1 definitions differ across papers, so the figure does not compare F1 values. CAS and OSGA keep the weights frozen but fit a steering vector offline (CAS: ridge regression on 161 synthesized, manually verified samples; OSGA: backpropagation on one sample selected by CHAIR); neither uses labels at deployment.}
\label{fig:F2}
\end{figure*}

\subsection{Methods not in Tab.~\ref{tab:compare}}
\label{sec:F.1}

Fig.~\ref{fig:F2} also shows methods not in Tab.~\ref{tab:compare} (SPIN, IAT, AdaIAT, OSGA, RLAIF-V and OWL).

\begin{itemize}[leftmargin=*,itemsep=2pt]
\item \textbf{Reference on a Qwen backbone.} On Qwen2-VL-7B, VGD reduces CHAIRi / CHAIRs by 45\% / 42\% at @rec90 and by 63\% / 64\% at @rec70; its backbone and images differ from ours.
\end{itemize}

\subsection{Accounting of cost}
\label{sec:F.2}

Published methods often report a per-token latency of ``about 1$\times$'', but their full pipelines also include hyperparameters tuned per model on labeled benchmarks (SPIN, PAI, VISTA~\cite{li2025the}), steering vectors fitted by ridge regression on synthesized, manually verified data (CAS) or by backpropagation on a sample selected with ground truth (OSGA), and GPT-generated paired data (HIME~\cite{akl2026hime}, Nullu~\cite{yang2025nullu}); HIME reports CHAIR with beam size 3, and SPIN reads the attention weights of every head at each step and requires eager attention. Our one-time preparation consists of 100--200 unlabeled cold-start images and a development set whose labels are used only for layer selection (Secs.~\ref{sec:C.3} and \ref{sec:C.6}).
\twocolumn

\FloatBarrier
\setcounter{section}{6}
\section{Implementation Details, Configuration and Cost}
\label{sec:G}

\subsection{Setup and configuration}
\label{sec:G.1}

\paragraph{Models and layers.}
Layers are counted from 1. For all four backbones, the layers are selected with the same selection procedure and layer-switching rule of Sec.~\ref{sec:C.3} (four signals with equal weights), on the hidden states of each backbone's 1{,}160 development images; in no case does the grid optimum pass the switching rule (Sec.~\ref{sec:C.6}).

\begin{center}
\small
\setlength{\tabcolsep}{5pt}
\renewcommand{\arraystretch}{1.15}
\begin{tabular}{lcccccc}
\toprule
Backbone (layers) & Precision & Scorer & S1 & S2 & $d$ & $e$ \\
\midrule
LLaVA-1.5-7B (32) & fp16 & 13 & 12 & 14 & 24 & 28 \\
Qwen3-VL-8B (36) & bf16 & 20 & 20 & 20 & 24 & 32 \\
Qwen3.5-9B (32) & bf16 & 14 & 16 & 11 & 23 & 29 \\
Qwen2.5-VL-7B (28) & bf16 & 16 & 16 & 16 & 19 & 25 \\
\bottomrule
\end{tabular}
\end{center}

\paragraph{Decoding and prompts.}
Decoding is greedy (beam size 5 is additionally reported for LLaVA-1.5-7B), with at most 512 new tokens. COCO uses ``Please describe this image in detail.'' and AMBER ``Describe this image.''; each model's official chat template is used, with the thinking mode of Qwen3.5-9B disabled; images for the Qwen family are resized to at most 262{,}144 pixels. The self-asking prompt is ``Is there a \{c\} in this image? Please answer yes or no.'' In the $\pyes$ formula of Sec.~\ref{sec:judge}, $\mathcal{Y}$ and $\mathcal{N}$ are \{Yes, yes\} and \{No, no\}, each word taken as a single token of the model's vocabulary (for LLaVA-1.5-7B, the SentencePiece token with the word-start marker).

\paragraph{Data.}
\begin{itemize}[leftmargin=*,itemsep=1pt,topsep=2pt]
\item Cold start: for COCO, the first 200 of 1{,}000 images drawn with fixed seeds from the 5{,}000 val2017 images; for AMBER, the first 100 of the 1{,}004 generative-task images after a fixed-seed shuffle. Neither overlaps the development set; the COCO cold start does not overlap the COCO test set, and the 100 AMBER cold-start images are among the 1{,}004 but not in the 744-image subset.
\item Development set (layer selection and value setting): 1{,}000 COCO val2017 images (two halves of 500, excluding the 1{,}000 images above and the images used in the design phase) and 160 AMBER images (two halves of 80), 1{,}160 in total.
\item Test set: for COCO, 500 images drawn with seed 1994 from val2014 after removing all val2017 ids and the 882 val2014 images used in the design phase, the same for all four backbones. For AMBER, all 1{,}004 generative-task images by convention (including the cold-start and development images), with the remaining 744 reported separately as a slice of the same run. The CHAIR unified 500 images (the list released by Yue et al.~\cite{yue2024less}) are run additionally as the reproduction check of Sec.~\ref{sec:G.6} (Tab.~\ref{tab:G6}).
\item Neither the fitting nor the online calibration of the verifier reads ground truth.
\item The rules of the reachable count (Sec.~\ref{sec:G.7}) come from two further disjoint groups of 800 COCO val2017 images used in the design phase (the 800 images of the 1{,}000-image draw outside the cold start, and 800 drawn afterwards), which overlap none of the sets above; only their official annotations are used.
\end{itemize}

\paragraph{Baselines.}
The original model; self-ask only (the scorer removed, with the same rollback and laddered ban, and a threshold at the same memory quantile); scorer only; published methods compared by the values reported in their papers (Sec.~\ref{sec:F}).

\paragraph{Configuration of \ours (Tab.~\ref{tab:main}).}
\begin{itemize}[leftmargin=*,itemsep=1pt,topsep=2pt]
\item Operating points: LLaVA-1.5-7B at $q=50\%$, with 20\%, 30\% and 40\% also reported ; Qwen3-VL-8B at $q=35\%$, with 20\% and 28\% also reported; Qwen2.5-VL-7B and Qwen3.5-9B run only at $q=35\%$.
\item Intervention: laddered ban, recognition table, fluency check, English only, at most 20 rollbacks per image (Sec.~\ref{sec:G.3}); after truncation, dangling formatting symbols at the end are removed (symbols only, no letters changed).
\item The outputs of the test runs receive no offline post-processing; costs are measured on the same machine.
\end{itemize}

\paragraph{Implementation options.}
In the test runs, self-asking reuses the KV cache of the image prefix; the scorer layer is read at each step through a forward hook, so that the framework does not return all layers; the memory step reuses the real-image hidden states from generation and re-reads only the gray image; and BLAS threads are limited to 8 when the scorer is re-trained. Re-reading only the gray image saves forward passes, and the others only shorten computation time.

\subsection{Cost}
\label{sec:G.2}

The cold start is done once before deployment and is not counted in the wall-clock: the original model describes and re-reads 200 COCO and 100 AMBER unlabeled images (about 400--700 mentions), and the scorer is self-trained once on the CPU. The wall-clock multiples include the per-image re-reading, memory updates and scorer re-training; they are measured in a single process with exclusive use of one RTX PRO 6000 Blackwell, batch size 1, excluding model loading and the cold start. Over the settings of \ours listed in Tab.~\ref{tab:main} and Tab.~\ref{tab:E2}, the complete wall-clock is 1.07--1.20$\times$ on COCO and 1.14--1.50$\times$ on AMBER.

\subsection{Intervention module: laddered ban, rewrite blocking and external vocabulary}
\label{sec:G.3}

\paragraph{Rewrite blocking.}
After a refusal, the model often writes the same object back in another form, mainly of four kinds: misspellings and non-word compounds (\emph{banannas}, \emph{desklamp}), split forms and derivations (\emph{hand bag}, \emph{cellular device}), a switch of script or language (\emph{teddy} followed by the Chinese word for bear, \emph{toilet} followed by the Russian word for paper in Cyrillic, \emph{mel\'{o}n}), and punctuation joins (\emph{toilet.paper}). This module handles only such hidden hallucinations; it has no tunable similarity threshold, and its only external parameter is the cap of 20 rollbacks. A rewrite into a hypernym or a vague expression is not blocked, because blocking such expressions would remove real objects.

\paragraph{Recognition: is a written word another form of the refused object?}
A hit by any of three parts---normalization, the lookup in the recognition table, and the blocking signal---counts as the same refusal; the table below lists what each part does.

\begin{table*}[t]
\centering
\small
\setlength{\tabcolsep}{4pt}
\renewcommand{\arraystretch}{1.15}
\begin{tabular}{p{0.17\textwidth}p{0.62\textwidth}p{0.15\textwidth}}
\toprule
Part & What it does & What it reads \\
\midrule
Normalization & Same stem (Porter), plural and inflection variants, edit distance of at most 1 (an adjacent transposition counts once), truncated names, split or glued forms (joined with the previous word, the word forms exactly a name, with at most one letter off), a name embedded in a non-word; a real word is compared only with the refused object's own spellings. A stem match counts only for noun use (\emph{seated}, \emph{containing}, \emph{tied} are not \emph{seat}, \emph{container}, \emph{tie}); one letter off applies only to non-words absent from the dictionary (\emph{sick}, \emph{pigs}, \emph{cap} are not \emph{sink}, \emph{pugs}, \emph{car}) & The written text \\
Look up the one-hot encoding of the recognition table & Synonym rows: 1{,}711 words (official spellings, first-sense synonyms, compounds); class rows: 6{,}282 entries (hypernym chain or definition genus word of the WordNet main sense); person rows: 11{,}577 word forms; phrase rule: an adjective derived from the object name followed by an open head noun (\emph{grassy area}, \emph{snowy terrain}, \emph{glacial formations}), with derivation following WordNet and no word vectors. The build uses only WordNet, SemCor usage counts and word vectors (word vectors only for sister classes), followed by a manual revision & The written text + the table \\
Blocking signal & The token just blocked by the decoder, joined with the letters already written, is the beginning of a name of the refused object: the model is writing this name, which counts as a catch on the spot & The generation process \\
\bottomrule
\end{tabular}
\end{table*}

Looking up the one-hot encoding of the recognition table works as follows: each word of the table is one row and each object class one column; a word's row has a 1 in the column of every class the word stands for and 0 elsewhere, and the codes are fixed by the table, not learned. A written word, after normalization, hits a refused object when it has a row and that row has a 1 in the refused object's column.

\paragraph{Manual revision of the recognition table.}
The table derived from WordNet by fixed steps (containing 147 of the 251 official CHAIR synonyms) is revised after an error analysis on the development set: entries that cause false blocks are removed and none is added, 180 in total (55 synonym pairs, 46 class pairs and 79 person words). The removed entries follow five patterns: synonyms from the wrong sense (the first sense of \emph{bike} is motorcycle, so \emph{motorcycle} was treated as \emph{bicycle}; \emph{commode} brought in \emph{can}, \emph{pot} and \emph{stool}); wrong senses in class rows (\emph{lid} as the eyelid, \emph{scooter} as a water scooter); class rows treating a sister class, a part or a place as the same object (\emph{tea} $\to$ \emph{coffee}, \emph{frame} $\to$ \emph{window}, \emph{bathroom} $\to$ \emph{toilet}); person rows using the first sense (\emph{tiger}, \emph{toaster}, \emph{coaster}, \emph{controller}); and person rows with words used mainly as adjectives or surnames on signs (\emph{blonde}, \emph{melancholic}; \emph{Smith}, \emph{Miller}). Replayed on the recognition records of the development-set runs, the recognitions removed are almost all false blocks; the correct ones lost are mainly activity words on AMBER (\emph{paddling}, \emph{sails}) and phrasal expressions on COCO (\emph{entertainment system}, \emph{Dell setups}).

\paragraph{English only.}
After the first refusal in an image, all tokens containing non-ASCII letters are banned, identically on the four backbones; earlier text (e.g., Chinese signs copied from the image) and the original model, which never refuses, are not affected, nor is LLaVA-1.5-7B. Without this rule, Qwen3-VL-8B writes a refused object in Chinese in 83 of its captions on the 1{,}004 AMBER images, and in none with the rule; captions switching script fall from 71 to 0 for Qwen2.5-VL-7B, and captions containing CJK characters from 150 to 2 for Qwen3.5-9B (4 for the original model itself). Foreign names spelled in Latin letters remain (e.g., \emph{urinoir} instead of \emph{urinal} on Qwen3-VL-8B), about once per thousand captions.

\paragraph{Ladder: how far to back off at each catch.}
A refusal by the judge and a hit of the recognition step both count as a ``catch'', and the back-off depends on the number $k$ of catches so far in the current sentence. At $k=1$, this word is backed off (together with the first half of a compound when a noun immediately precedes it, \emph{a large cruise [ship]}); at $k=2$--4, the back-off goes to the start of the noun phrase, where the original first token is banned; at $k\ge 5$, the sentence is closed: if the part before the refused span can stand as a sentence (with a predicate, ending in a noun, with no dangling complement or coordination), a period is added there; otherwise the sentence is deleted and writing continues with the next one. The first sentence is never deleted. The objects of a closed or deleted sentence (including the refused one) are locked: a later catch of a locked object closes its sentence at once, and a second close at the same sentence start ends the caption. Each image has at most 20 rollbacks. The 20th backs off to the start of the current segment, deletes it and continues, with checks as usual; after that, the next refusal deletes its sentence and ends the caption (in the first sentence, a period is added before the refused noun phrase instead). If the cap is used up in the first sentence and no part that can form a sentence remains before the refused position (fewer than 4 words, or not closable), the sentence is kept as a whole and no longer checked, and the object bans are paused so that the model does not evade them by misspelling; checks and bans resume after this sentence, and the next refusal deletes its sentence and ends the caption.

\paragraph{Fluency check: rewritten sentences must read fluently.}
After the first back-off, each completed sentence is checked by rules, with no extra forward pass and no tunable value: garbled non-words (4 letters or more, neither a WordNet word nor in the word list, e.g., \emph{desklab}, \emph{pedestrienses}), a dangling article or preposition at the end of a sentence (\emph{held in the,}), consecutive repeated words or punctuation (\emph{the the}), a whole sentence repeating earlier text (word-trigram overlap of 80\% or more), and the same object coordinated twice (\emph{a tennis racket and a tennis racquet}). A sentence that fails backs off one segment from the first faulty word, counted as one rollback; when the first sentence fails a second time at the same position, it is closed at the preceding point where it can form a sentence.

Non-words are judged with two word lists: all lower-case words in the captions of the two original models (LLaVA-1.5-7B, Qwen3-VL-8B) on the 1{,}000 COCO development images (8{,}199 words), and a general English list, namely all alphabetic entries of the spaCy \texttt{en\_core\_web\_lg} vector table in lower case (247{,}568 words), which keeps image-specific words (\emph{gelato}, \emph{motocross}, \emph{verizon}) from being judged as non-words. Text in quotation marks, italics and URLs (text in the image, Latin scientific names) is, like capitalized names, not judged as a non-word, while recognition and bans still check it. The general list is used only for this rule; ``non-words'' in normalization are still judged by WordNet, so misspellings of a refused object (e.g., \emph{motorcyle}) are still recognized.

\paragraph{Ablation of the intervention components.}
The judge is that of Sec.~\ref{sec:method}, and each group changes only the intervention components (Tab.~\ref{tab:G1}).

\begin{table*}[t]
\centering
\caption{\textbf{Ablation of the intervention components} (development set, 100 images per group; LLaVA-1.5-7B $q=30\%$, Qwen3-VL-8B $q=35\%$; official vocabulary, reachable hallucinations on COCO). ``Single word only'' removes together the laddered back-off, the fluency check and the deletion of the current segment at the rollback cap: each refusal only backs off to before that mention and bans it. ``Write-back'': captions found by the rules to contain another form of a refused object, not checked by blind reading. Wall: complete wall-clock relative to the original model.}
\label{tab:G1}
\small
\setlength{\tabcolsep}{5pt}
\renewcommand{\arraystretch}{1.15}
\begin{tabular}{lrrrrrrrrrr}
\toprule
Group & Real & Lost real & Halluc. & CHAIRs & CHAIRi & Cover & F1 & Rollbacks & Write-back & Wall \\
\midrule
\multicolumn{11}{l}{\textit{LLaVA-1.5-7B $\cdot$ COCO}} \\
Original & 256 & 0 & 95 & 62.0 & 27.1 & 80.8 & 76.6 & 0 & -- & 1.00$\times$ \\
Ours & 211 & 49 & 9 & 9.0 & 4.1 & 66.6 & 78.6 & 329 & 0 & 1.16$\times$ \\
Recognition table off & 223 & 39 & 15 & 15.0 & 6.3 & 70.3 & 80.4 & 201 & 26 & 1.27$\times$ \\
Single word only & 232 & 30 & 40 & 31.0 & 14.7 & 73.2 & 78.8 & 543 & 6 & 1.34$\times$ \\
\midrule
\multicolumn{11}{l}{\textit{LLaVA-1.5-7B $\cdot$ AMBER}} \\
Original & 358 & 0 & 68 & 43.0 & 16.0 & 60.3 & 70.2 & 0 & -- & 1.00$\times$ \\
Ours & 301 & 80 & 6 & 6.0 & 2.0 & 50.7 & 66.8 & 325 & 1 & 1.30$\times$ \\
Recognition table off & 308 & 72 & 8 & 7.0 & 2.5 & 51.9 & 67.7 & 212 & 14 & 1.34$\times$ \\
Single word only & 310 & 71 & 8 & 8.0 & 2.5 & 52.2 & 68.0 & 343 & 5 & 1.33$\times$ \\
\midrule
\multicolumn{11}{l}{\textit{Qwen3-VL-8B $\cdot$ COCO}} \\
Original & 236 & 0 & 57 & 39.0 & 19.5 & 74.4 & 77.4 & 0 & -- & 1.00$\times$ \\
Ours & 195 & 46 & 8 & 6.0 & 3.9 & 61.5 & 75.0 & 207 & 1 & 1.06$\times$ \\
Recognition table off & 193 & 47 & 8 & 6.0 & 4.0 & 60.9 & 74.5 & 126 & 4 & 1.07$\times$ \\
Single word only & 194 & 45 & 9 & 7.0 & 4.4 & 61.2 & 74.6 & 153 & 1 & 1.06$\times$ \\
\midrule
\multicolumn{11}{l}{\textit{Qwen3-VL-8B $\cdot$ AMBER}} \\
Original & 564 & 0 & 104 & 56.0 & 15.6 & 94.9 & 89.4 & 0 & -- & 1.00$\times$ \\
Ours & 437 & 138 & 7 & 7.0 & 1.6 & 73.6 & 84.2 & 388 & 3 & 1.17$\times$ \\
Recognition table off & 458 & 120 & 7 & 7.0 & 1.5 & 77.1 & 86.5 & 274 & 5 & 1.17$\times$ \\
Single word only & 456 & 117 & 11 & 9.0 & 2.4 & 76.8 & 86.0 & 421 & 2 & 1.21$\times$ \\
\bottomrule
\end{tabular}
\end{table*}

\paragraph{External vocabulary (Sec.~\ref{sec:external}).}
The external vocabulary depends on no official vocabulary and is derived from WordNet by fixed rules: of the 55{,}191 one-word nouns consisting of letters only, those whose most common sense is a concrete thing (artifact, organism, food, body part, plant part, natural object, body of water, geological formation) and that WordNet uses mainly as nouns are kept, 21{,}451 words. During generation, the lemma of each candidate noun (lower-case letters only, at least 3 letters, position and color words removed) is used to look up the one-hot encoding of the external vocabulary; a hit, or a word of the official AMBER vocabulary, enters the decision. The rest of the configuration is as in Sec.~\ref{sec:method}, and the cold-start memory is rebuilt with the same trigger rule. On LLaVA-1.5-7B, the complete wall-clock in the runs of Tab.~\ref{tab:G2} is 1.29$\times$ / 1.37$\times$ with the official vocabulary at $q=30\%$ / $40\%$ and 1.37$\times$ / 1.49$\times$ with the external vocabulary, owing to more questions (4.6 $\to$ 6.9 at $q=30\%$) and more rollbacks per image (3.2 $\to$ 3.8); it was not measured on Qwen3-VL-8B. Tab.~\ref{tab:G2} uses labels from blind image judging: on 100 AMBER development images, every noun in the captions is judged as present in the image or not (2{,}047 image--noun pairs for Qwen3-VL-8B, one judge per item, blind to the group), which gives CHAIR, Cover and F1 without relying on the official vocabulary.

\begin{table*}[t]
\centering
\caption{\textbf{External vocabulary and official vocabulary} (100 AMBER development images, labels from blind image judging; hallucinated and real objects are de-duplicated image--noun pairs; ``out of vocab.'' counts the hallucinations whose noun is not in the official AMBER vocabulary)}
\label{tab:G2}
\small
\setlength{\tabcolsep}{6pt}
\renewcommand{\arraystretch}{1.15}
\begin{tabular}{llrrrrrr}
\toprule
Backbone & Scope, $q$ & Real & Halluc. & Out of vocab. & CHAIR & Cover & F1 \\
\midrule
Qwen3-VL-8B & Original & 1273 & 85 & 47 & 6.3 & 78.2 & 85.3 \\
Qwen3-VL-8B & Official $q=20\%$ & 1223 & 58 & 53 & 4.5 & 75.2 & 84.1 \\
Qwen3-VL-8B & Official $q=35\%$ & 1093 & 49 & 49 & 4.3 & 67.2 & 78.9 \\
Qwen3-VL-8B & External $q=20\%$ & 1145 & 18 & 7 & 1.5 & 70.4 & 82.1 \\
Qwen3-VL-8B & External $q=30\%$ & 1039 & 7 & 2 & 0.7 & 63.9 & 77.7 \\
Qwen3-VL-8B & External $q=35\%$ & 990 & 5 & 1 & 0.5 & 60.8 & 75.5 \\
Qwen3-VL-8B & External $q=40\%$ & 826 & 1 & 0 & 0.1 & 50.8 & 67.3 \\
LLaVA-1.5-7B & Original & 483 & 75 & 18 & 13.4 & 78.2 & 82.1 \\
LLaVA-1.5-7B & Official $q=30\%$ & 441 & 29 & 25 & 6.2 & 71.4 & 81.1 \\
LLaVA-1.5-7B & Official $q=40\%$ & 418 & 36 & 33 & 7.9 & 67.6 & 78.0 \\
LLaVA-1.5-7B & External $q=30\%$ & 422 & 7 & 0 & 1.6 & 68.3 & 80.6 \\
LLaVA-1.5-7B & External $q=40\%$ & 377 & 5 & 1 & 1.3 & 61.0 & 75.4 \\
\bottomrule
\end{tabular}
\end{table*}

\begin{itemize}[leftmargin=*,itemsep=1pt,topsep=2pt]
\item No group exceeds the original model in F1. Of the 20 words labeled absent by the official annotation, the single judge marked 4 as present, so the hallucination counts may be underestimated.
\end{itemize}

\paragraph{Check on new images: 500 AMBER images not used in development.}
The first 500 images, in official order, of the 744 AMBER images outside the cold start and the development set are run in three groups on Qwen3-VL-8B: the original model (from the test run), the official vocabulary only and the external vocabulary, both at $q=35\%$. All (image, noun) pairs in the captions of the three groups are de-duplicated and shuffled, and two judges (two independent runs of Claude Opus 5.5) each judge all 9{,}403 items while looking at the image, blind to the group and to each other; on the 7{,}389 items to which both give a present/absent verdict, 96\% agree. An item counts as a hallucination only if both judge it absent (426 items) and as a real object only if both judge it present (6{,}686); the 277 opposite verdicts and the remaining 2{,}014 items (unsure, not an object) are not counted. Differences carry 95\% intervals from 2{,}000 paired bootstrap resamples over images (Tab.~\ref{tab:G3}).

\begin{table*}[t]
\centering
\caption{\textbf{External vocabulary, 500 new images} (Qwen3-VL-8B, $q=35\%$, two blind image judges; counted by (image, noun) pairs, which differs from the counting of the official AMBER script; in / out of vocab.\ denotes whether a noun is in the official AMBER vocabulary)}
\label{tab:G3}
\small
\setlength{\tabcolsep}{8pt}
\renewcommand{\arraystretch}{1.15}
\begin{tabular}{lrrrrrr}
\toprule
 & \multicolumn{3}{c}{Halluc.} & \multicolumn{3}{c}{Real} \\
\cmidrule(lr){2-4}\cmidrule(lr){5-7}
Group & All & \begin{tabular}[b]{@{}c@{}}In\\vocab.\end{tabular} & \begin{tabular}[b]{@{}c@{}}Out of\\vocab.\end{tabular} & All & \begin{tabular}[b]{@{}c@{}}In\\vocab.\end{tabular} & \begin{tabular}[b]{@{}c@{}}Out of\\vocab.\end{tabular} \\
\midrule
Original & 312 & 159 & 153 & 5{,}750 & 2{,}778 & 2{,}972 \\
Official vocabulary only & 210 & 0 & 210 & 4{,}940 & 2{,}002 & 2{,}938 \\
External vocabulary & 9 & 2 & 7 & 4{,}346 & 2{,}311 & 2{,}035 \\
\bottomrule
\end{tabular}
\end{table*}

\begin{itemize}[leftmargin=*,itemsep=1pt,topsep=2pt]
\item With the official vocabulary only, out-of-vocabulary hallucinations are 57 more than for the original model [+29, +85]. With the external vocabulary, out-of-vocabulary hallucinations are 203 fewer than with the official vocabulary only [$-$234, $-$171], and all hallucinations are 303 fewer than for the original model [$-$341, $-$266] ($-$97\%).
\item The cost is in real objects: 594 fewer than with the official vocabulary only [$-$758, $-$428] ($-$12\%), and 1{,}404 fewer than for the original model [$-$1{,}560, $-$1{,}252] ($-$24\%). The refusal rate $q$ is computed over all checked mentions; when the checking scope is enlarged, part of the 35\% refusal budget goes to out-of-vocabulary nouns, so real objects within the vocabulary increase.
\item The direction agrees with the 100 development images (Tab.~\ref{tab:G2}); with 500 images, none of the intervals above contains 0.
\end{itemize}

\paragraph{New data: MMBench scene images.}
Only the images of MMBench~\cite{liu2024mmbench} are used, not its multiple-choice questions: 100 images are drawn with a fixed random seed from the ``scene'' category of en\_dev whose source is not COCO, and none was used for development or tuning. Qwen3-VL-8B runs with the external vocabulary at $q=35\%$ and starts from the unlabeled cold-start memory of AMBER (100 images), with the memory and threshold updated online along the test stream, so this is also a cross-dataset cold start. Two judges independently judge all 1{,}889 (image, noun) pairs blind while looking at the image, the second in reverse order, counted as above: 1{,}505 items are judged present by both, 79 absent by both and 11 opposite; the remaining 294 are not counted (Tab.~\ref{tab:G4}).

\begin{table}[t]
\centering
\caption{\textbf{New data: 100 MMBench scene images} (Qwen3-VL-8B, external vocabulary $q=35\%$, cold-start memory from AMBER; two blind image judges)}
\label{tab:G4}
\small
\setlength{\tabcolsep}{4pt}
\renewcommand{\arraystretch}{1.15}
\begin{tabular}{lrrr}
\toprule
Group & Halluc. & \begin{tabular}[b]{@{}c@{}}Images with\\halluc.\end{tabular} & Real \\
\midrule
Original & 77 & 51 & 1{,}343 \\
External vocabulary $q=35\%$ & 3 & 3 & 1{,}017 \\
\bottomrule
\end{tabular}
\end{table}

\begin{itemize}[leftmargin=*,itemsep=1pt,topsep=2pt]
\item Hallucinations are 74 fewer [$-$94, $-$56] ($-$96\%), and real objects 326 fewer [$-$398, $-$255] ($-$24\%), the same cost as at the same setting on the new AMBER images.
\end{itemize}

\paragraph{Complementarity of the external vocabulary and the recognition table.}
The external vocabulary enlarges the checking scope, so that a refused object rewritten as another real word in it is checked again; the recognition table recognizes rewrites of a refused object, whose misspellings and split forms are not in the external vocabulary. Tab.~\ref{tab:G5} compares the four combinations of the two, with the rest of the configuration unchanged.

\begin{table}[t]
\centering
\caption{\textbf{Combinations of the external vocabulary and the recognition table} (100 AMBER development images; only objects refused within the official vocabulary are counted; ``write-back'' means that a caption contains another form of a refused object that the original model does not write on the same image; candidates are first found with the rules of Sec.~\ref{sec:G.3}, then confirmed one by one by a judge who reads the sentences, blind to the group)}
\label{tab:G5}
\small
\setlength{\tabcolsep}{5pt}
\renewcommand{\arraystretch}{1.15}
\begin{tabular}{llrrr}
\toprule
Scope & \begin{tabular}[b]{@{}c@{}}Recog.\\table\end{tabular} & \begin{tabular}[b]{@{}c@{}}Captions\\with refusals\end{tabular} & \begin{tabular}[b]{@{}c@{}}Write-back\\captions\end{tabular} & \begin{tabular}[b]{@{}c@{}}Write-back\\forms\end{tabular} \\
\midrule
\multicolumn{5}{l}{\textit{LLaVA-1.5-7B, $q=30\%$}} \\
Official & On & 65 & 1 & 1 \\
Official & Off & 63 & 12 & 13 \\
External & On & 54 & 0 & 0 \\
External & Off & 55 & 11 & 11 \\
\midrule
\multicolumn{5}{l}{\textit{Qwen3-VL-8B, $q=35\%$}} \\
Official & On & 83 & 2 & 2 \\
Official & Off & 83 & 4 & 4 \\
External & On & 63 & 1 & 1 \\
External & Off & 64 & 1 & 1 \\
\bottomrule
\end{tabular}
\end{table}

\begin{itemize}[leftmargin=*,itemsep=1pt,topsep=2pt]
\item The recognition table reduces write-backs the most, the external vocabulary reduces them further, and the fewest occur with both on; on Qwen3-VL-8B, the external vocabulary alone ties with the combination, at 1 caption each. The external vocabulary cannot replace the recognition table. The counts are small, and the results are exploratory.
\end{itemize}

\subsection{Exact rollback on hybrid linear-attention backbones}
\label{sec:G.4}

The linear-attention layers of Qwen3.5 store recurrent states, which cannot be truncated like a KV cache. During generation, state snapshots of the last 16 steps are kept (including the logits of the next token); at a rollback they are restored directly, and the full-attention layers truncate the extra positions. On Qwen3.5-9B (4 images $\times$ 4 truncation points), the 30 tokens continued after restoring are identical position by position to step-by-step generation (logit difference 0); if instead the prompt state is read in again in one pass, the continuation diverges in 3 of the 16 cases (the chunked and step-by-step algorithms differ slightly in bf16).

\subsection{Token-level agreement with standard decoding when the intervention is off}
\label{sec:G.5}

All groups in this paper, including the original model, use the same decoding function. With the intervention off, its outputs agree token by token with those of transformers \texttt{model.generate} (greedy, \texttt{repetition\_penalty=1.0}, the same inputs; when two logits are exactly equal in bf16, the token with the smaller id is taken, as \texttt{generate} does): LLaVA-1.5-7B 12/12 (greedy and beam size 5), Qwen2.5-VL-7B 6/6, Qwen3-VL-8B 6/6, and Qwen3.5-9B 6/6 after removing one trailing newline after \texttt{<|im\_end|>} (this paper stops at \texttt{<|im\_end|>} following the chat template).

\subsection{Reproduction of the unified evaluation protocol}
\label{sec:G.6}

The numbers of REVERSE~\cite{wu2025generate} and VGD~\cite{yang2026hallucinations} in Tab.~\ref{tab:compare} come from their unified setting: the CHAIR unified 500 images (the val2014 image list released by Yue et al.~\cite{yue2024less}) and the AMBER generative task, with the prompts of Sec.~\ref{sec:G.1}, the original LLaVA-1.5 conversation format, fp16 and greedy decoding. The rows of \ours use the 500 COCO images of Sec.~\ref{sec:setup} (same prompt, decoding and evaluation script, different images) and all 1{,}004 AMBER images, so Tab.~\ref{tab:compare} compares reductions relative to each paper's original model. Of the unified 500 images, 3 are cold-start images of this paper, 17 are in the development set, and 59 more were used in the design phase (47 val2017 and 12 val2014 images), so the test set is drawn separately (Sec.~\ref{sec:G.1}); the two 500-image sets share only 7 images.
\begin{itemize}[leftmargin=*,itemsep=1pt,topsep=2pt]
\item \textbf{Consistent evaluation.} Scoring the original-model captions released by Yue et al.\ with our CHAIR code gives CHAIRs 50.0 / CHAIRi 15.4, identical to the unified-protocol table of REVERSE / VGD; after tokenization, the prompt is token-identical to the original.
\item \textbf{AMBER reproduction.} Original model CHAIR 7.7 / Cover 51.7 / Hal 36.0 / Cog 4.2 (in that table 7.8 / 51.0 / 36.4 / 4.2); on the discriminative questions, following the official practice, Acc 71.6 / F1 74.5 (AMBER paper 72.0 / 74.7).
\item \textbf{Differences of the original model.} Our original model obtains on the unified 500 images CHAIRs 50.2 / CHAIRi 13.9, versus 50.0 / 15.4 in the unified-protocol table. Only 20 / 500 of its captions are identical word for word to the released ones (21 / 500 with eager attention); the difference comes from the software stack (the original LLaVA repository with an old transformers version versus the new HF implementation), so Tab.~\ref{tab:compare} lists both original-model rows.
\item \textbf{Runs of \ours on the unified 500 images} (Tab.~\ref{tab:G6}; the cold-start and development images among them are not removed). The reductions agree with those on the test set of Sec.~\ref{sec:setup} (68\% / 69\% at $q=30\%$, 93\% / 94\% at $q=50\%$); F1 drops more on these images (on the test set 82.9 $\to$ 82.0 / 69.6).
\end{itemize}

\begin{table*}[t]
\centering
\caption{\textbf{Runs on the CHAIR unified 500 images} (LLaVA-1.5-7B, official CHAIR script; the cold start is still the 200 val2017 images of Sec.~\ref{sec:setup}; parentheses: reachable; changes are relative to the original model in this table; cost not measured)}
\label{tab:G6}
\small
\setlength{\tabcolsep}{4pt}
\renewcommand{\arraystretch}{1.15}
\begin{tabular}{lrrrrcrcrrrr}
\toprule
Group & Real & \begin{tabular}[b]{@{}c@{}}Lost\\real\end{tabular} & Halluc. & \begin{tabular}[b]{@{}c@{}}Reach.\\halluc.\end{tabular} & \begin{tabular}[b]{@{}c@{}}CHAIRs\\(reach.)\end{tabular} & Change & \begin{tabular}[b]{@{}c@{}}CHAIRi\\(reach.)\end{tabular} & Change & Cover & F1 & \begin{tabular}[b]{@{}c@{}}Roll-\\backs\end{tabular} \\
\midrule
Original & 1250 & 0 & 430 & 424 & 50.2 (49.0) & & 13.9 (13.7) & & 78.8 & 82.3 & 0 \\
Ours $q=30\%$ & 1065 & 202 & 96 & 90 & 17.4 (16.2) & $-$65\% & 4.1 (3.9) & $-$71\% & 67.1 & 79.0 & 1226 \\
Ours $q=50\%$ & 800 & 458 & 27 & 17 & 5.2 (3.2) & $-$90\% & 1.6 (1.2) & $-$88\% & 50.4 & 66.7 & 2423 \\
\bottomrule
\end{tabular}
\end{table*}

\subsection{Reachable hallucinations and blind image judging}
\label{sec:G.7}

\paragraph{Reachable hallucinations: a count of this paper.}
The official CHAIR script and vocabulary are unchanged, and all methods report the official numbers as usual; the rows of \ours additionally report the reachable count, side by side with the official one.
\begin{itemize}[leftmargin=*,itemsep=1pt,topsep=2pt]
\item \textbf{Computation.} The count is computed purely by rules, without human or model judgment. Each caption is read again with a rule-based extractor: the synonym table and two-word table (e.g., \emph{traffic light}, \emph{teddy bear}, \emph{toilet seat}) are the same as in the official CHAIR, but word forms are lemmatized only as nouns (WordNet noun lemmas), so \emph{seated}, \emph{tied} and \emph{bearing} are not, as in the official script, tagged for part of speech and then lemmatized to \emph{seat}, \emph{tie} and \emph{bear}; negated mentions (\emph{no dog}) are skipped; \emph{orange} used as a color (\emph{an orange cat}) does not count as the fruit. When \emph{toilet} also occurs in the caption, the official script does not count \emph{seat} as a chair, and the extractor does the same. An object category is a reachable hallucination only if two conditions both hold: the official script counts it as a hallucination of this image, and the extractor also reads it as an object mention in this caption. During generation, the judge checks exactly these extracted mentions; they are what the judge can ``reach'', hence the name.
\item \textbf{Source of the rules.} The rules come from an error analysis on the original-model captions of the two groups of design-phase COCO val2017 images (800 each, Sec.~\ref{sec:G.1}): the items that the official script counts as hallucinations but the extractor does not read as object mentions are grouped by category and triggering form. The most frequent are \emph{orange} as a color, \emph{seated} lemmatized to \emph{seat} and then counted as a chair, and \emph{tie} / \emph{tied} as verbs; the items fall into three patterns (not a noun; a noun not referring to the object; negation), written as the rules above. Prior work reports the same literal-matching miscounts of CHAIR: THRONE~\cite{kaul2024throne}, ALOHa~\cite{al2024aloha} and Feng et al.~\cite{feng2024more} point out that word-by-word matching against a fixed vocabulary counts words that do not refer to objects as objects.
\item \textbf{Check of the rules.} In the COCO test runs (all settings of LLaVA-1.5-7B and Qwen3-VL-8B), the items removed by the rules number 232 after de-duplication by sentence; they were read in two independent passes by Claude Opus 5.5 (sentences only, without the image): 221 indeed do not refer to the object (verbs, color words, other objects), 7 are negations, and the two passes agree completely. The other 4 all come from ``orange fruit'' in the same image, which can be read either as ``oranges'' or as ``orange-colored fruit''; the reading takes the former, while the rule reads the modifier before the noun as a color, so the count of this image differs by 1.
\item \textbf{What the rules remove.} On the COCO 500 test runs, the rules remove 127 of the 440 official hallucinations of the original Qwen3-VL-8B and 123 of the 137 of \ours at $q=35\%$ (Tab.~\ref{tab:main}). The table below lists the removed items by rule (the totals equal Tab.~\ref{tab:main}).
\begin{table*}[t]
\centering
\footnotesize
{\itshape Items removed by the reachable rules on the COCO 500 test runs}\\[3pt]
\small
\setlength{\tabcolsep}{6pt}
\renewcommand{\arraystretch}{1.15}
\begin{tabular}{lrrrrrrrr}
\toprule
Group & \begin{tabular}[b]{@{}c@{}}Off.\\halluc.\end{tabular} & Reach. & Removed & \begin{tabular}[b]{@{}c@{}}\emph{orange}\\color\end{tabular} & \begin{tabular}[b]{@{}c@{}}Non-noun\\forms\end{tabular} & Neg. & Other & \begin{tabular}[b]{@{}c@{}}Words /\\caption\end{tabular} \\
\midrule
LLaVA-1.5-7B orig. & 431 & 422 & 9 & 2 & 5 & 2 & 0 & 91.7 \\
Qwen3-VL-8B orig. & 440 & 313 & 127 & 80 & 44 & 2 & 1 & 318.3 \\
Qwen2.5-VL-7B orig. & 214 & 163 & 51 & 26 & 24 & 1 & 0 & 156.2 \\
Qwen3.5-9B orig. & 491 & 351 & 140 & 83 & 57 & 0 & 0 & 332.4 \\
Qwen3-VL-8B ours $q{=}35\%$ & 137 & 14 & 123 & 70 & 50 & 2 & 1 & 309.2 \\
\bottomrule
\end{tabular}
\end{table*}
\begin{table*}[t]
\centering
\caption{\textbf{Hallucination counts under three measures} (test runs; official = official CHAIR script; reachable = official minus the items removed by the three rules of the Computation paragraph; the later columns classify the reachable items by blind image judging; bold: the minimum of each hallucination-count column)}
\label{tab:G7}
\footnotesize
\setlength{\tabcolsep}{4pt}
\renewcommand{\arraystretch}{1.15}
\begin{tabular}{lrrrrrrrrr}
\toprule
Group & \begin{tabular}[c]{@{}c@{}}Official\\halluc.\end{tabular} & Reachable & \begin{tabular}[c]{@{}c@{}}Absent in\\both passes\end{tabular} & \begin{tabular}[c]{@{}c@{}}Absent in at\\least one pass\end{tabular} & \begin{tabular}[c]{@{}c@{}}Word-sense\\mismatch\end{tabular} & \begin{tabular}[c]{@{}c@{}}Present in\\image\end{tabular} & \begin{tabular}[c]{@{}c@{}}Category\\boundary\end{tabular} & \begin{tabular}[c]{@{}c@{}}Other\\disagreement\\or unclear\end{tabular} & \begin{tabular}[c]{@{}c@{}}Images with\\image-confirmed\\halluc.\end{tabular} \\
\midrule
Original & 440 & 313 & 58 & 99 & 114 & 33 & 41 & 26 & 54 / 500 \\
Ours $q=20\%$ & 192 & 67 & 8 & 15 & 3 & 25 & 14 & 10 & 8 / 500 \\
Ours $q=28\%$ & 160 & 38 & 3 & 6 & 1 & 18 & 6 & 7 & 3 / 500 \\
Ours $q=35\%$ & \textbf{137} & \textbf{14} & \textbf{0} & \textbf{2} & 1 & 9 & 1 & 1 & \textbf{0 / 500} \\
Self-ask only $q=35\%$ & 143 & 21 & 2 & \textbf{2} & 2 & 13 & 2 & 2 & 2 / 500 \\
\bottomrule
\end{tabular}
\end{table*}
The most frequent non-noun forms are \emph{seated} and \emph{tied}: 22 and 11 for the original Qwen3-VL-8B, 18 and 6 for Qwen2.5-VL-7B, and 28 and 20 for Qwen3.5-9B; the 5 items of LLaVA-1.5-7B are \emph{seating} (3), \emph{seated} (1) and \emph{skiing} (1). The three Qwen backbones lose far more items to the rules than LLaVA-1.5-7B; the counts point to two sources, frequent use of \emph{orange} as a color and of \emph{seated} and \emph{tied}, and their captions are also 1.7--3.6$\times$ as long. ``Other'' denotes items that fit none of the three rules.
\end{itemize}

\paragraph{Blind image judging (Qwen3-VL-8B, COCO 500 images).}
This check shows how many of the official and reachable hallucinations are confirmed by looking at the image. All reachable hallucinations of the original model, the three settings of \ours and self-ask only at $q=35\%$ (318 items after de-duplication by image and category) are pooled, 35 objects annotated as real are added as controls, and the items are shuffled and judged twice independently by Claude Opus 5.5 (the second pass in reverse order), blind to the group and the annotation. Each item is judged as absent from the image, present in the image, category boundary, word-sense mismatch (the written word does not refer to this category in the sentence, e.g., \emph{vessel} in \emph{vessel sink} counted as a boat) or unclear; only an item judged absent in both passes counts as an image-confirmed hallucination (Tab.~\ref{tab:G7}); out-of-vocabulary words in the captions are not judged.

\begin{itemize}[leftmargin=*,itemsep=1pt,topsep=2pt]
\item The agreement of the two passes is 81.8\% ($\kappa=0.76$); none of the 35 controls is judged absent in both passes (2 are judged absent in one pass). The columns ``present in image'' and ``category boundary'' are annotation misses.
\item Another model (Claude Sonnet) independently re-judged a sample of 60 items (20 each of: absent in both passes; present in image or category boundary in both passes; word-sense mismatch or disagreement; fixed random seed): it judged 17, 0 and 3 of them absent in the three groups, respectively; with the binary split ``absent / other'', its agreement with ``absent in both passes'' is 54 / 60.
\end{itemize}

\end{document}